\documentclass[10pt,journal,compsoc]{IEEEtran}

\ifCLASSOPTIONcompsoc
  \usepackage[nocompress]{cite}
\else
  \usepackage{cite}
\fi

\usepackage[utf8]{inputenc} 
\usepackage[T1]{fontenc}    
\usepackage{url}            
\usepackage{booktabs}       
\usepackage{amsfonts}       
\usepackage{amsmath}        
\usepackage{nicefrac}       
\usepackage{microtype}      
\usepackage{graphicx}
\usepackage{makecell}
\usepackage{multirow}
\usepackage{subcaption}
\newcommand{\ra}[1]{\renewcommand{\arraystretch}{#1}}
\usepackage[pagebackref=false,breaklinks=true,colorlinks,urlcolor=blue,citecolor=blue,linkcolor=blue,bookmarks=false]{hyperref}
\usepackage{tikz}

\newcommand{\etal}{\textit{et~al}\mbox{.}}
\newcommand{\eg}{e.g.,\ }
\newcommand{\ie}{i.e.,\ }
\newcommand{\bbR}{{\mathbb{R}}}

\newcommand{\revised}[1]{#1}

\newlength{\threeimg}

\newlength{\siximg}

\newlength{\eightimg}

\newlength{\tenimg}
\newlength{\elevenimg}

\newlength{\rowmargin}

\newlength\paramargin
\newlength\figmargin
\newlength\secmargin
\newlength\figcapmargin

\newboolean{revising}
\setboolean{revising}{true}

\ifthenelse{\boolean{revising}} {
\newcommand{\yylin}[1]{\textcolor{blue}{#1}}}
{
\newcommand{\yylin}[1]{}
}

\newcommand{\mpage}[2]
{
\begin{minipage}[b]{#1}\centering
#2
\end{minipage}
}

\begin{document}

\title{Show, Match and Segment: Joint \revised{Weakly Supervised} Learning of Semantic Matching and Object Co-segmentation}

\author{
Yun-Chun~Chen,
Yen-Yu~Lin,
Ming-Hsuan~Yang,
and~Jia-Bin~Huang
\IEEEcompsocitemizethanks{
\IEEEcompsocthanksitem Y.-C. Chen is with the Research Center for Information Technology Innovation, Academia Sinica, Taipei 115, Taiwan, 
E-mail: ycchen918@gmail.com 
\IEEEcompsocthanksitem Y.-Y. Lin is with the Department of Computer Science, National Chiao Tung University, Hsinchu 300, Taiwan,
E-mail: lin@cs.nctu.edu.tw
\IEEEcompsocthanksitem M.-H. Yang is with School of Engineering, University of California, Merced, CA, US, 
Google Research, and Yonsei University, 
E-mail: mhyang@ucmerced.edu
\IEEEcompsocthanksitem J.-B. Huang is with Department of Electrical and Computer Engineering, Virginia Tech, VA, US, 
E-mail: jbhuang@vt.edu}
%
}

\markboth{}%
{Shell \MakeLowercase{\textit{et al.}}: Bare Demo of IEEEtran.cls for Computer Society Journals}

\IEEEtitleabstractindextext{%

\begin{abstract}
We present an approach for jointly matching and segmenting object instances of the same category within a collection of images.
In contrast to existing algorithms that tackle the tasks of semantic matching and object co-segmentation in isolation, our method exploits the complementary nature of the two tasks.
The key insights of our method are two-fold.
First, the estimated dense correspondence \revised{fields} from semantic matching \revised{provide} supervision for object co-segmentation by enforcing consistency between the predicted masks from a pair of images.
Second, the predicted object masks from object co-segmentation in turn allow us to reduce the adverse effects due to background clutters for improving semantic matching.
Our model is end-to-end trainable and does not require supervision from manually annotated correspondences and object masks.
%
%
We validate the efficacy of our approach on \revised{five} benchmark datasets: TSS, Internet, PF-PASCAL, PF-WILLOW, \revised{and SPair-$71$k}, and show that our algorithm performs favorably against the state-of-the-art methods on both semantic matching and object co-segmentation tasks.
\end{abstract}

}

\maketitle

\IEEEdisplaynontitleabstractindextext

\IEEEpeerreviewmaketitle

\setlength{\threeimg}{0.328\textwidth}
\begin{figure*}[t]
  \centering
  \begin{subfigure}[b]{\threeimg}
    \centering\includegraphics[width=0.8\linewidth]{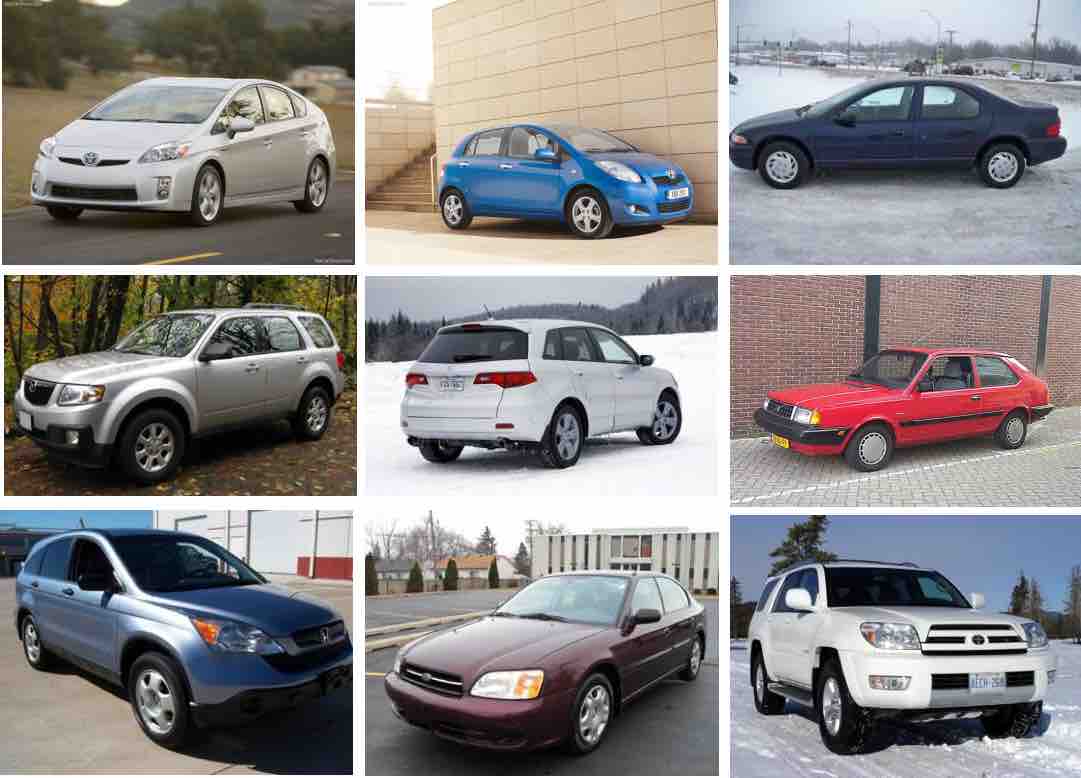}
    \vspace{\figcapmargin}
    \centering\caption*{A collection of images}
  \end{subfigure}
  \tikz{\draw[-,blue, densely dashed, thick](0,-1.8) -- (0,2.1);}
  \begin{subfigure}[b]{\threeimg}
    \centering\includegraphics[width=0.8\linewidth]{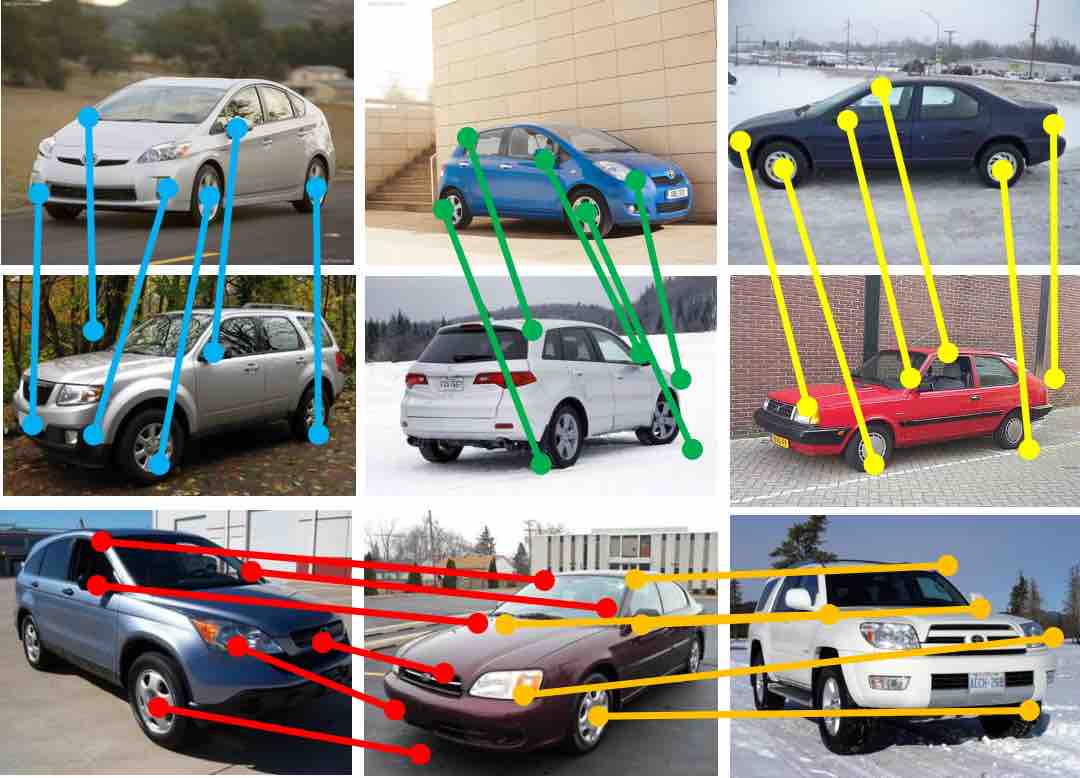}
    \vspace{\figcapmargin}
    \centering\caption*{Semantic matching}
  \end{subfigure}
  \begin{subfigure}[b]{\threeimg}
    \centering\includegraphics[width=0.8\linewidth]{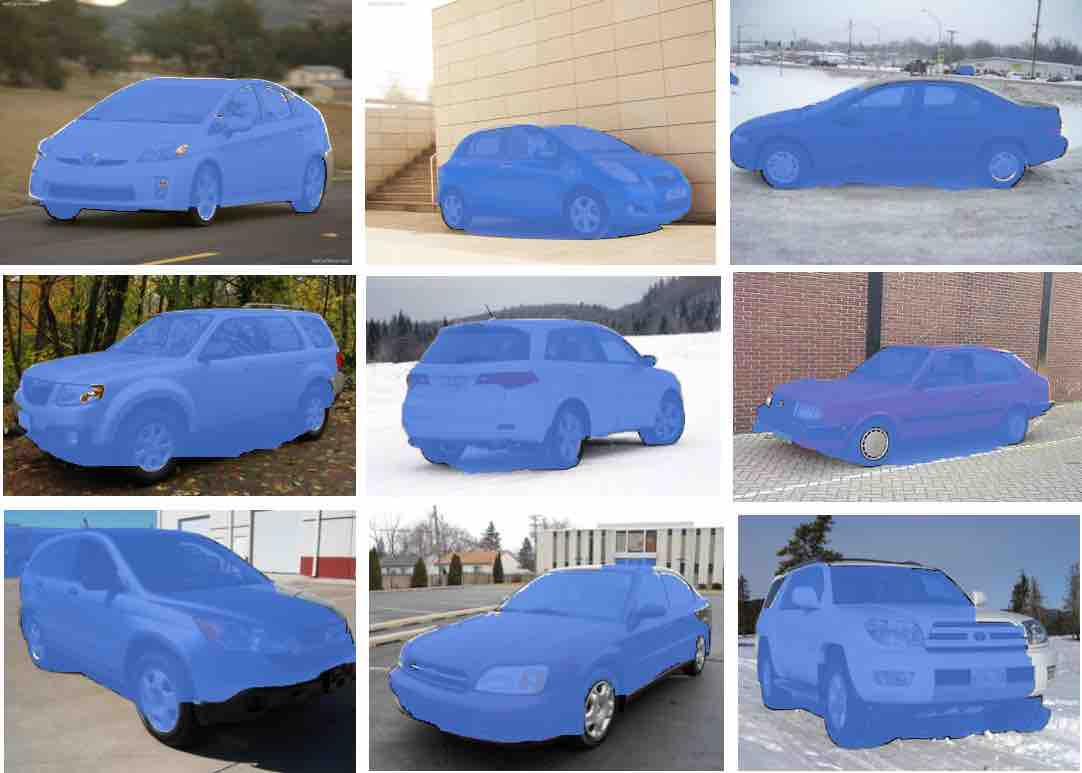}
    \vspace{\figcapmargin}
    \centering\caption*{Object co-segmentation}
  \end{subfigure}
  \caption{
  \textbf{Joint semantic matching and object co-segmentation.} 
  Semantic matching and object co-segmentation are two highly correlated tasks.
  However, existing methods often tackle these two tasks in isolation. 
  In this paper, we exploit the complementary nature of the two tasks and propose a cross-task consistency loss to couple the learning of the two tasks.
  By leveraging cross-task information, our algorithm produces more accurate and consistent results on both tasks.
  }
  \label{fig:teaser}
\end{figure*}

\IEEEraisesectionheading{\section{Introduction}\label{sec:introduction}}
\IEEEPARstart{W}{e} address the problem of jointly aligning and segmenting different object instances of the same category from a collection of images.
These two tasks, known as semantic matching and object co-segmentation (see Figure~\ref{fig:teaser}), are fundamental and active research topics in computer vision with applications ranging from object recognition~\cite{SIFTFlow}, semantic segmentation~\cite{shen}, 3D reconstruction~\cite{mustafa}, content-based image retrieval~\cite{retrieval}, to interactive image editing~\cite{ImageEditing}.
Nevertheless, due to the presence of background clutters, large intra-class appearance variations, and drastic diversities of scales, poses, and viewpoints, both semantic matching and object co-segmentation remain challenging.

\vspace{\paramargin}
{\flushleft {\bf Existing approaches and their drawbacks.}}
Numerous methods have been proposed to address the problems of semantic matching or object co-segmentation.
Earlier approaches for semantic matching rely on hand-engineered features and a geometric alignment model in an energy minimization framework~\cite{Taniai,SIFTFlow,ProposalFlow,DAISY}.
Similarly, conventional object co-segmentation algorithms do not involve feature learning~\cite{Chang15,Jerripothula16,Quan16,Wang17}.
The lack of end-to-end trainable features and inference pipelines often leads to limited performance.
\revised{
In light of this, recent methods leverage trainable descriptors and models for semantic matching~\cite{SCNet,UCN} and object co-segmentation~\cite{Yuan17,li2018deep}. 
While promising results have been reported, training these models~\cite{SCNet,UCN,Yuan17,li2018deep} requires strong supervision in the form of manually labeled ground truth such as keypoint correspondences for semantic matching and object masks for object co-segmentation.
}
However, constructing large-scale and diverse datasets is difficult since the labeling process is often expensive and labor-intensive.
The dependence on manual supervision restricts the scalability of such approaches.

To alleviate this issue, several weakly supervised methods for semantic matching~\cite{End-to-end,rocco2018neighbourhood,kim2018recurrent} and object co-segmentation~\cite{hsu2018co} have been proposed.
While these weakly supervised methods alleviate the need for collecting manually labeled datasets, two issues remain. 
First, existing algorithms for semantic matching~\cite{End-to-end,rocco2018neighbourhood,kim2018recurrent} implicitly enforce the background features from both images to be similar, suffering from the negative impact caused by background clutters.
\revised{
Second, an existing approach for object co-segmentation~\cite{hsu2018co} tends to segment only the most discriminative regions instead of the entire objects.
To address this issue, this method~\cite{hsu2018co} resorts to off-the-shelf object proposal algorithms to guide the learning of object co-segmentation.
However, leveraging the object proposal that is the most consistent with the estimated co-attention map as supervision to guide the model learning can be sub-optimal because the object proposals are generated \emph{independently} for each image and may not accurately highlight the co-occurrent objects in an image collection.
}

\vspace{\paramargin}
{\flushleft {\bf Our work.}}
In this paper, we propose to jointly tackle both semantic matching and object co-segmentation with a two-stream network in an end-to-end trainable fashion.
Our key insights are two-fold.
First, to suppress the effect of background clutters, the predicted object masks by object co-segmentation allow the model to focus on matching the segmented foreground regions while excluding background matching.
Second, the estimated dense correspondence fields by semantic matching provide supervision for enforcing the model to generate geometrically consistent object masks across images.
Therefore, we exploit the interdependency between the two network outputs, \ie the estimated dense correspondence fields and the predicted foreground object masks, by introducing the \emph{cross-network consistency loss}.
Incorporating this loss improves both tasks since it encourages two networks to generate more consistent explanations of the given image pair as shown in Figure~\ref{fig:motivation}.

The proposed training objective requires only weak image-level supervision (\ie image pairs containing common objects).
To facilitate the network training with such weak supervision, for semantic matching we develop \emph{cycle-consistent} losses that make the predicted image transformations more geometrically plausible.
\revised{
For object co-segmentation, motivated by the classic idea of enforcing the foreground histograms of different images to be similar (\ie histogram matching~\cite{rother2006cosegmentation}) and the co-attention loss in~\cite{hsu2018co,hsu2018unsupervised}, we develop a \emph{perceptual contrastive loss} that enhances the foreground appearance similarity between images while enforcing the figure-ground dissimilarity within each image.
} 
As shown in Figure~\ref{fig:motivation}, our model 
using joint learning addresses both tasks simultaneously, producing more accurate and consistent semantic matching and object co-segmentation results.

\setlength{\siximg}{0.16\textwidth}
\begin{figure*}[t]
  \centering
  \begin{subfigure}[b]{\siximg}
    \centering\includegraphics[width=\linewidth]{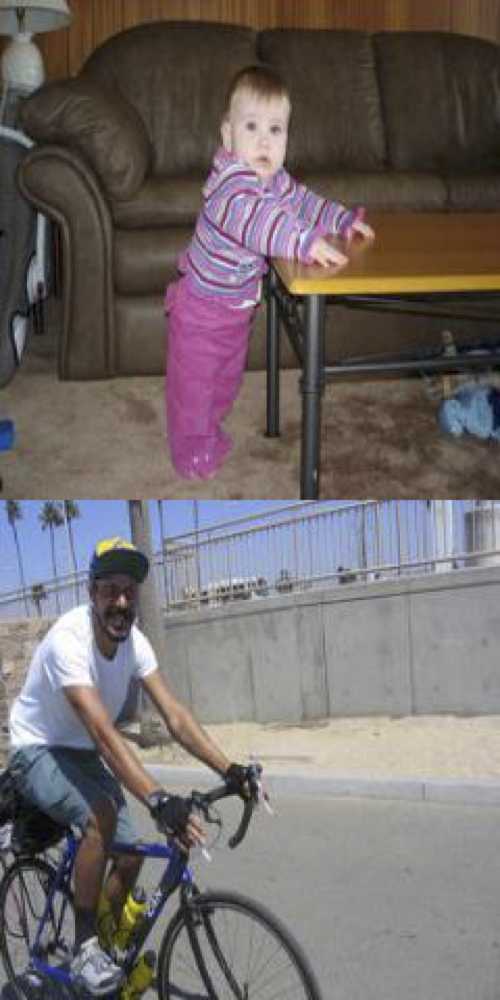}
    \vspace{-5mm}
    \centering\caption*{Input}
  \end{subfigure}
  \begin{subfigure}[b]{\siximg}
    \centering\includegraphics[width=\linewidth]{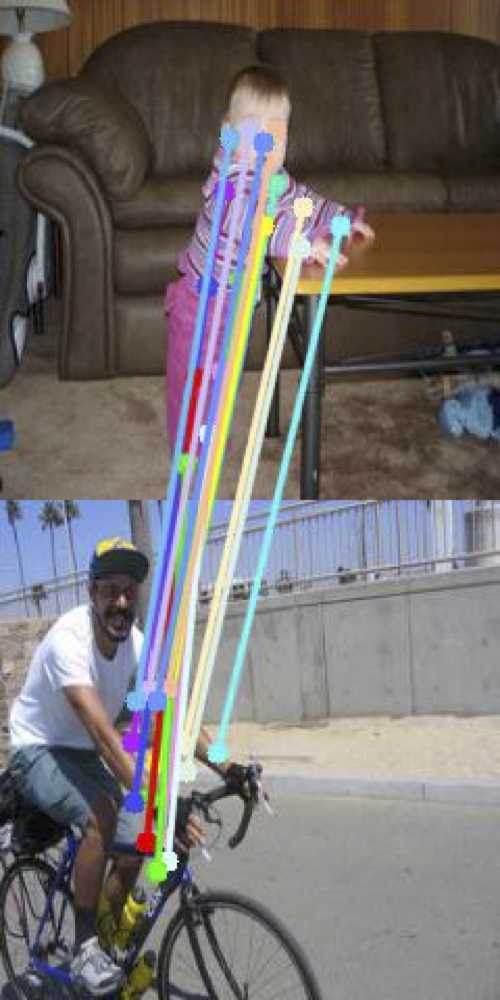}
    \vspace{-5mm}    
    \centering\caption*{Separate learning}
  \end{subfigure}
  \begin{subfigure}[b]{\siximg}
    \centering\includegraphics[width=\linewidth]{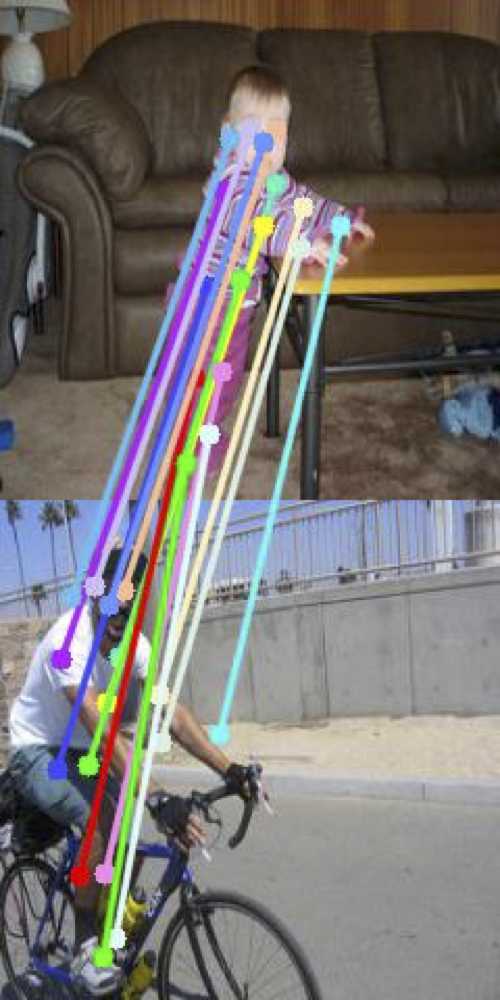}
    \vspace{-5mm}    
    \centering\caption*{Joint learning (Ours)}
  \end{subfigure}
  \begin{subfigure}[b]{\siximg}
    \centering\includegraphics[width=\linewidth]{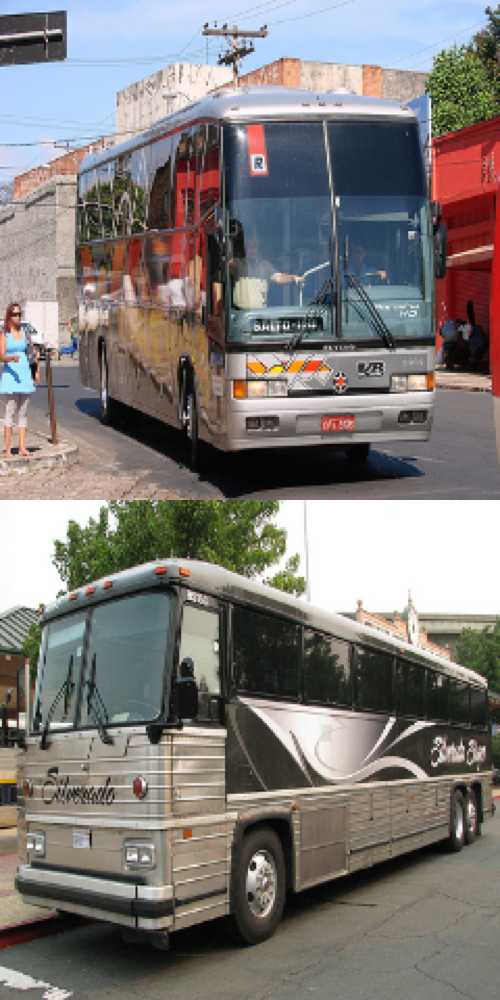}
    \vspace{-5mm}    
    \centering\caption*{Input}
  \end{subfigure}
  \begin{subfigure}[b]{\siximg}
    \centering\includegraphics[width=\linewidth]{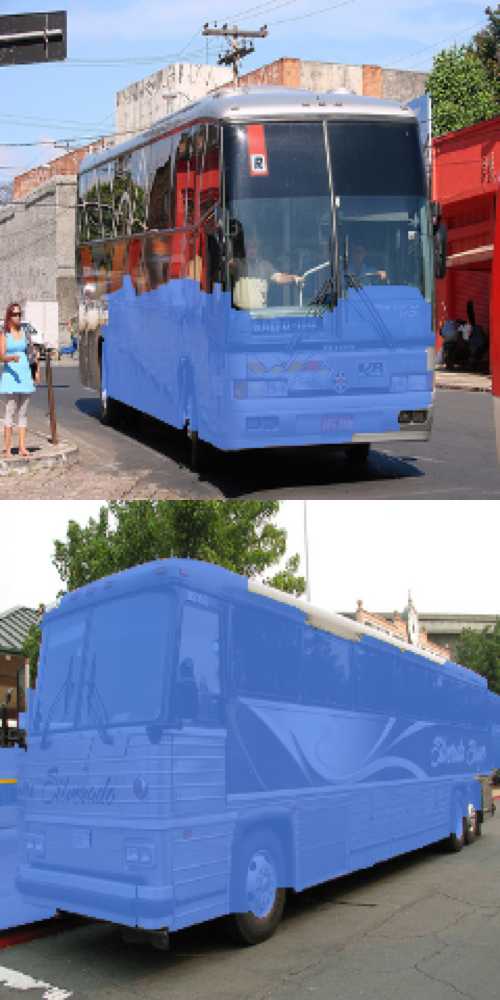}
    \vspace{-5mm}
    \centering\caption*{Separate learning}
  \end{subfigure}
  \begin{subfigure}[b]{\siximg}
    \centering\includegraphics[width=\linewidth]{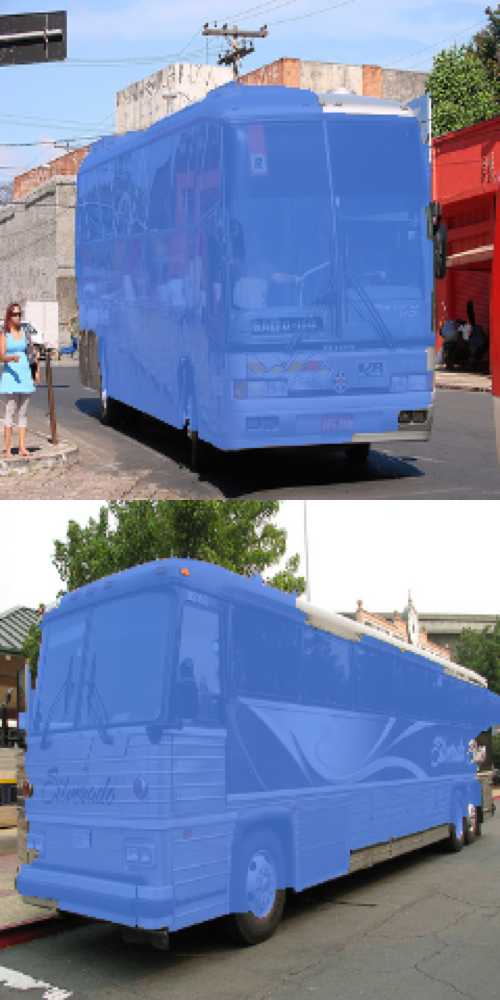}
    \vspace{-5mm}    
    \centering\caption*{Joint learning (Ours)}
  \end{subfigure}
  \caption{
  \textbf{Separate learning vs. joint learning.}
  Addressing semantic matching (\emph{left}) or object co-segmentation (\emph{right}) in isolation often suffers from the effect of background clutters (for semantic matching) or only focuses on segmenting the discriminative parts (for object co-segmentation).
  In this work, we exploit the property that the predicted object masks allow the model to suppress the negative impact due to background clutters while the estimated dense correspondence fields provide supervision for object co-segmentation.
  We couple the learning of both tasks through a cross-network consistency loss and show that joint learning improves the performance of both tasks.
  }
  \label{fig:motivation}
\end{figure*}

\vspace{\paramargin}
{\flushleft {\bf Our contributions.}}
First, we present a weakly-supervised and end-to-end trainable algorithm for joint semantic matching and object co-segmentation.
Second, we propose a cross-network consistency loss that enforces the consistency between the estimated dense correspondence fields and the predicted object masks, resulting in significant performance improvement for both tasks.
\revised{
Third, motivated by the histogram matching idea and the co-attention loss in~\cite{hsu2018co,hsu2018unsupervised}, we develop a perceptual contrastive loss that allows the model to segment the co-occurrent objects from an image collection.
} 
Fourth, we conduct extensive experiments on \revised{five} benchmark datasets including the TSS~\cite{Taniai}, Internet~\cite{rubinstein}, PF-PASCAL~\cite{ProposalFlow}, PF-WILLOW~\cite{ProposalFlow}, \revised{and SPair-$71$k~\cite{min2019spair}.}
Extensive evaluations with existing semantic matching and object co-segmentation methods demonstrate that the proposed algorithm achieves the state-of-the-art performance on both tasks.

\section{Related Work}
\label{sec:RelatedWork}

Semantic matching and object co-segmentation have been extensively studied in the literature.
In this section, we review several topics relevant to our approach.

\vspace{\paramargin}
{\flushleft {\bf Semantic matching.}}
Semantic matching algorithms can be grouped into two categories depending on the adopted feature descriptors: 
(1) \emph{hand-crafted descriptor based} methods~\cite{Taniai,SIFTFlow,ProposalFlow,ProposalFlow-CVPR,DAISY,hu2016progressive,hu2015matching,hsu2015robust,Deformable,AnchorNet,ObjAware,ufer2017deep,GMK,DFF} and 
(2) \emph{trainable descriptor based} methods~\cite{SCNet,UCN,FCSS,FCSS-PAMI,DCTM-PAMI,DCTM,CNNGeo,End-to-end,PARN,ncnet,RTN,WarpNet,A2Net,gaur2017weakly,novotny2018self,rafidirect,laskar2018semi}.
Hand-crafted descriptor based methods often leverage SIFT~\cite{SIFT} or HOG~\cite{HoG} features along with geometric matching models to solve correspondence matching by energy minimization.
However, hand-crafted descriptors are pre-defined and cannot adapt to various tasks.
%
%
%
%
\revised{
Trainable descriptor based methods either use pre-trained~\cite{ufer2017deep} or trainable CNN features for semantic matching~\cite{UCN,SCNet,FCSS,FCSS-PAMI,DCTM,DCTM-PAMI,rafidirect,laskar2018semi}. 
While these methods demonstrate significant performance gain over those using hand-crafted features, they require manual correspondence annotations for training~\cite{ufer2017deep,UCN,SCNet,FCSS,FCSS-PAMI,DCTM,DCTM-PAMI,rafidirect,laskar2018semi}.
} 

%
\revised{
Recently, several weakly supervised approaches~\cite{AnchorNet,End-to-end,PARN,ncnet,RTN,WarpNet,gaur2017weakly,CNNGeo,A2Net,novotny2018self} have been proposed to relax the dependence on keypoint-based supervision.
}
%
The AnchorNet~\cite{AnchorNet} learns a set of filters with geometrically consistent responses across different object instances to establish inter-image correspondences.
The AnchorNet model, however, is not end-to-end trainable due to the use of the hand-engineered alignment model.
The WarpNet~\cite{WarpNet} considers fine-grained image matching with small-scale and pose variations via aligning objects across images through known deformation.
However, the application domain is relatively ideal since the objects are located in the image centers with limited translations, scale variations, and background clutters.
Gaur~\etal~\cite{gaur2017weakly} propose an optimization algorithm that learns a latent space to cluster semantically related object parts.
%
%
%
\revised{
While this method provides geometric invariance to some degree, their approach cannot handle affine transformations across images that frequently occur in the context of semantic matching.
To address this issue, several methods have been proposed.
A number of methods learn to predict geometric transformations for image pairs in a self-supervised fashion~\cite{CNNGeo,novotny2018self,A2Net}.
Rocco~\etal~\cite{End-to-end} present a weakly supervised semantic matching network using a differentiable soft inlier scoring module.
}
The PARN~\cite{PARN} estimates locally-varying affine transformation fields across semantically similar images in a coarse-to-fine manner.
Motivated by the procedure of non-rigid image registration between an image pair, the RTNs~\cite{RTN} use recurrent networks to progressively compute dense correspondences between two images.
In addition to estimating geometric transformations between an image pair~\cite{End-to-end,PARN,RTN}, another line of research focuses on establishing dense per-pixel correspondences without using any geometric models~\cite{ncnet}.
Rocco~\etal~\cite{ncnet} establish dense correspondences by analyzing the neighborhood pattern in the 4D space using 4D convolutional layers.
Similar to these methods~\cite{AnchorNet,End-to-end,PARN,RTN,ncnet}, our model also does not require ground-truth correspondences from training images.
Our method differs from these methods in two aspects.
First, our method explicitly models foreground masks from object co-segmentation, and thereby effectively reduces the negative impacts of background clutters on semantic matching.
Second, we enforce the cycle consistency constraints on the predicted geometric transformations through two losses, resulting in more accurate and consistent matching results.

\vspace{\paramargin}
{\flushleft {\bf Object co-segmentation.}}
Object co-segmentation algorithms can be categorized into two groups: (1) {\em graph-based}~\cite{Chang15,Jerripothula16,Quan16,Wang17} and (2) {\em clustering-based}~\cite{Joulin12,Lee15,Tao17} approaches. 
Graph-based methods first construct a graph to encode the relationships between object instances from different images and formulate object co-segmentation as a labeling problem. 
Clustering-based methods, on the other hand, assume that common objects share similar appearances and achieve co-segmentation by finding tight clusters. 
Existing methods in both groups use hand-crafted features such as SIFT~\cite{SIFT}, HOG~\cite{HoG}, or texton~\cite{shotton2009textonboost} for describing a set of object candidates extracted from super-pixels or region-based proposals.
Recently, learning based methods~\cite{Yuan17,li2018deep,hsu2018co,chen2018semantic} have been developed for object co-segmentation.
While significant improvement has been shown, these methods~\cite{Yuan17,li2018deep,chen2018semantic} require costly foreground masks for training and are not applicable to unseen object categories.
%
%
%
%
%
%
%
%
\revised{
To address the absence of manual supervision, Hsu~\etal~\cite{hsu2018co} propose the co-attention loss that allows the model to segment the co-occurrent objects from an image collection.
However, using the co-attention loss alone might lead to sup-optimal performance because models may only segment the discriminative regions in each image.
To tackle the issue, Hsu~\etal~\cite{hsu2018co} further develop a mask loss by taking single-image objectness into account and leveraging off-the-shelf object proposal algorithms to regularize the learning of object co-segmentation.
Similar to Hsu~\etal~\cite{hsu2018co}, our method does not require manually labeled object masks and can segment objects of unseen categories.
Our algorithm differs from them~\cite{hsu2018co} in three aspects.
First, our method does not rely on the detected object proposals as supervision for each image to guide the network training.
Second, the estimated geometric transformations from semantic matching provide supervision for object co-segmentation by enforcing the predicted object masks from the given image pair to be geometrically consistent.
Third, our model further takes into account the correlation map when performing object co-segmentation.
}

\vspace{\paramargin}
{\flushleft {\bf Joint semantic matching and object co-segmentation.}}
Several methods explore joint semantic correspondence and object co-segmentation.
Rubinstein~\etal~\cite{rubinstein} carry out object co-segmentation by exploiting the dense correspondence fields derived by the SIFT flow~\cite{SIFTFlow}.
Taniai~\etal~\cite{Taniai} develop a hierarchical Markov random field (MRF) model for joint dense matching and object co-segmentation.
However, these methods employ hand-crafted descriptors.
Motivated by these methods~\cite{rubinstein,Taniai}, our approach leverages the complementary nature of the two tasks and develop cross-network loss functions that couple the two networks during optimization for improving the performance of the individual tasks.
After optimization, the learned semantic matching and the object co-segmentation models can be applied jointly or independently.

\vspace{\paramargin}
{\flushleft {\bf Meta-supervision via coupled network training.}} 
%
%
\revised{
Enforcing consistency across different network outputs has been used in several vision applications including image translations~\cite{zhu2017unpaired,DRIT,DRIT++,MUNIT}, depth and ego-motion~\cite{zhou2017unsupervised,DFNet}, and shape reconstruction~\cite{tulsiani2018multi}.
}
In this work, we design a cross-network objective function to make the semantic matching and object co-segmentation networks complementary to each other, and demonstrate that coupled training of the two heterogeneous networks leads to significant performance improvement on both tasks.

\vspace{\paramargin}
{\flushleft {\bf Cycle consistency.}}
Cycle consistency constraints have been used to regularize network training for numerous vision tasks.
\revised{
Exploiting cycle-consistency properties along the time axis allows us to align different videos~\cite{dwibedi2019temporal}, establish visual correspondences in unlabeled videos~\cite{wang2019learning}, and better interpolate video frames~\cite{liu2019deep}.
In the context of visual question answering, enforcing cycle consistency between low information modality (\eg an answer) and high information modality (\eg a question) can alleviate the requirement of paired training data~\cite{shah2019cycle}.
In motion analysis, enforcing forward-backward consistency constraints has been shown effective for detecting occlusion while learning optical flow~\cite{UnFlow,DFNet} or depth prediction~\cite{wong2019bilateral}.
}
In image-to-image translation, enforcing cycle consistency allows the model to learn the mappings between domains without paired data~\cite{CycleGan,DRIT,MUNIT}.
In unsupervised domain adaptation, exploiting cross-domain invariance in the label space results in more consistent task predictions for unlabeled images of different domains~\cite{CrDoCo}.
%
%
Similar to these methods, the idea of cycle consistency is also extensively applied to semantic matching.
The FlowWeb~\cite{FlowWeb} enforces cycle consistency constraints to establish globally-consistent dense correspondences.
Zhou~\etal~\cite{Multi-match} address multi-image matching by jointly learning feature matching and enforcing cycle consistency. 
However, these methods~\cite{Multi-match,FlowWeb} adopt hand-crafted descriptors, which may not adapt to unseen object category given for matching.
While trainable descriptor based methods~\cite{3D-cycle} are proposed to alleviate this limitation, this method~\cite{3D-cycle} utilizes an additional 3D CAD model to form a cross-instance loop between synthetic and real images for establishing dense correspondences.
Namely, this method~\cite{3D-cycle} requires four images for computing the cycle consistency loss.
In contrast, our method does not need any additional data to guide network training.
Experimental results show that with the two developed cycle consistency losses, our model produces consistent matching results, resulting in significant performance gain.

\begin{figure*}[t]
  \begin{center}
  \includegraphics[width=.9\linewidth]{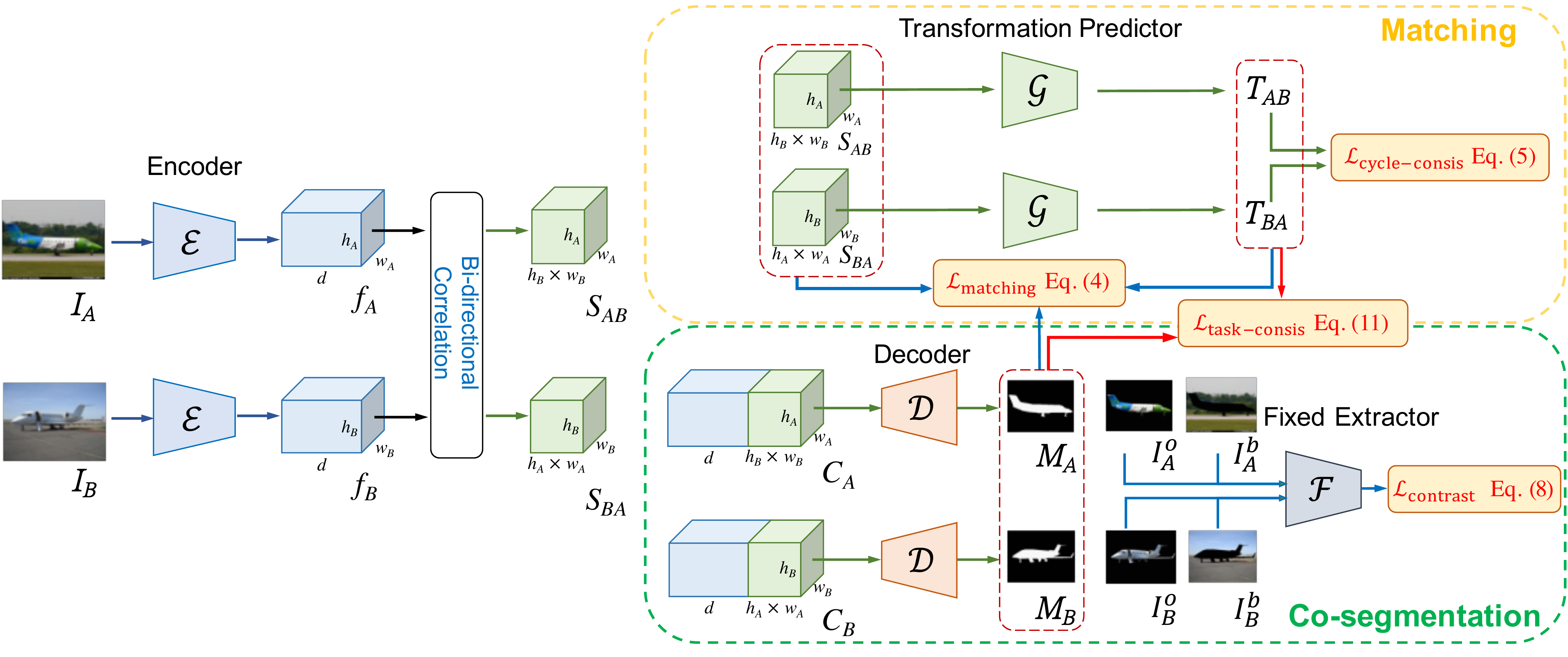}
  \caption{\textbf{Overview of the proposed model.} 
  Our model is a two-stream network: (\emph{top}) semantic matching network and (\emph{bottom}) object co-segmentation network.
  Our model consists of four main CNN sub-networks: an encoder $\cal{E}$ (for extracting features from the input images), a transformation predictor $\cal{G}$ (for estimating the geometric transformation between an input image pair), a decoder $\cal{D}$ (for producing object masks), and an ImageNet-pretrained and fixed ResNet-$50$ feature extractor $\cal{F}$ (for computing the perceptual contrastive loss).
  The model training is driven by four loss functions, including the foreground-guided matching loss $\mathcal{L}_\mathrm{matching}$, forward-backward consistency loss $\mathcal{L}_\mathrm{cycle-consis}$, perceptual contrastive loss $\mathcal{L}_\mathrm{contrast}$, and cross-network consistency loss $\mathcal{L}_\mathrm{task-consis}$.
  }
  \label{Fig:Model}
  \end{center}
\end{figure*}

\section{Proposed Algorithm}

In this section, we first provide an overview of the proposed approach.
We then describe the objective functions for joint semantic matching and object co-segmentation followed by the implementation details.

\subsection{Algorithmic overview}

Given a set of $N$ images $\mathcal{I} = \{I_i\}_{i=1}^N$ containing objects of a specific category, our goal is to learn a model that can determine the geometric correspondences between the input image pairs while segmenting the common objects in $\mathcal{I}$ \emph{without knowing the object class a priori}.
Our formulation for joint semantic matching and object co-segmentation is \emph{weakly-supervised} since it requires only weak image-level supervision in the form of training image pairs containing objects of one particular class. 
No ground-truth keypoint correspondences and object masks are used in the training stage.

\vspace{\paramargin}
{\flushleft {\bf Network.}}
As shown in Figure~\ref{Fig:Model}, our model is composed of four CNN sub-networks: an {\em encoder} $\cal{E}$, a {\em transformation predictor} $\cal{G}$, a {\em decoder} $\cal{D}$, and an {\em ImageNet-pretrained ResNet-$50$ feature extractor} $\cal{F}$.
Given an input image pair, we first use the encoder $\cal{E}$ to encode the content of each image.
We then apply a correlation layer for computing matching scores for every pair of features from two images. 
The correlation layer has been extensively applied in the other context, including optical flow~\cite{FlowNet}, stereo~\cite{kendall2017end,huang2018deepmvs}, and video object segmentation~\cite{hu2018videomatch,voigtlaender2019feelvos}.
%
Here, taking two tensors of matching scores as inputs, we apply a transformation predictor $\cal{G}$ to estimate the geometric transformation that aligns the two images.
To generate object masks, we use the fully convolutional network decoder $\cal{D}$ for object co-segmentation.
%
To capture the co-occurrence information, we concatenate the encoded image features with the correlation maps.
Our decoder $\cal{D}$ then takes the concatenated features as inputs to generate object segmentation masks.
The use of both feature maps and correlation maps for object co-segmentation has been proposed in \cite{li2018deep}.
%
%
However, the method in \cite{li2018deep} requires manually annotated object masks for training. 
To enable the decoder $\cal{D}$ to segment the co-occurrent objects from the given image pair \emph{without} supervision from ground-truth object masks, we leverage the perceptual contrastive loss $\mathcal{L}_\mathrm{contrast}$ using the ImageNet-pretrained ResNet-$50$~\cite{ResNet} feature extractor $\cal{F}$ to enforce the appearance similarity between the segmented foreground across images and the dissimilarity between the masked foreground and background within each image.

\vspace{\paramargin}
{\flushleft {\bf Training losses.}}
Our training objective consists of five loss functions.
First, the foreground-guided matching loss $\mathcal{L}_\mathrm{matching}$ minimizes the distance between corresponding features based on the estimated geometric transformation.
Unlike existing feature learning methods for semantic matching~\cite{End-to-end,CNNGeo,UCN}, our model explicitly takes the predicted object masks into account to suppress the negative impacts caused by background clutters.
Second, the cross-network consistency loss $\mathcal{L}_\mathrm{task-consis}$ penalizes the inconsistency of the predicted object masks of an input image pair and the estimated geometric transformations between that pair.
Such a cross-network loss couples the networks during training and provides supervisory signals for both semantic matching and object co-segmentation.
Third, the perceptual contrastive loss $\mathcal{L}_\mathrm{contrast}$ guides the decoder $\cal{D}$ to produce object co-segments with higher inter-image foreground similarity and large intra-image figure-ground separation.
Fourth, both the forward-backward consistency loss $\mathcal{L}_\mathrm{cycle-consis}$ and transitivity consistency loss $\mathcal{L}_\mathrm{trans-consis}$ (applied to three input images at a time) regularize the network training by enforcing the predicted geometric transformations to be consistent across multiple images.
Specifically, the training objective $\mathcal{L}$ is defined by
\begin{equation}
  \begin{split}
  \mathcal{L} & = \mathcal{L}_\mathrm{matching} \\ 
  & + \lambda_\mathrm{cycle}\cdot\mathcal{L}_\mathrm{cycle-consis} + \lambda_\mathrm{trans}\cdot\mathcal{L}_\mathrm{trans-consis} \\ 
  & + \lambda_\mathrm{contrast}\cdot\mathcal{L}_\mathrm{contrast} + \lambda_\mathrm{task}\cdot\mathcal{L}_\mathrm{task-consis},
  \label{eq:FullObj}
  \end{split}
\end{equation}
where $\lambda_\mathrm{cycle}$, $\lambda_\mathrm{trans}$, $\lambda_\mathrm{contrast}$, and $\lambda_\mathrm{task}$ are the hyper-parameters used to control the relative importance of the respective loss terms.
%

\subsection{Semantic matching}

\vspace{\paramargin}
{\flushleft\textbf{Foreground-guided matching loss} $\mathcal{L}_\mathrm{matching}$.}
%
%
\revised{
Given an image pair $(I_A,I_B)$, the encoder $\mathcal{E}$ represents the images with the feature maps $f_{A}\in\mathbb{R}^{h_A\times w_A\times d}$ and $f_{B}\in\mathbb{R}^{h_B\times w_B\times d}$, where $h_A \times w_A$ is the spatial size of $f_A$, $h_B \times w_B$ is the spatial size of $f_B$, and $d$ is the number of channels.
}
We apply a correlation layer to $f_{A}$ and $f_{B}$, and obtain the correlation map $S_{AB}\in\mathbb{R}^{h_A\times w_A\times h_B\times w_B}$, where the element $S_{AB}(i,j,s,t) = S_{AB}(\mathbf{p},\mathbf{q})$ records the normalized inner product between the feature vectors extracted at two locations $\mathbf{p} = [i, j]^\top$ in $f_{A}$ and $\mathbf{q} = [s, t]^\top$ in $f_{B}$.
The correlation map $S_{AB}$ can be reshaped to a three-dimensional tensor with dimensions $h_A$, $w_A$, and $(h_B \times w_B)$, \ie $S_{AB} \in \bbR^{h_A \times w_A \times (h_B \times w_B)}$.
As such, $S_{AB}$ can be interpreted as a dense $h_A \times w_A$ grid with $(h_B \times w_B)$-dimensional local features.
With the reshaped $S_{AB}$, we use the transformation predictor $\cal{G}$~\cite{CNNGeo} to estimate a geometric transformation $T_{AB}$ which warps $I_A$ to $\tilde{I}_A$ so that $\tilde{I}_A$ and $I_B$ can be well aligned.

With the geometric transformation $T_{AB}$, we can identify and remove geometrically inconsistent correspondences.
Consider a correspondence with the endpoints $(\mathbf{p} \in \mathcal{P}_A, \mathbf{q} \in \mathcal{P}_B)$, where $\mathcal{P}_A$ and $\mathcal{P}_B$ are the domains of all spatial coordinates of $f_A$ and $f_B$, respectively.
We refer the distance $\| T_{AB}(\mathbf{p}) - \mathbf{q} \|$ as the {\em projection error} of this correspondence with respect to transformation $T_{AB}$.
Following Rocco~\etal~\cite{End-to-end}, we introduce a correspondence mask $m_A \in \bbR^{h_A \times w_A \times (h_B \times w_B)}$ to determine if the correspondences are geometrically consistent with transformation $T_{AB}$.
Specifically, $m_A$ is of the form
\begin{equation}
  \begin{split}
  m_{A}{(\mathbf{p},\mathbf{q})} = &
  \begin{cases}
    1, & \text{if $\|T_{AB}(\mathbf{p})-\mathbf{q}\| \leq \varphi$},\\
    0, & \text{otherwise},
  \end{cases}\mbox{,}
  \\
  & \mbox{for $\mathbf{p}\in\mathcal{P}_A$ and $\mathbf{q}\in\mathcal{P}_B$,}
  \end{split}
  \label{eq:CorrMask}
  \vspace{-3mm}
\end{equation} 
where $\varphi$ is a pre-defined threshold. 
In \revised{Eq.}~(\ref{eq:CorrMask}), a correspondence is considered geometrically consistent with transformation $T_{AB}$ if its projection error is not larger than the threshold $\varphi$.
Empirically, we set the threshold $\varphi$ to $1$ in all experiments.

For the correspondence with the endpoints $(\mathbf{p}\in\mathcal{P}_A, \mathbf{q}\in\mathcal{P}_B)$, the correlation map $S_{AB}(\mathbf{p},\mathbf{q})$ and the correspondence mask $m_{A}(\mathbf{p},\mathbf{q})$ capture its appearance and geometric consensus, respectively. 
When focusing on point $\mathbf{p}\in\mathcal{P}_A$, we compute the matching score of location $\mathbf{p}$ by
\begin{equation}
  s_{A}{(\mathbf{p})} = \sum_{\mathbf{q}\in\mathcal{P}_{B}}^{}m_{A}{(\mathbf{p},\mathbf{q})}\cdot {S}_{AB} (\mathbf{p},\mathbf{q}).
  \label{eq:MatchScore}
\end{equation}

%
%
%
%
Semantic matching often suffers from false positives caused by background clutters and false negatives caused by large intra-class variations.
In this work, we exploit object co-segmentation, where object-level similarity complements patch-level similarity in semantic matching, to address the aforementioned issues.
Specifically, we consider the object mask $M_A$ estimated by the decoder $\mathcal{D}$ for object co-segmentation, and resize it to resolution $h_A \times w_A$ for guiding the matching loss $\mathcal{L}_\mathrm{matching}$. 
Our foreground-guided matching loss is formulated as
%
\begin{equation}
  \begin{split}
  & \mathcal{L}_\mathrm{matching}(I_A,I_B,M_A,M_B;\mathcal{E},\mathcal{G},\mathcal{D}) \\
  & = - \bigg(\sum_{\mathbf{p}\in\mathcal{P}_{A}}^{} s_A{(\mathbf{p})}\cdot M_{A}{(\mathbf{p})} + \sum_{\mathbf{q}\in\mathcal{P}_{B}}^{}s_{B}{(\mathbf{q})}\cdot M_B{(\mathbf{q})}\bigg),
  \end{split}
  \label{eq:MatchLoss}
\end{equation}
where $s_{B}$ and $M_B$ are similarly defined as $s_{A}$ and $M_A$, respectively.
The negative sign in \revised{Eq.}~(\ref{eq:MatchLoss}) indicates that maximizing the matching score is equivalent to minimizing the foreground-guided matching loss $\mathcal{L}_\mathrm{matching}$. 
The loss $\mathcal{L}_\mathrm{matching}$ encourages the transformation predictor $\cal{G}$ to generate transformations $T_{AB}$ and $T_{BA}$ with which the corresponding foreground features across the two images are as similar as possible.

\vspace{\paramargin}
{\flushleft {\bf Cycle consistency.}}
For an image pair $I_A$ and $I_B$, the transformation predictor $\mathcal{G}$ estimates a geometric transformation $T_{AB}$ which can warp $I_A$ to $\tilde{I}_A$ such that $\tilde{I}_A$ aligns $I_B$ well.
However, the large capacity of the transformation predictor $\mathcal{G}$ often leads to situations where various transformations can warp $I_A$ to $\tilde{I}_A$ such that $\tilde{I}_A$ aligns $I_B$ well.
Namely, multiple points on $I_A$ can match well a single point on $I_B$.
These cases imply that using the foreground-guided matching loss $\mathcal{L}_\mathrm{matching}$ alone is insufficient to reliably train the transformation predictor $\mathcal{G}$ under the weakly supervised setting since no ground-truth correspondences are available to guide the search of the correct geometric transformations.
To address this issue, we simultaneously estimate bi-directional geometric transformations $T_{AB}$ and $T_{BA}$ and explicitly enforce the predicted transformations to be geometrically plausible and consistent across multiple images.
As such, exploiting the cycle consistency constraints greatly reduces the feasible space of transformations and can serve as a regularization term in training the transformation predictor $\mathcal{G}$, avoiding the degenerate solutions.
We develop the following two loss functions to exploit such geometric consistency constraints.

%

\vspace{\paramargin}
{\flushleft\textbf{1) Forward-backward consistency loss $\mathcal{L}_\mathrm{cycle-consis}$.}}
Consider the correlation maps $S_{AB}$ and $S_{BA}$ generated from images $I_A$ and $I_B$, our transformation predictor $\mathcal{G}$ predicts two transformations: (1) $T_{AB}$: mapping points from image $I_A$ to $I_B$ and (2) $T_{BA}$: mapping points from image $I_B$ to $I_A$.
The forward-backward consistency enforces the property $T_{BA}(T_{AB}(\mathbf{p}))\approx \mathbf{p}$ for any $\mathbf{p}\in\mathcal{P}_{A}$.
Similarly, the consistency can also be applied in the reverse direction by enforcing $T_{AB}(T_{BA}(\mathbf{q}))\approx \mathbf{q}$ for any $\mathbf{q}\in\mathcal{P}_{B}$.
We introduce the forward-backward consistency loss $\mathcal{L}_\mathrm{cycle-consis}$ as 
%
%
\begin{equation}
  \begin{aligned}
    \mathcal{L}_\mathrm{cycle-consis}&(I_A,I_B;\mathcal{E},\mathcal{G})\\
    = &~ \frac{1}{\|\mathcal{P}_{A}\|}\sum_{\mathbf{p}\in\mathcal{P}_{A}}^{} \|T_{BA}(T_{AB}(\mathbf{p}))- \mathbf{p}\| \\
    + &~ \frac{1}{\|\mathcal{P}_{B}\|}\sum_{\mathbf{q}\in\mathcal{P}_{B}} \|T_{AB}(T_{BA}(\mathbf{q})) - \mathbf{q}\|,
  \end{aligned}
  \label{eq:cycle-consis}
\end{equation}
where $\|\mathcal{P}_{A}\|$ is the number of pixel coordinates in $\mathcal{P}_A$ and $\|T_{BA}(T_{AB}(\mathbf{p})) - \mathbf{p}\|$ is the re-projection error between coordinate $\mathbf{p}$ and the re-projected coordinate $T_{BA}(T_{AB}(\mathbf{\mathbf{p}}))$.

\vspace{\paramargin}
{\flushleft\textbf{2) Transitivity consistency loss $\mathcal{L}_\mathrm{trans-consis}$.}}
The idea of forward-backward consistency between an image pair can be extended to the \emph{transitivity} consistency across multiple images, \eg three images.
Considering the case of three images $I_A$, $I_B$, and $I_C$, we first estimate three geometric transformations $T_{AB}$, $T_{BC}$, and $T_{CA}$.
Transitivity consistency in this case states that for any coordinate $\mathbf{p} \in \mathcal{P}_A$, the property $T_{CA}(T_{BC}(T_{AB}(\mathbf{p}))) \approx \mathbf{p}$ holds.
Thus, we introduce the transitivity consistency loss $\mathcal{L}_\mathrm{trans-consis}$ as
%
\begin{equation}
  \begin{aligned}
    \mathcal{L}_\mathrm{trans-consis}&(I_A,I_B,I_C;\mathcal{E},\mathcal{G}) \\
    = &~ \frac{1}{\|\mathcal{P}_{A}\|}\sum_{\mathbf{p}\in\mathcal{P}_{A}}^{}\|T_{CA}(T_{BC}(T_{AB}(\mathbf{p}))) - \mathbf{p}\|. \\
  \end{aligned}
  \label{eq:trans-consis}
\end{equation}

\vspace{\paramargin}
{\flushleft {\bf Avoiding degenerate solutions.}}
We note that degenerate solutions may exist for the optimization problem based on the foreground-guided matching loss 
and cycle-consistent losses. 
That is, the correspondence masks $m_A$ and $m_B$ are made zero everywhere, and the transformation predictor $\mathcal{G}$ always predicts identity transformation.
To address this issue, we use the transformation predictor $\mathcal{G}$~\cite{CNNGeo} pre-trained on a large-scale synthetic dataset.
The pre-trained model provides good initialization, thus alleviating the issue of rendering degenerate solutions.
%

\subsection{Object co-segmentation}

To segment the common objects in a pair of images $(I_A, I_B)$ and enhance the performance based on semantic matching, our network estimates object co-segmentation masks by minimizing the perceptual contrastive loss $\mathcal{L}_\mathrm{contrast}$ and the cross-network consistency loss $\mathcal{L}_\mathrm{task-consis}$.

\vspace{\paramargin}
{\flushleft\textbf{Perceptual contrastive loss} $\mathcal{L}_\mathrm{contrast}$.}
Given an image pair $(I_A, I_B)$, the respective feature maps $f_A$ and $f_B$, and the reshaped correlation maps $S_{AB}$ and $S_{BA}$, to capture the inter-image co-occurrence information, we first generate the concatenated feature representations $C_A \in \mathbb{R}^{h_A \times w_A \times (h_B \times w_B + d)}$ by concatenating $f_A$ with $S_{AB}$ for $I_A$, and $C_B \in \mathbb{R}^{h_B \times w_B \times (h_A \times w_A + d)}$ by concatenating $f_B$ with $S_{BA}$ for $I_B$.
As shown in Figure~\ref{Fig:Model}, the decoder $\cal{D}$ takes the concatenated feature maps $C_A$ and $C_B$ as inputs and produces object masks $M_A \in \bbR^{H_A \times W_A}$ and $M_B \in \bbR^{H_B \times W_B}$ for input images $I_A$ and $I_B$, respectively.
%
%
\revised{
To facilitate the decoder $\cal{D}$ segmenting the co-occurrent objects, we introduce the perceptual contrastive loss $\mathcal{L}_\mathrm{contrast}$ that enhances the quality of the object co-segmentation masks produced by the decoder $\cal{D}$ based on two criteria: (1) high foreground object similarity \emph{across} the images~\cite{rother2006cosegmentation,hsu2018co,hsu2018unsupervised} and (2) high foreground-background discrepancy \emph{within} each image~\cite{hsu2018co,hsu2018unsupervised}.
} 

%
%
%
\revised{
Following the scheme in~\cite{hsu2018co,hsu2018unsupervised}, we first generate the \emph{object image} $I_i^o$ and the \emph{background image} $I_i^b$ for each image $I_i$ by
\begin{equation}
  I_i^o = M_i\otimes I_i \; \textmd{ and } \; I_i^b = (1-M_i)\otimes I_i \textmd{ for } i \in \{A,B\},
\end{equation} 
where $\otimes$ denotes the pixel-wise multiplication between the two operands.
} 
We then apply an ImageNet-pretrained ResNet-$50$~\cite{ResNet} network $\cal{F}$ to $I_i^o$ and $I_i^b$ and extract their semantic feature vectors $\mathcal{F}(I_i^o)$ and $\mathcal{F}(I_i^b)$, respectively.
The perceptual contrastive loss $\mathcal{L}_\mathrm{contrast}$ is defined by
\begin{equation}
  \mathcal{L}_\mathrm{contrast}(I_A,I_B;\mathcal{E},\mathcal{D},\mathcal{F}) = d_{AB}^+ + d_{AB}^-,
  \label{eq:contrast}
\end{equation}
where the two aforementioned criteria are respectively imposed on $d_{AB}^+$ and $d_{AB}^-$:
\begin{equation}
  d_{AB}^+ = \frac{1}{c}\|\mathcal{F}(I_A^o)-\mathcal{F}(I_B^o)\|^2 \textmd{ and}
  \label{eq:pos-d-AB}
\end{equation}
\begin{equation}\scriptsize
  d_{AB}^- = \max\Bigg(0, m - \frac{1}{2c}\bigg(\|\mathcal{F}(I_A^o)-\mathcal{F}(I_A^b)\|^2 + \|\mathcal{F}(I_B^o)-\mathcal{F}(I_B^b)\|^2\bigg)\Bigg).
  \label{eq:neg-d-AB}
\end{equation}
%

In \revised{Eq.}~\eqref{eq:neg-d-AB}, the constant $c$ set to be $2,048$ is the dimension of the semantic features produced by $\cal{F}$~\cite{ResNet}, and the margin $m$ set to be $2$ is the cutoff threshold.

\begin{figure}[t]
  \begin{center}
  \includegraphics[width=\linewidth]{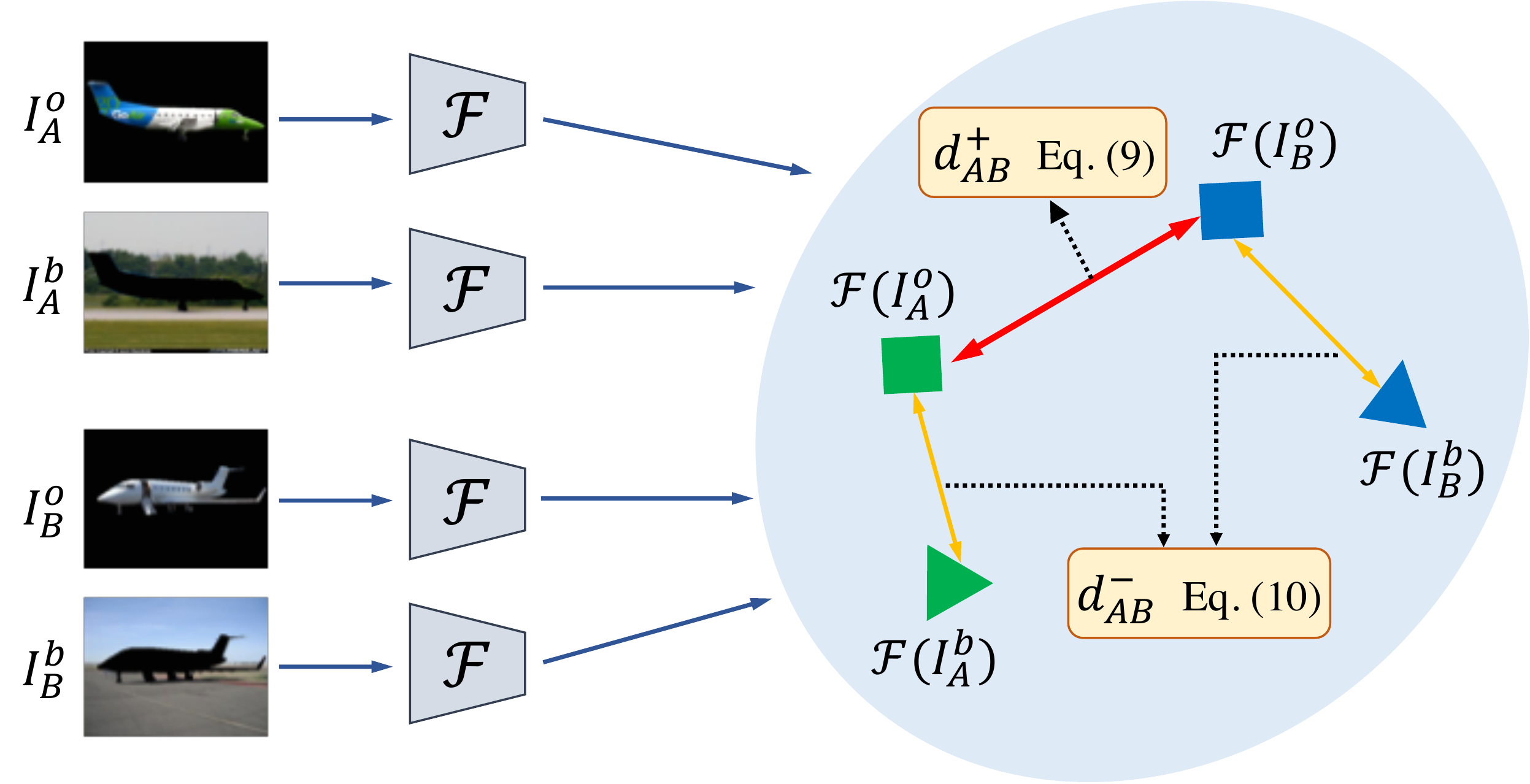}
  \caption{\textbf{Illustration of the perceptual contrastive loss $\mathcal{L}_\mathrm{contrast}$.} 
  The proposed perceptual contrastive loss is developed based on two criteria:
  %
  (1) low inter-image foreground object distinctness ($d_{AB}^+$) and
  (2) high intra-image foreground-background discrepancy ($d_{AB}^-$).
  }
  \label{fig:contrast-loss}
  \end{center}
\end{figure}

As shown in Figure~\ref{fig:contrast-loss}, minimizing the perceptual contrastive loss $\mathcal{L}_\mathrm{contrast}$ in \revised{Eq.}~(\ref{eq:contrast}) entails minimizing the inter-image foreground object distinctness in \revised{Eq.}~(\ref{eq:pos-d-AB}) while maximizing the intra-image foreground-background discrepancy in \revised{Eq.}~(\ref{eq:neg-d-AB}).
We note that minimizing $d_{AB}^+$ in \revised{Eq.}~(\ref{eq:pos-d-AB}) is equivalent to trainable matching foreground histograms~\cite{rother2006cosegmentation} between a pair of images.
%
%
\revised{
While both the proposed perceptual contrastive loss and the co-attention loss in~\cite{hsu2018co} share similar spirits, \ie aiming at minimizing the inter-image foreground object distinctness (Eq. (9)) and maximizing the intra-image foreground-background discrepancy (Eq. (10)), there is one major difference.
Since the object image $I_i^o$ and background image $I_i^b$ for each image $I_i$ are inherently \emph{different}, minimizing the co-attention loss~\cite{hsu2018co} may be dominated by maximizing the intra-image foreground-background discrepancy (\ie $d_{ij}^{-}$), which might lead to sub-optimal performance.
The formulation of our perceptual contrastive loss, on the other hand, is capable of avoiding this degenerate solution by introducing a cutoff threshold $m$ in $d_{AB}^{-}$ (Eq. (10)).
When the object images and background images are dissimilar to some extent, \ie $\frac{1}{2c}\bigg(\|\mathcal{F}(I_A^o)-\mathcal{F}(I_A^b)\|^2 + \|\mathcal{F}(I_B^o)-\mathcal{F}(I_B^b)\|^2\bigg) > m$, our model will stop maximizing the intra-image foreground-background discrepancy, and focuses only on minimizing the inter-image foreground object distinctness.
}

\vspace{\paramargin}
{\flushleft\textbf{Cross-network consistency loss} $\mathcal{L}_\mathrm{task-consis}$.}
Using the perceptual contrastive loss $\mathcal{L}_\mathrm{contrast}$ alone for object co-segmentation may generate object masks that highlight only the discriminative parts rather than the entire objects.
As shown in the right example of separate learning for object co-segmentation in Figure~\ref{fig:motivation}, the windows of the top bus are not correctly segmented.
Without manually annotated object masks for guiding the training process of the decoder $\cal{D}$, the dense correspondence fields estimated from semantic matching can provide supervision to address this issue.
Our key insight is that the predicted object masks $M_A$ and $M_B$ should be geometrically consistent with the learned geometric transformations $T_{AB}$ and $T_{BA}$.
We thus propose a \emph{cross-network consistency loss} $\mathcal{L}_\mathrm{task-consis}$ that bridges the outputs of the semantic matching network and the object co-segmentation network.
This loss enforces the consistency between the learned geometric transformations $T_{AB}$ and $T_{BA}$ and the predicted object masks $M_A$ and $M_B$.
To this end, we use $T_{AB}$ to warp $M_A$ and encourage that the warped mask $\tilde{M}_A$ and $M_B$ highly overlap.
We compute the symmetric binary cross-entropy loss and define the cross-network consistency loss $\mathcal{L}_\mathrm{task-consis}$ as
\begin{equation}
  \begin{split}
  \mathcal{L}_\mathrm{task-consis}&(I_A,I_B;\mathcal{E},\mathcal{G},\mathcal{D}) \\
  = &~ \mathcal{L}_\mathrm{bce}(\tilde{M}_A,M_B) + \mathcal{L}_\mathrm{bce}(\tilde{M}_B,M_A),
  \end{split}
  \label{eq:ConsisLoss}
\end{equation}
where $\mathcal{L}_\mathrm{bce}(\tilde{M}_A,M_B)$ computes the binary cross-entropy loss between $\tilde{M}_A$ and $M_B$, and is defined by
\begin{equation}
  \begin{split}
  \mathcal{L}_\mathrm{bce}&(\tilde{M}_A,M_B) \\
  = &~ - \frac{1}{H_B \times W_B}\Bigg(\sum_{i,j}^{}{\tilde{M}_A(i,j)\log\bigg(M_B(i,j)\bigg)} \\
  &~ + \sum_{i,j}^{}{\bigg(1-\tilde{M}_A(i,j)\bigg)\log\bigg(1-M_B(i,j)\bigg)}\Bigg).
  \end{split}
  \label{eq:BCELoss}
\end{equation}

The cross-network consistency loss $\mathcal{L}_\mathrm{task-consis}$ provides supervisory signals for both tasks without the need of ground-truth keypoint correspondences and object masks.

While the model consists of four individual CNN sub-networks, our method \emph{end-to-end} and \emph{jointly} optimizes the training objective in \revised{Eq.}~(\ref{eq:FullObj}) using weak supervision.
The networks for the two tasks are coupled during training, but can be applied independently for each task during inference.

\subsection{Implementation details}

We implement our model using PyTorch.
We adopt the ResNet-$101$~\cite{ResNet} as the encoder $\cal{E}$ and use the feature activations from the \texttt{conv4-23} layer as our feature map.
Similar to~\cite{CNNGeo}, our transformation predictor $\cal{G}$ is a cascade of two modules predicting an affine transformation and a thin plate spline (TPS) transformation, respectively.
We initialize the encoder $\cal{E}$ and the transformation predictor $\cal{G}$ from those in~\cite{WeakMatchNet}.
We construct the decoder $\cal{D}$ with a siamese structure using four blocks, each of which contains one deconvolutional layer and two convolutional layers.
The decoder $\cal{D}$ is randomly initialized.
We add skip connections between each block of the encoder $\cal{E}$ and the decoder $\cal{D}$.
Our network $\cal{F}$ is an ImageNet-pretrained ResNet-$50$~\cite{ResNet} and remains fixed during training. 
All images are resized to the resolution of $240 \times 240$ in advance.
We perform data augmentation by horizontal flipping and random cropping the input images.
We train our model using the ADAM optimizer~\cite{kingma2014adam} with an initial learning rate of $5 \times 10^{-8}$.
For the transitivity consistency loss, 
the input triplets are randomly selected within a mini-batch.
We sample $10 \times 10 = 100$ spatial coordinates for computing the forward-backward consistency loss 
and the transitivity consistency loss.
%
\revised{
We adopt the GrabCut~\cite{rother2004grabcut} method to convert the outputs of the decoder (\ie soft masks) to the object masks (\ie hard masks).
}

\section{Experimental Results}

In this section, we first describe the experimental settings, and then present the quantitative and qualitative evaluation with comparisons to the state-of-the-art methods on \revised{five} benchmark datasets for semantic matching and object co-segmentation.
The source code, the pre-trained models, and additional results are available at \url{https://yunchunchen.github.io/MaCoSNet/}.


\subsection{Evaluation metrics and datasets}

Here we describe the evaluation metrics for semantic matching and object co-segmentation as well as the \revised{five} datasets.

\vspace{\paramargin}
{\flushleft {\bf Evaluation metrics.}}
We evaluate our proposed method on both semantic matching and object co-segmentation tasks.
To measure the performance of semantic matching, we use the commonly used \emph{percentage of correct keypoints} (PCK) metric~\cite{PCK} which calculates the percentage of keypoints whose re-projection errors are less than a given threshold.
The re-projection error is the Euclidean distance $d(T_{AB}(\mathbf{p}), \mathbf{p}^*)$ between the locations of the warped keypoint $T_{AB}(\mathbf{p})$ and the ground-truth keypoint $\mathbf{p}^*$.
The threshold is defined by $\alpha\cdot\max(H,W)$ where $H$ and $W$ are the height and width of the annotated object bounding box on the image, respectively.
We report the performance under different values of $\alpha$. 

For object co-segmentation, we adopt the \emph{precision} $\mathcal{P}$ and the \emph{Jaccard index} $\mathcal{J}$.
The precision $\mathcal{P}$ measures the percentage of correctly classified pixels.
The Jaccard index $\mathcal{J}$ is the ratio of the intersection area of the predicted foreground objects and the ground truths to their union area.
Since background pixels are taken into account in the precision metric, this measure may not precisely reflect the quality of object co-segmentation results.
In contrast, the Jaccard index $\mathcal{J}$ is considered more reliable since it focuses on foreground objects.

\vspace{\paramargin}
{\flushleft {\bf Datasets.}} 
We conduct the experiments on \revised{five} public benchmarks, including the TSS~\cite{Taniai}, Internet~\cite{rubinstein}, PF-PASCAL~\cite{ProposalFlow}, PF-WILLOW~\cite{ProposalFlow}, \revised{and SPair-$71$k~\cite{min2019spair}} datasets.
We use the TSS dataset~\cite{Taniai} for evaluating joint semantic matching and object co-segmentation as the TSS dataset contains the ground-truth annotations for both tasks. 
For object co-segmentation, we use a more challenging Internet dataset~\cite{rubinstein}.
For semantic matching, we use the PF-PASCAL~\cite{ProposalFlow}, PF-WILLOW~\cite{ProposalFlow}, \revised{and SPair-$71$k~\cite{min2019spair}} datasets.
The details of these datasets can be found on the project website of this work. 

\setlength{\tenimg}{0.085\textwidth}

\setlength{\rowmargin}{1.0mm}

\begin{figure*}[t]
\begin{center}

\mpage{\tenimg}{\includegraphics[width=\linewidth]{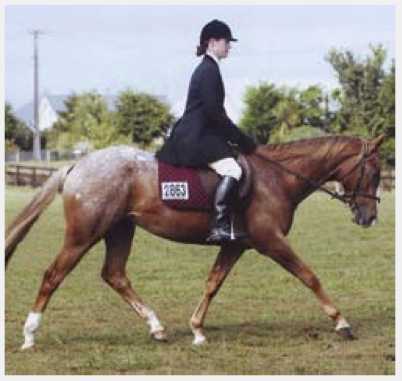}} \hfill
\mpage{\tenimg}{\includegraphics[width=\linewidth]{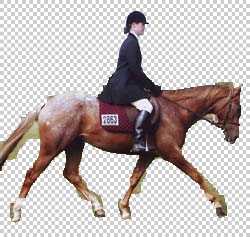}} \hfill
\mpage{\tenimg}{\includegraphics[width=\linewidth]{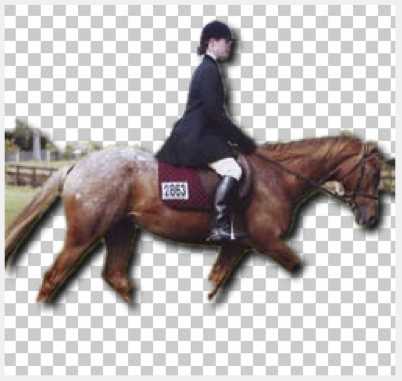}} \hfill
\mpage{\tenimg}{\includegraphics[width=\linewidth]{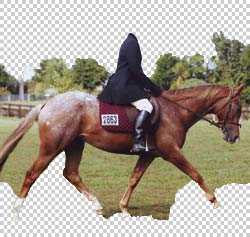}} \hfill
\mpage{\tenimg}{\includegraphics[width=\linewidth]{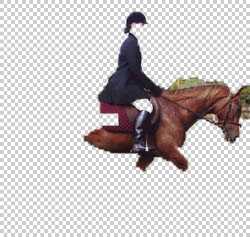}} \hfill
\mpage{\tenimg}{\includegraphics[width=\linewidth]{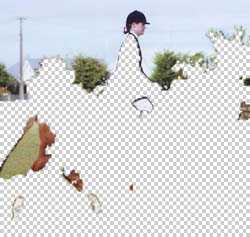}} \hfill
\mpage{\tenimg}{\includegraphics[width=\linewidth]{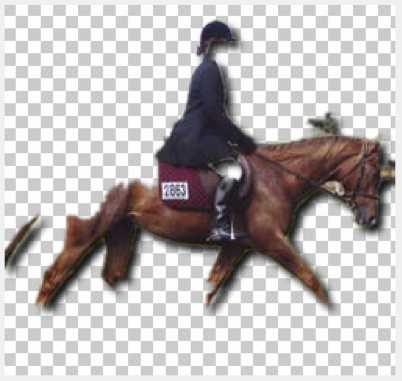}} \hfill
\mpage{\tenimg}{\includegraphics[width=\linewidth]{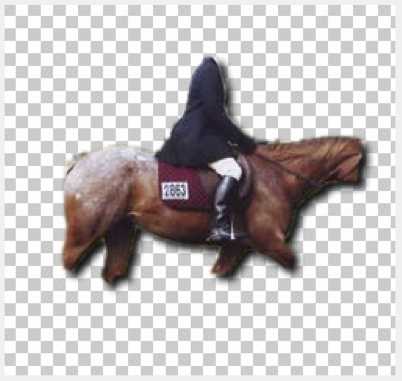}} \hfill
\mpage{\tenimg}{\includegraphics[width=\linewidth]{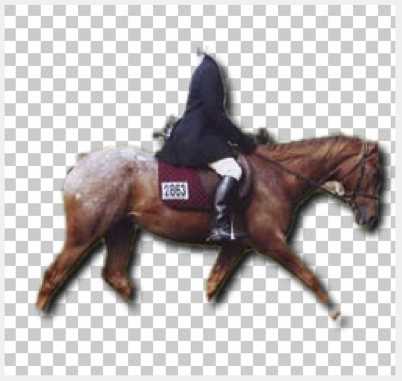}} \hfill
\mpage{\tenimg}{\includegraphics[width=\linewidth]{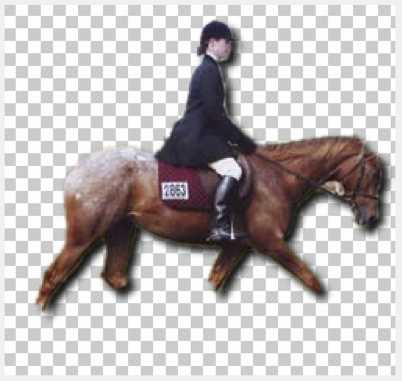}} \\

\vspace{\rowmargin}
\mpage{\tenimg}{\includegraphics[width=\linewidth]{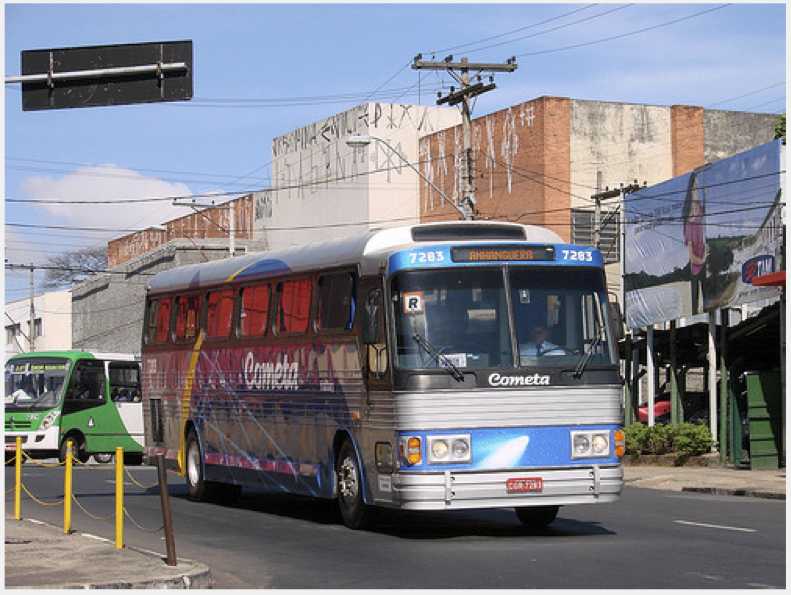}} \hfill
\mpage{\tenimg}{\includegraphics[width=\linewidth]{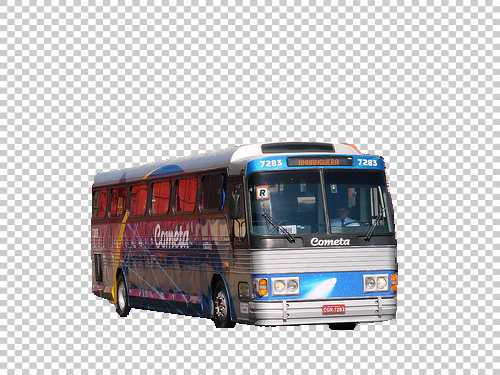}} \hfill
\mpage{\tenimg}{\includegraphics[width=\linewidth]{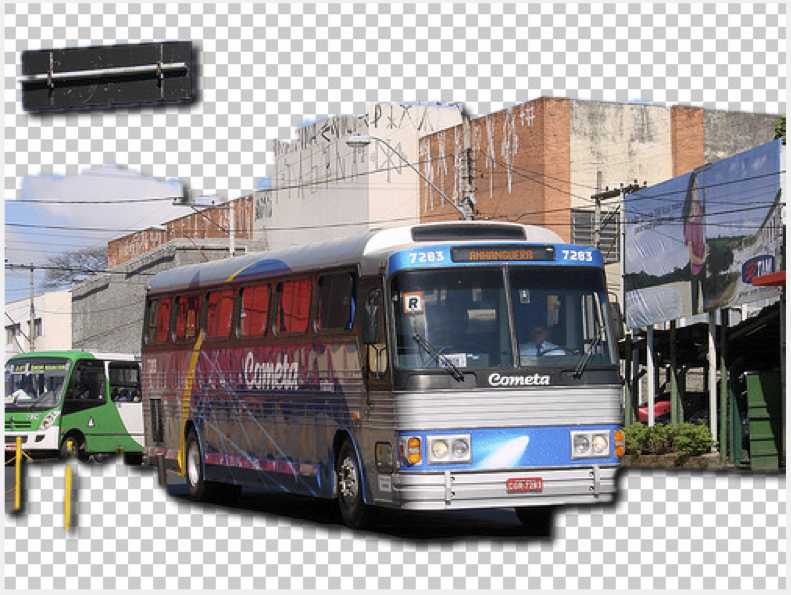}} \hfill
\mpage{\tenimg}{\includegraphics[width=\linewidth]{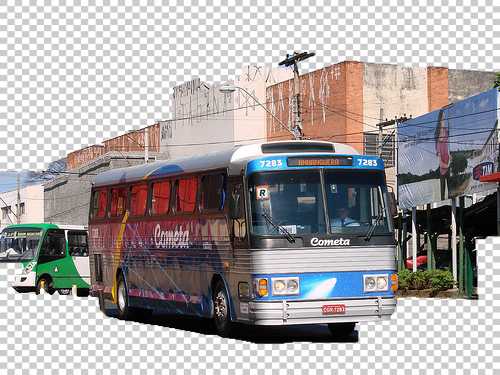}} \hfill
\mpage{\tenimg}{\includegraphics[width=\linewidth]{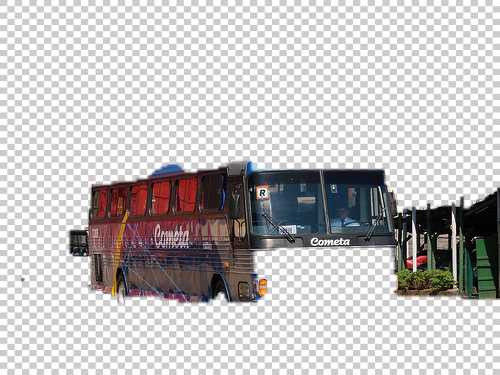}} \hfill
\mpage{\tenimg}{\includegraphics[width=\linewidth]{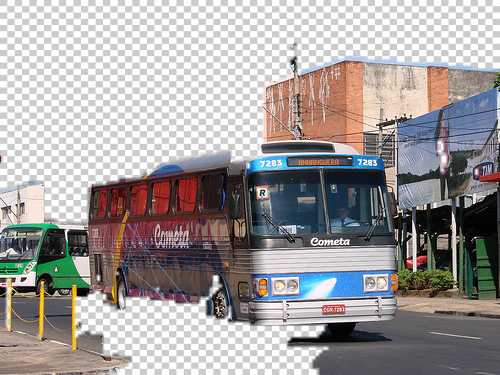}} \hfill
\mpage{\tenimg}{\includegraphics[width=\linewidth]{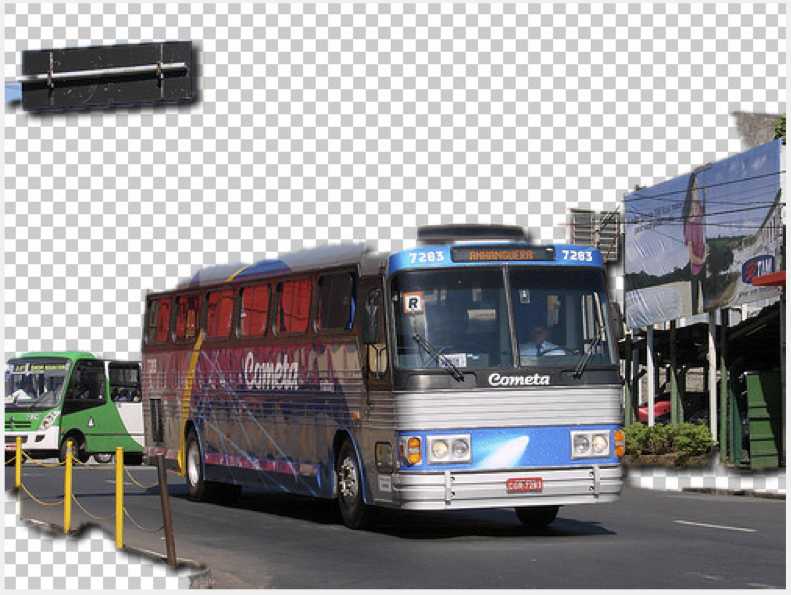}} \hfill
\mpage{\tenimg}{\includegraphics[width=\linewidth]{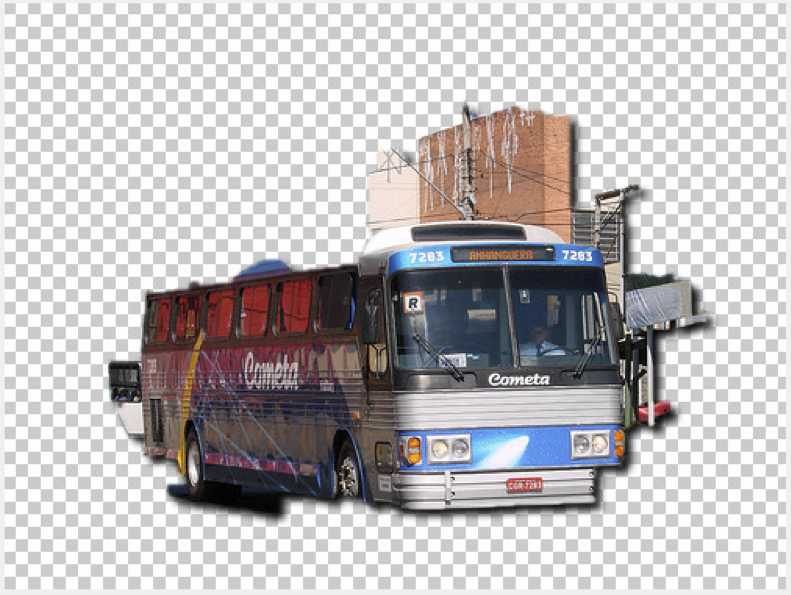}} \hfill
\mpage{\tenimg}{\includegraphics[width=\linewidth]{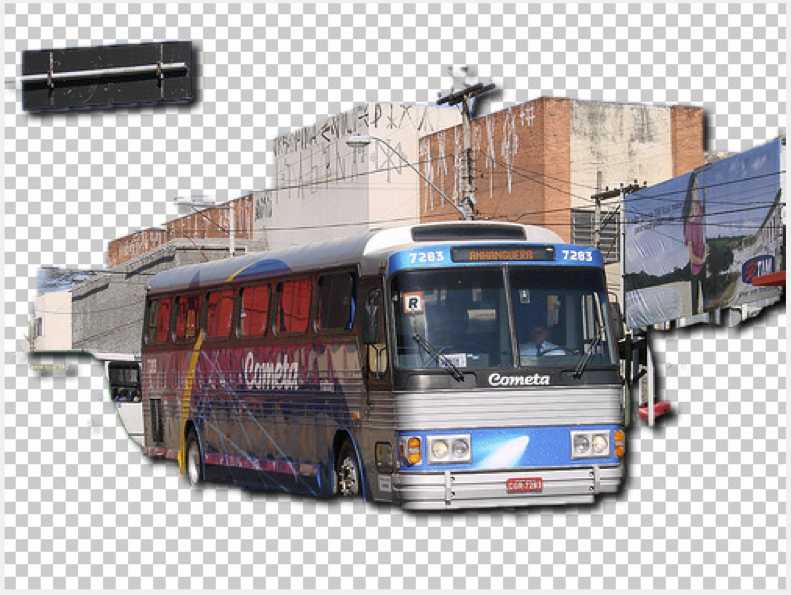}} \hfill
\mpage{\tenimg}{\includegraphics[width=\linewidth]{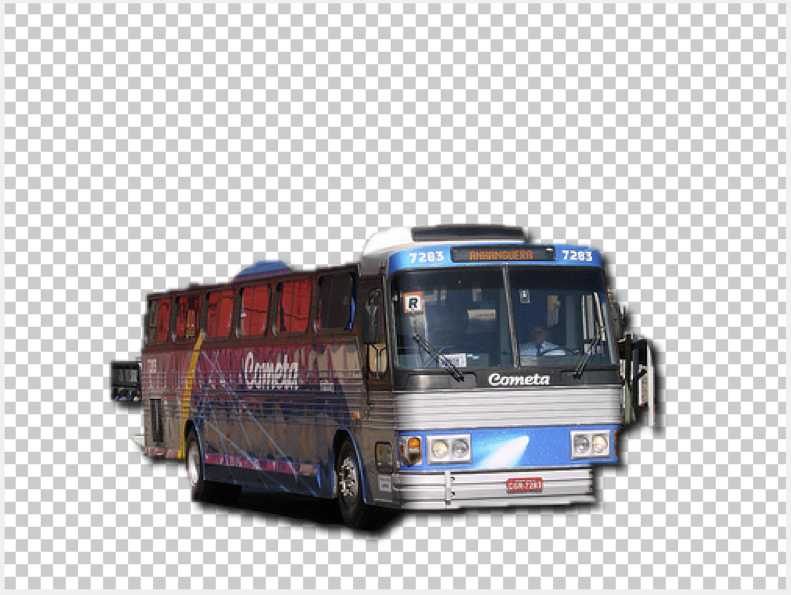}} \\

\vspace{\rowmargin}
\mpage{\tenimg}{\includegraphics[width=\linewidth]{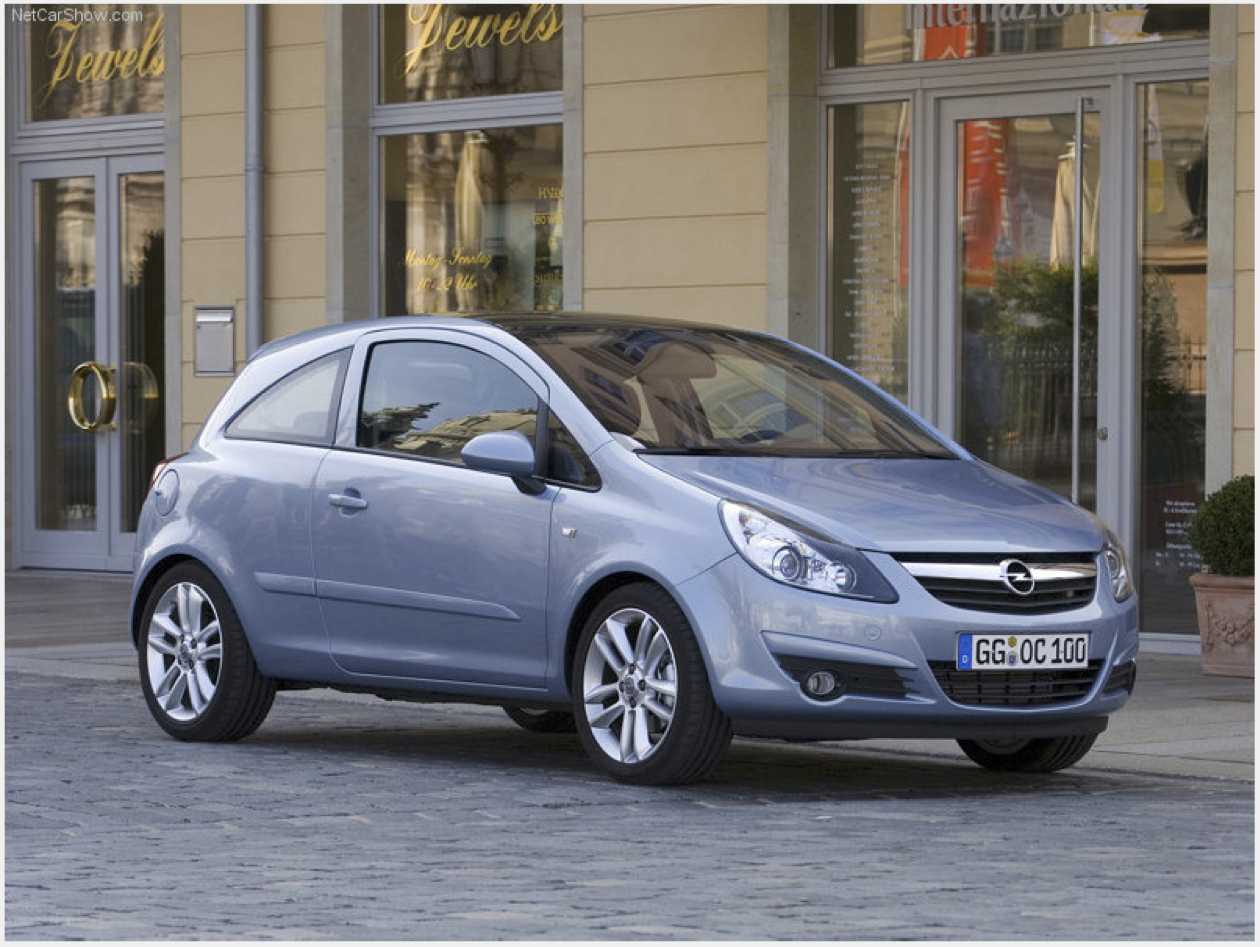}} \hfill
\mpage{\tenimg}{\includegraphics[width=\linewidth]{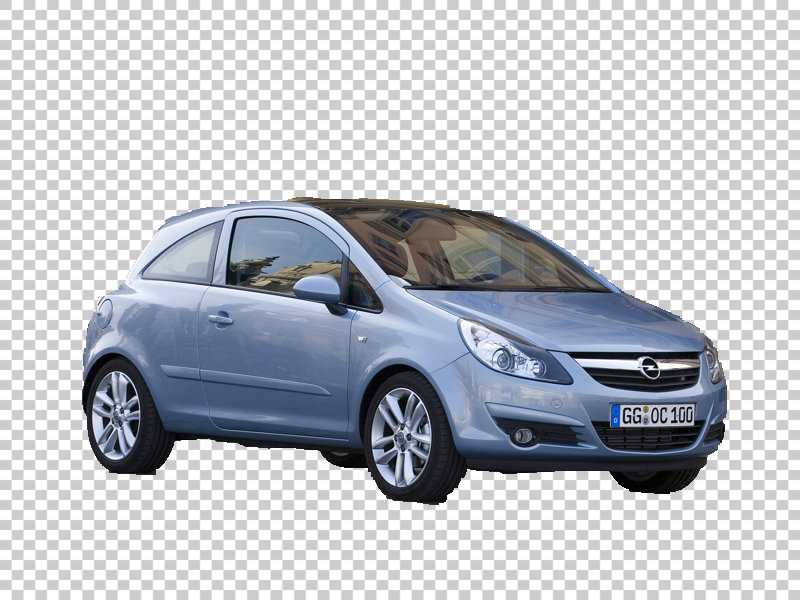}} \hfill
\mpage{\tenimg}{\includegraphics[width=\linewidth]{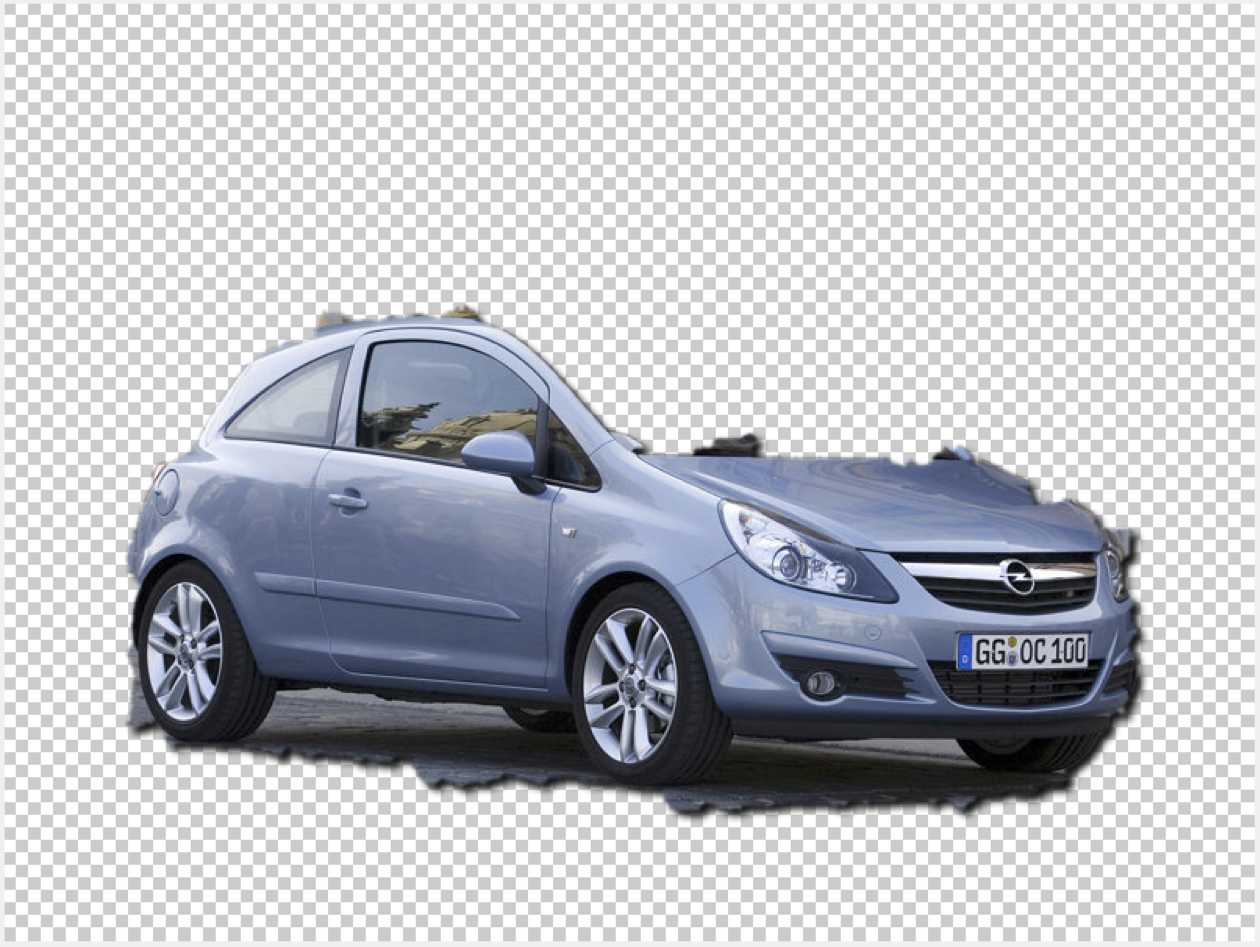}} \hfill
\mpage{\tenimg}{\includegraphics[width=\linewidth]{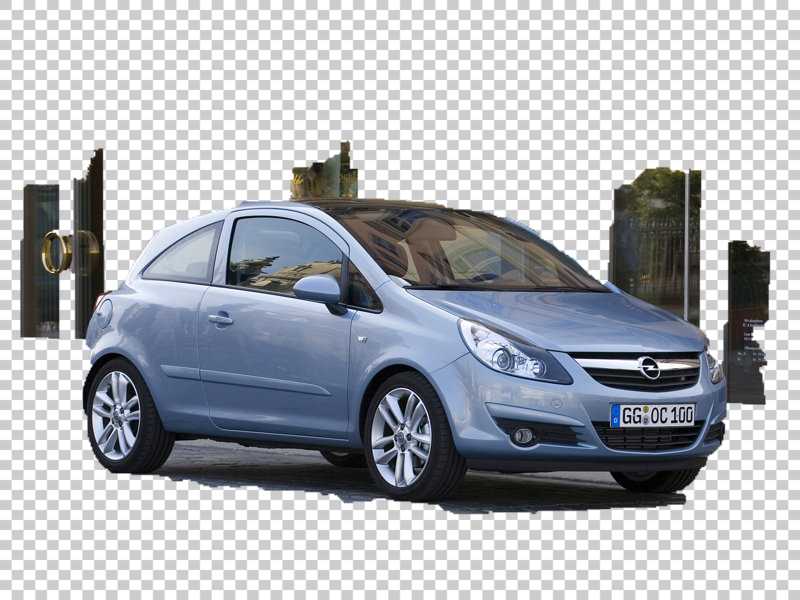}} \hfill
\mpage{\tenimg}{\includegraphics[width=\linewidth]{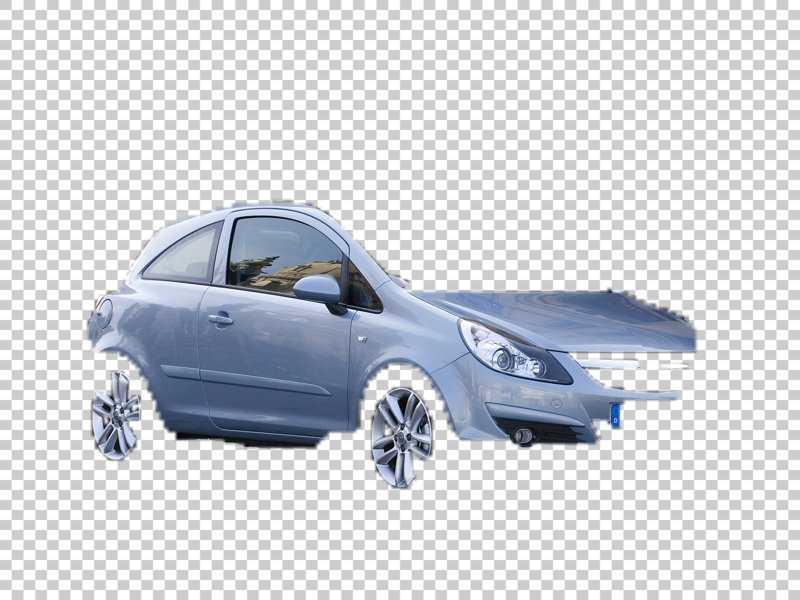}} \hfill
\mpage{\tenimg}{\includegraphics[width=\linewidth]{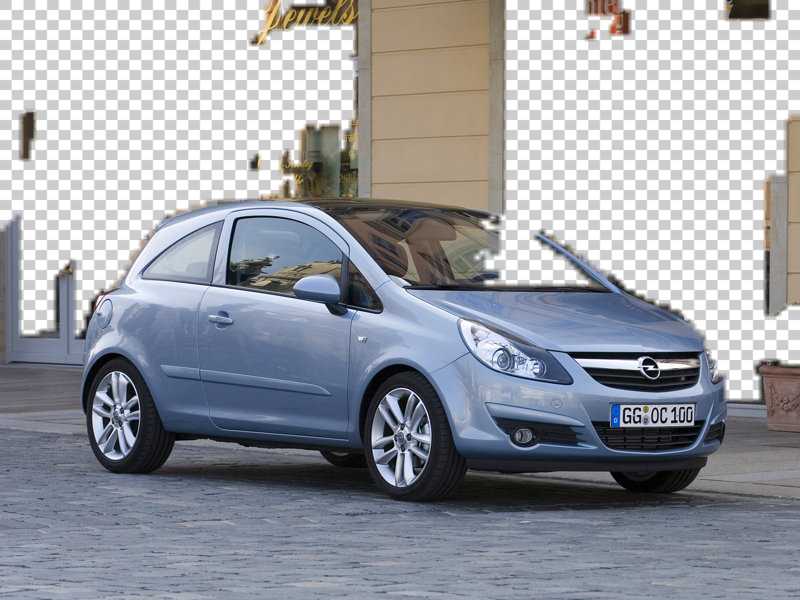}} \hfill
\mpage{\tenimg}{\includegraphics[width=\linewidth]{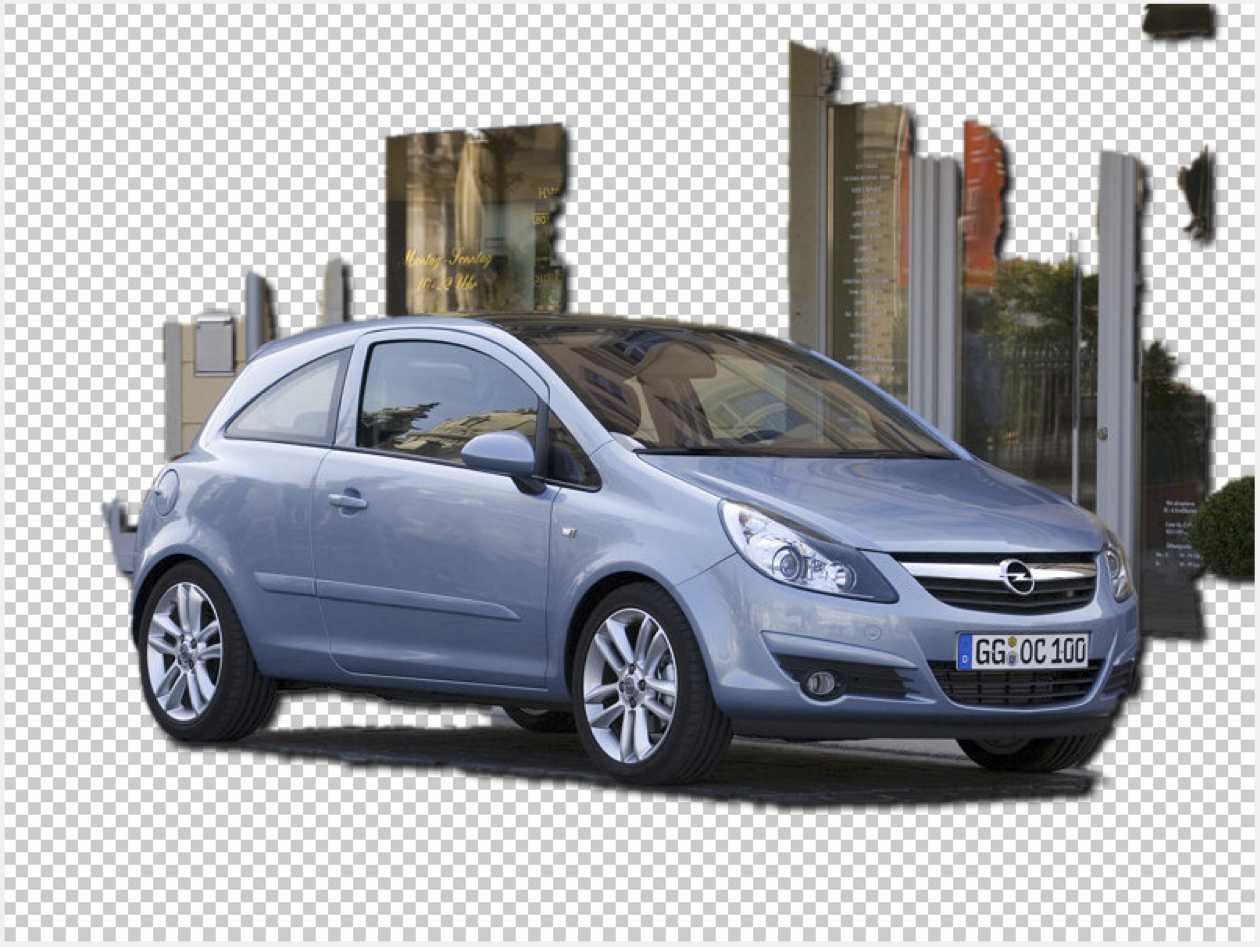}} \hfill
\mpage{\tenimg}{\includegraphics[width=\linewidth]{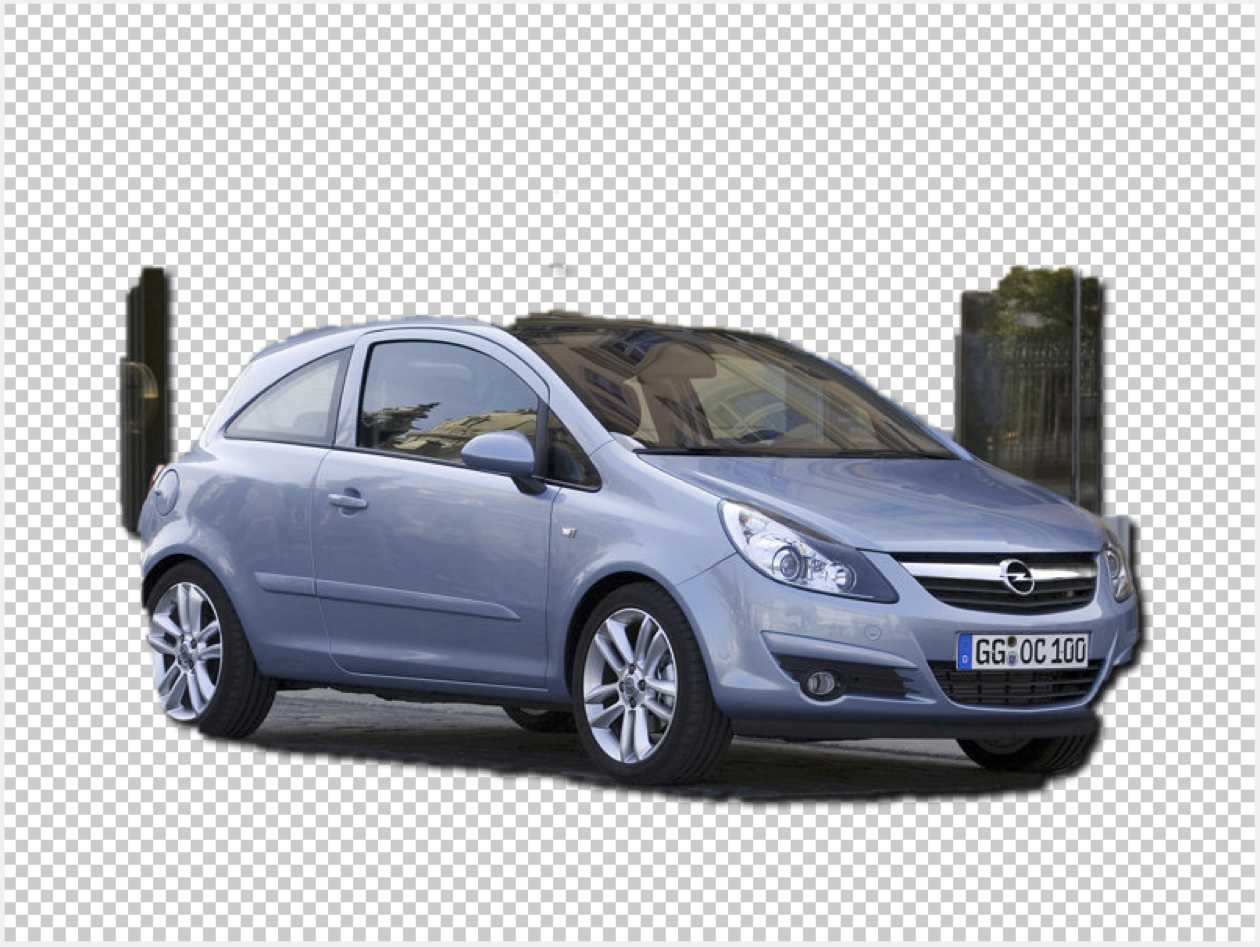}} \hfill
\mpage{\tenimg}{\includegraphics[width=\linewidth]{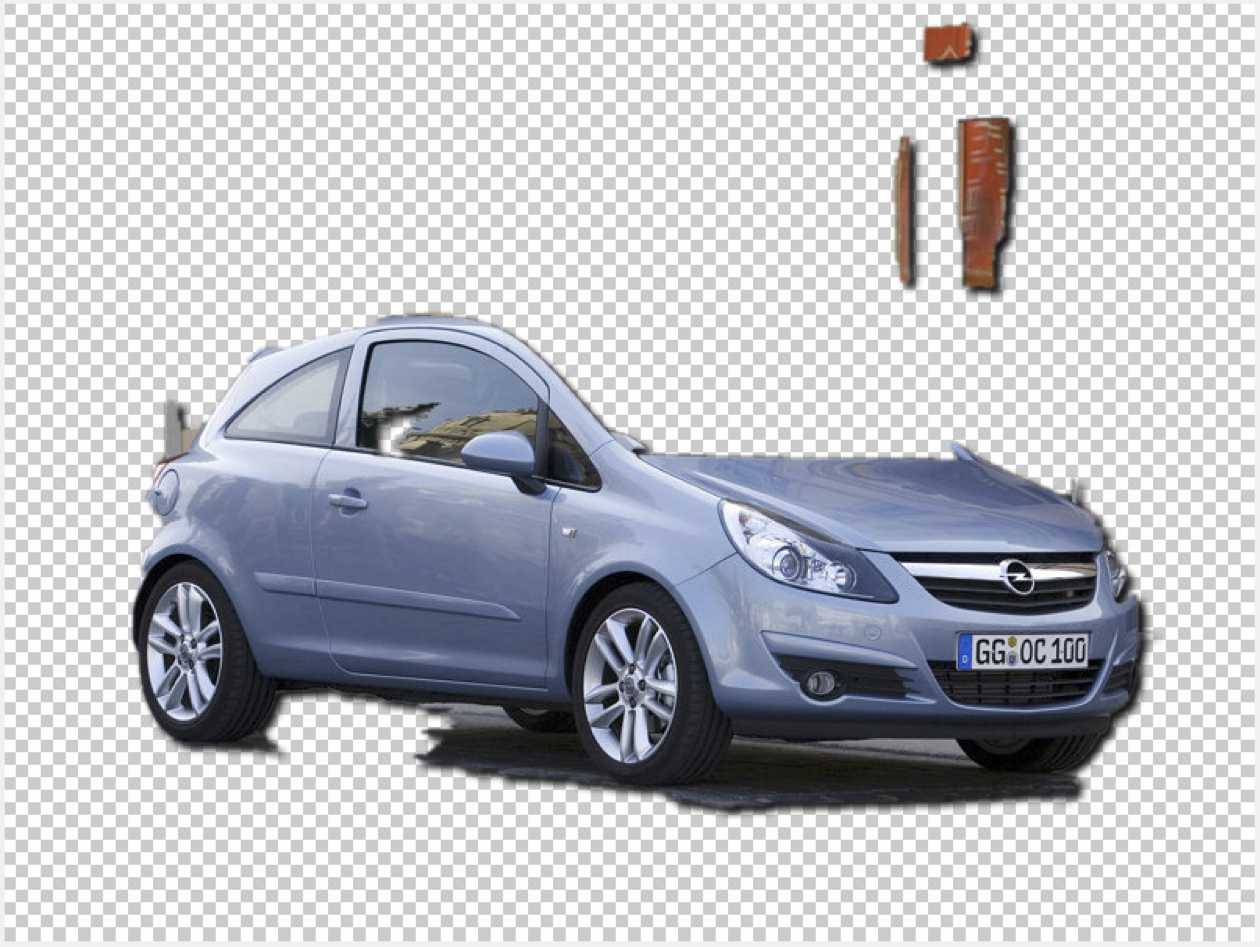}} \hfill
\mpage{\tenimg}{\includegraphics[width=\linewidth]{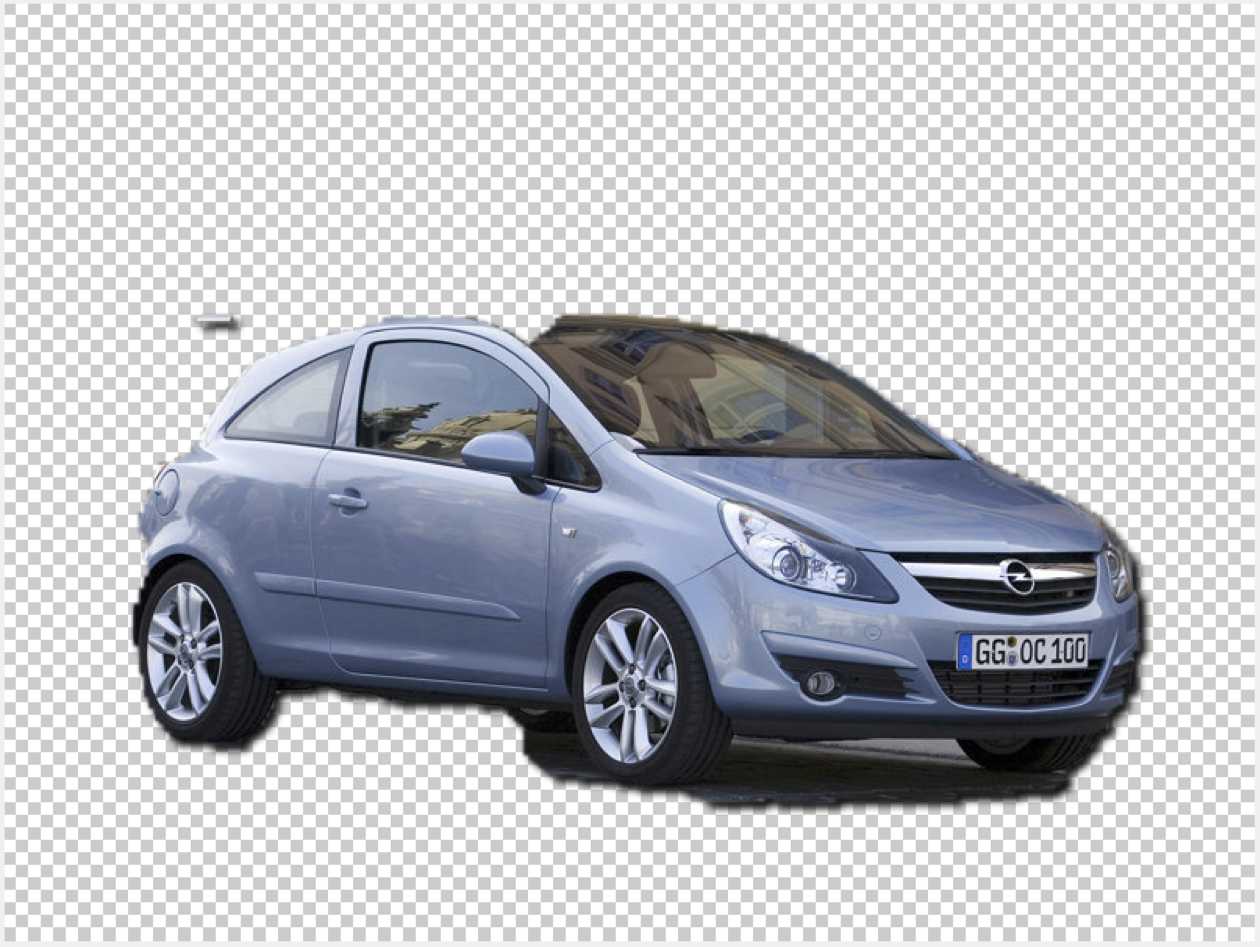}} \\
\vspace{-2mm}
\mpage{\tenimg}{\scriptsize Images} \hfill
\mpage{\tenimg}{\scriptsize Ground truth} \hfill
\mpage{\tenimg}{\scriptsize \cite{Chang15}} \hfill
\mpage{\tenimg}{\scriptsize \cite{Taniai}} \hfill
\mpage{\tenimg}{\scriptsize \cite{faktor2013co}} \hfill
\mpage{\tenimg}{\scriptsize \cite{Joulin10}} \hfill
\mpage{\tenimg}{\scriptsize \cite{Lee15}} \hfill
\mpage{\tenimg}{\scriptsize \cite{Jerripothula16}} \hfill
\mpage{\tenimg}{\scriptsize \cite{Jerripothula17}} \hfill
\mpage{\tenimg}{\scriptsize Ours} \\

\caption{\textbf{Visual comparisons of object co-segmentation on the TSS dataset~\cite{Taniai}.} 
}

\label{fig:TSS-coseg}
\end{center}

\end{figure*}

\begin{table}[t]
  \caption{
  \textbf{Semantic matching results on the TSS dataset~\cite{Taniai}.}
  `Strong' denotes the method is learned with keypoint supervision.
  `Weak' denotes the method is learned with image-level supervision.
  %
  %
  The bold and underlined numbers indicate the top two results, respectively.
  }
  \label{table:TSS-matching}
    \scriptsize
    \ra{1.2}
    \begin{center}
      \resizebox{\linewidth}{!} 
      {
      \begin{tabular}{l|cc|c|c|c|c}
    \toprule
    Method & Descriptor & Supervision & FG3DCar & JODS & PASCAL & Avg. \\
    \midrule
    SIFT Flow~\cite{SIFTFlow} & SIFT~\cite{SIFT} & - & 0.632 & 0.509 & 0.360 & 0.500 \\
    DSP~\cite{Deformable} & SIFT~\cite{SIFT} & - & 0.487 & 0.465 & 0.382 & 0.445 \\
    TSS~\cite{Taniai} & HOG~\cite{HoG} & - & 0.829 & 0.595 & 0.483 & 0.636 \\
    Proposal Flow + LOM~\cite{ProposalFlow,ProposalFlow-CVPR} & HOG~\cite{HoG} & - & 0.786 & 0.653 & 0.531 & 0.657 \\
    DAISY~\cite{DAISY} & DAISY~\cite{DAISY} & - & 0.636 & 0.373 & 0.338 & 0.449 \\
    DFF~\cite{DFF} & DAISY~\cite{DAISY} & - & 0.495 & 0.304 & 0.224 & 0.341 \\
    LSS~\cite{shechtman2007matching} & LSS~\cite{shechtman2007matching} & - & 0.644 & 0.349 & 0.359 & 0.451 \\
    DASC~\cite{kim2015dasc} & DASC~\cite{kim2015dasc} & - & 0.668 & 0.454 & 0.261 & 0.461 \\
    MatchNet~\cite{han2015matchnet} & AlexNet~\cite{krizhevsky2012imagenet} & - & 0.561 & 0.380 & 0.270 & 0.404 \\
    3D-guided~\cite{3D-cycle} & 3D-guided~\cite{3D-cycle} & - & 0.721 & 0.514 & 0.436 & 0.556 \\
    OHG~\cite{ObjAware} & HOG~\cite{HoG} & - & 0.875 & 0.708 & \textbf{0.729} & \textbf{0.771} \\
    \midrule
    UCN~\cite{UCN} & GoogLeNet~\cite{szegedy2015going} & Strong & 0.853 & 0.672 & 0.511 & 0.679 \\
    FCSS~\cite{FCSS,FCSS-PAMI} & FCSS~\cite{FCSS,FCSS-PAMI} & Strong & 0.830 & 0.656 & 0.494 & 0.660 \\
    Proposal Flow + LOM~\cite{ProposalFlow,ProposalFlow-CVPR} & FCSS~\cite{FCSS,FCSS-PAMI} & Strong & 0.839 & 0.635 & 0.582 & 0.685 \\
    DCTM~\cite{DCTM,DCTM-PAMI} & VGG-16~\cite{VGG} & Strong & 0.790 & 0.611 & 0.528 & 0.630 \\
    DCTM~\cite{DCTM,DCTM-PAMI} & FCSS~\cite{FCSS,FCSS-PAMI} & Strong & 0.891 & 0.721 & 0.610 & 0.740 \\
    SCNet-A~\cite{SCNet} & VGG-16~\cite{VGG} & Strong & 0.774 & 0.574 & 0.476 & 0.608 \\
    SCNet-AG~\cite{SCNet} & VGG-16~\cite{VGG} & Strong & 0.764 & 0.600 & 0.463 & 0.609 \\
    SCNet-AG+~\cite{SCNet} & VGG-16~\cite{VGG} & Strong & 0.776 & 0.608 & 0.474 & 0.619 \\
    CNNGeo~\cite{CNNGeo} & VGG-16~\cite{VGG} & Strong & 0.835 & 0.656 & 0.527 & 0.673 \\
    CNNGeo~\cite{CNNGeo} & ResNet-101~\cite{ResNet} & Strong & 0.886 & 0.758 & 0.560 & 0.735 \\
    \midrule
    CNNGeo w/ Inlier~\cite{End-to-end} & ResNet-101~\cite{ResNet} & Weak & 0.892 & 0.758 & 0.562 & 0.737 \\
    Ours w/o co-seg~\cite{WeakMatchNet} & ResNet-101~\cite{ResNet} & Weak & \underline{0.907} & \underline{0.781} & 0.565 & 0.751 \\
    Ours & ResNet-101~\cite{ResNet} & Weak & \textbf{0.908} & \textbf{0.783} & \underline{0.615} & \underline{0.769} \\
    \midrule
    \midrule
    RTNs~\cite{RTN} & VGG-16~\cite{VGG} & Weak & 0.893 & 0.762 & 0.591 & 0.749 \\
    RTNs~\cite{RTN} & FCSS~\cite{FCSS,FCSS-PAMI} & Weak & 0.889 & 0.775 & 0.611 & 0.758 \\
    RTNs~\cite{RTN} & ResNet-101~\cite{ResNet} & Weak & \underline{0.901} & \underline{0.782} & 0.633 & 0.772 \\
    PARN~\cite{PARN} & VGG-16~\cite{VGG} & Weak & 0.876 & 0.716 & \underline{0.688} & 0.760 \\
    PARN~\cite{PARN} & ResNet-101~\cite{ResNet} & Weak & 0.895 & 0.759 & \textbf{0.712} & \underline{0.788} \\
    Ours & ResNet-101~\cite{ResNet} & Weak & \textbf{0.912} & \textbf{0.790} & 0.673 & \textbf{0.792} \\
    \bottomrule
    \end{tabular}
      }
    \end{center}
\end{table}


\begin{table}[t]
  \caption{
  \textbf{Experimental results of object co-segmentation on the TSS dataset~\cite{Taniai}.} 
  The bold and underlined numbers indicate the top two results, respectively. }
  \label{table:TSS-coseg}
  \scriptsize
    \ra{1.2}
    \begin{center}
      \resizebox{\linewidth}{!} 
      {
      \begin{tabular}{l|c|cc|cc|cc|cc}
      \toprule
      \multirow{2}{*}{Method} & \multirow{2}{*}{Descriptor} & \multicolumn{2}{c|}{FG3DCar} & \multicolumn{2}{c|}{JODS} & \multicolumn{2}{c|}{PASCAL} & \multicolumn{2}{c}{Avg.} \\
      & & ${\cal P}$ & ${\cal J}$ & ${\cal P}$ & ${\cal J}$ & ${\cal P}$ & ${\cal J}$ & ${\cal P}$ & ${\cal J}$ \\
      \midrule 
      SIFT Flow~\cite{SIFTFlow} & SIFT~\cite{SIFT} & 0.661 & 0.42 & 0.557 & 0.24 & 0.628 & 0.41 & 0.615 & 0.36 \\
      DSP~\cite{Deformable} & SIFT~\cite{SIFT} & 0.502 & 0.29 & 0.454 & 0.22 & 0.496 & 0.34 & 0.484 & 0.28 \\
      Hati~\etal~\cite{Hati16} & SIFT~\cite{SIFT} & 0.785 & 0.47 & 0.778 & 0.31 & 0.701 & 0.31 & 0.755 & 0.36 \\
      Chang~\etal~\cite{Chang15} & SIFT~\cite{SIFT} & 0.872 & 0.67 & 0.851 & 0.52 & 0.723 & 0.40 & 0.815 & 0.53 \\
      MRW~\cite{Lee15} & SIFT~\cite{SIFT} & 0.784 & 0.63 & 0.730 & 0.46 & 0.804 & 0.66 & 0.773 & 0.58 \\
      Jerripothula~\etal~\cite{Jerripothula17} & SIFT~\cite{SIFT} & 0.885 & 0.76 & 0.822 & 0.57 & 0.830 & 0.56 & 0.846 & 0.63 \\
      Jerripothula~\etal~\cite{Jerripothula16} & SIFT~\cite{SIFT} & 0.913 & 0.78 & 0.900 & 0.65 & \underline{0.880} & \underline{0.73} & 0.898 & 0.72 \\
      Faktor~\etal~\cite{faktor2013co} & HOG~\cite{HoG} & 0.873 & 0.69 & 0.859 & 0.54 & 0.771 & 0.50 & 0.834 & 0.58 \\
      Joulin~\etal~\cite{Joulin10} & SIFT~\cite{SIFT} & 0.651 & 0.46 & 0.626 & 0.32 & 0.587 & 0.40 & 0.621 & 0.39 \\
      DFF~\cite{DFF} & DAISY~\cite{DAISY} & 0.704 & 0.33 & 0.696 & 0.21 & 0.601 & 0.21 & 0.667 & 0.25 \\
      TSS~\cite{Taniai} & HOG~\cite{HoG} & 0.877 & 0.76 & 0.761 & 0.50 & 0.778 & 0.65 & 0.805 & 0.63 \\
      \midrule
      Ours w/o matching & ResNet-101~\cite{ResNet} & \underline{0.958} & \underline{0.88} & \underline{0.911} & \underline{0.71} & 0.829 & 0.61 & \underline{0.899} & \underline{0.73} \\
      Ours & ResNet-101~\cite{ResNet} & \textbf{0.963} & \textbf{0.90} & \textbf{0.940} & \textbf{0.77} & \textbf{0.939} & \textbf{0.86} & \textbf{0.947} & \textbf{0.84} \\
      \bottomrule
      \end{tabular}
      }
    \end{center}
\end{table}

\subsection{Joint matching and co-segmentation}

{\flushleft {\bf Results on the TSS dataset.}} 
Table~\ref{table:TSS-matching} shows the quantitative results of semantic matching on the TSS~\cite{Taniai} dataset.
In this experiment, we set the hyper-parameters as follows: $\lambda_\mathrm{cycle}=5$, $\lambda_\mathrm{trans}=5$, $\lambda_\mathrm{contrast}=10$, and $\lambda_\mathrm{task}=10$.
%
Overall, the proposed method achieves favorable performance against all competing approaches.
%
%
%
In the bottom block of Table~\ref{table:TSS-matching}, we follow the PARN~\cite{PARN} and RTNs~\cite{RTN}, and resize all images to the larger dimension to $100$ (\ie resizing $\max(H,W)$ to $100$).
The proposed method also performs favorably against all competing methods.


\setlength{\eightimg}{0.11\textwidth}
\setlength{\rowmargin}{0.5mm}

\begin{figure*}[t]
\begin{center}

\mpage{\eightimg}{\includegraphics[width=\linewidth]{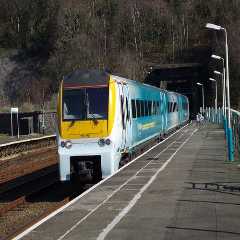}} 
\mpage{\eightimg}{\includegraphics[width=\linewidth]{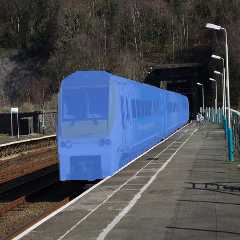}} 
\mpage{\eightimg}{\includegraphics[width=\linewidth]{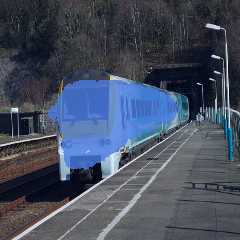}} 
\mpage{\eightimg}{\includegraphics[width=\linewidth]{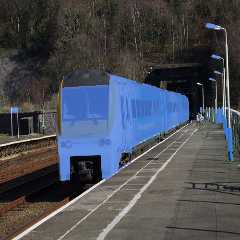}} \hfill
\mpage{\eightimg}{\includegraphics[width=\linewidth]{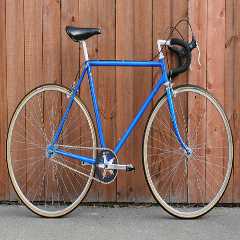}} 
\mpage{\eightimg}{\includegraphics[width=\linewidth]{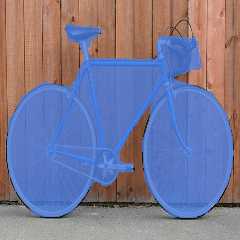}} 
\mpage{\eightimg}{\includegraphics[width=\linewidth]{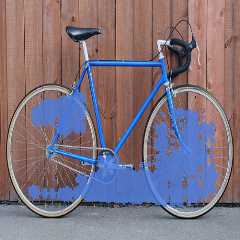}} 
\mpage{\eightimg}{\includegraphics[width=\linewidth]{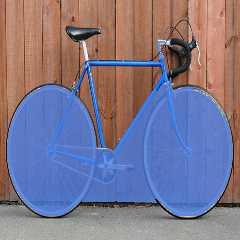}} \\

\vspace{0.25mm}
\mpage{\eightimg}{\includegraphics[width=\linewidth]{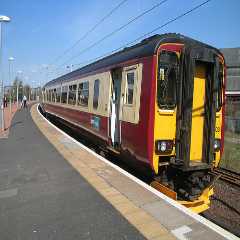}} 
\mpage{\eightimg}{\includegraphics[width=\linewidth]{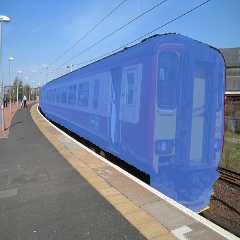}} 
\mpage{\eightimg}{\includegraphics[width=\linewidth]{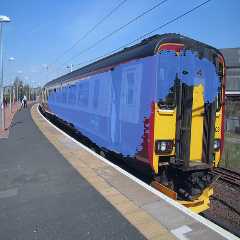}} 
\mpage{\eightimg}{\includegraphics[width=\linewidth]{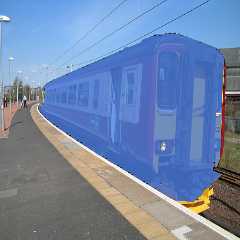}} \hfill
\mpage{\eightimg}{\includegraphics[width=\linewidth]{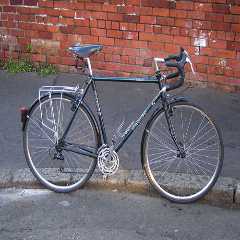}} 
\mpage{\eightimg}{\includegraphics[width=\linewidth]{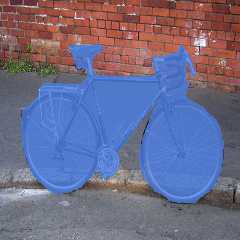}} 
\mpage{\eightimg}{\includegraphics[width=\linewidth]{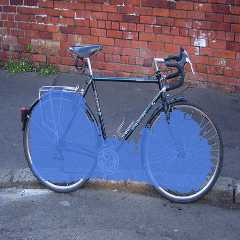}} 
\mpage{\eightimg}{\includegraphics[width=\linewidth]{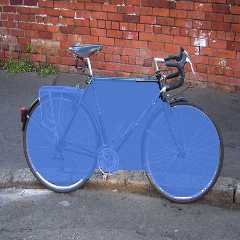}} \\

\vspace{\rowmargin}
\mpage{\eightimg}{\includegraphics[width=\linewidth]{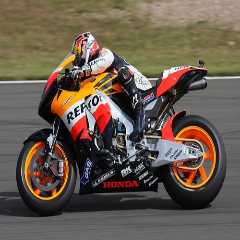}} 
\mpage{\eightimg}{\includegraphics[width=\linewidth]{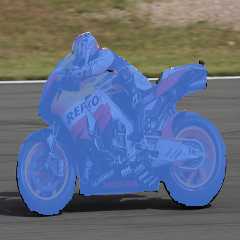}} 
\mpage{\eightimg}{\includegraphics[width=\linewidth]{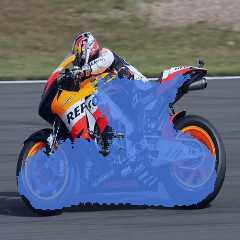}} 
\mpage{\eightimg}{\includegraphics[width=\linewidth]{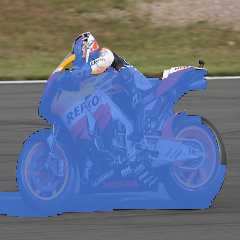}} \hfill
\mpage{\eightimg}{\includegraphics[width=\linewidth]{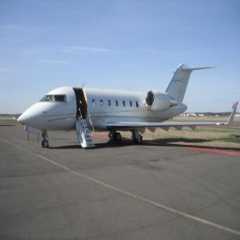}} 
\mpage{\eightimg}{\includegraphics[width=\linewidth]{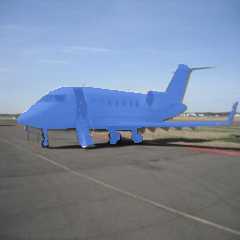}} 
\mpage{\eightimg}{\includegraphics[width=\linewidth]{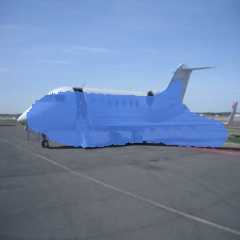}} 
\mpage{\eightimg}{\includegraphics[width=\linewidth]{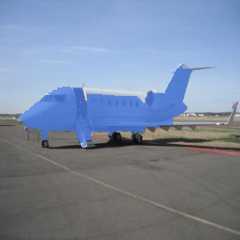}} \\

\vspace{0.25mm}
\mpage{\eightimg}{\includegraphics[width=\linewidth]{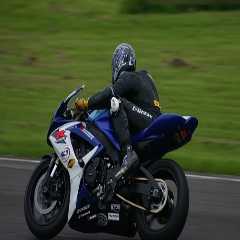}} 
\mpage{\eightimg}{\includegraphics[width=\linewidth]{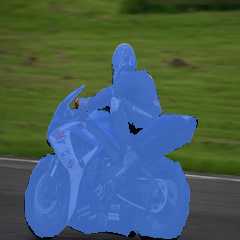}} 
\mpage{\eightimg}{\includegraphics[width=\linewidth]{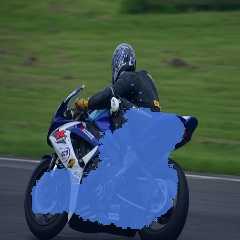}} 
\mpage{\eightimg}{\includegraphics[width=\linewidth]{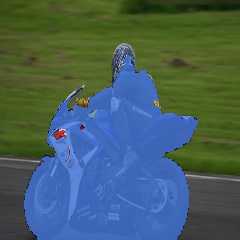}} \hfill
\mpage{\eightimg}{\includegraphics[width=\linewidth]{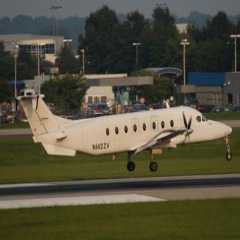}} 
\mpage{\eightimg}{\includegraphics[width=\linewidth]{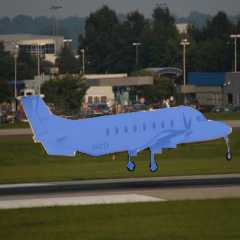}} 
\mpage{\eightimg}{\includegraphics[width=\linewidth]{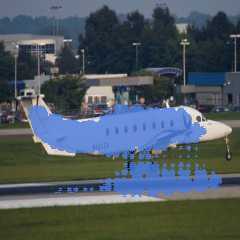}} 
\mpage{\eightimg}{\includegraphics[width=\linewidth]{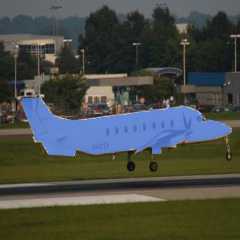}} \\
\vspace{-1mm}
\mpage{\eightimg}{\scriptsize Images} 
\mpage{\eightimg}{\scriptsize Ground truth}
\mpage{\eightimg}{\scriptsize w/o matching} 
\mpage{\eightimg}{\scriptsize w/ matching} \hfill
\mpage{\eightimg}{\scriptsize Images}
\mpage{\eightimg}{\scriptsize Ground truth} 
\mpage{\eightimg}{\scriptsize w/o matching}
\mpage{\eightimg}{\scriptsize w/ matching}

\caption{\textbf{The advantage of joint semantic matching and object co-segmentation.} We present four examples from the TSS dataset~\cite{Taniai}. Integrating semantic matching with object co-segmentation helps improve the quality of co-segmentation.
}

\label{Fig:TSS-Coseg}

\end{center}
\end{figure*}

Table~\ref{table:TSS-coseg} shows the quantitative results of object co-segmentation on the TSS~\cite{Taniai} dataset.
The proposed method achieves a precision of $94.7\%$ and a Jaccard index of $84\%$, and performs favorably against the state-of-the-art methods by a large margin $4.9\%$ in precision and $12\%$ in Jaccard index against the best competitor~\cite{Jerripothula16}.
We attribute the significant performance gains to two factors.
First, unlike most existing methods, our proposed approach tackles object co-segmentation with an end-to-end trainable model.
Second, integrating semantic matching further improves object co-segmentation.
Figure~\ref{fig:TSS-coseg} presents visual comparisons of object co-segmentation with existing methods.
The proposed method generates more accurate co-segmentation results, particularly when images contain drastic background clutters and large intra-class appearance variations.

{\flushleft {\bf Effect of joint learning.}} 
We conduct an ablation study on joint learning of semantic matching and object co-segmentation. 
We investigate two variants:
(1) \emph{Ours w/o co-seg}: disabling the object co-segmentation network stream (falling back to our preliminary results \emph{WeakMatchNet}~\cite{WeakMatchNet}) and 
(2) \emph{Ours w/o matching}: disabling the semantic matching stream.
%
%
Table~\ref{table:TSS-matching} and Table~\ref{table:TSS-coseg} show the quantitative results of these two variant methods.
For semantic matching, our model suffers a performance loss of $1.8\%$ in average PCK at $\alpha = 0.05$.
For object co-segmentation, our results show a drop of $4.8\%$ in precision and a $11\%$ drop in Jaccard index.
The proposed coupled training approach significantly improves the performance for both tasks.
%
%
In particular, the improvement over \emph{Ours w/o co-seg}~\cite{WeakMatchNet} indicates the benefits of explicit object mask estimation using object co-segmentation.
%
%

Figure~\ref{Fig:TSS-Coseg} shows four examples of the qualitative object co-segmentation results.
The object co-segmentation model (\emph{Ours w/o matching}) may focus only on the most discriminative parts as reflected by the motorbike example (the model focuses on segmenting the wheels of the motorbike).
Many false positives and false negatives are generated due to drastic appearance variations.
With the guidance of geometric transformations inferred from semantic matching, our joint training model significantly alleviates these unfavorable false positives and false negatives, resulting in more accurate and consistent object co-segmentation results.

\begin{table}[t]
  \scriptsize
  \caption{
  \textbf{Experimental results of object co-segmentation on the Internet dataset~\cite{rubinstein}.}
  Marker $^*$ indicates that method is learned with strong supervision (\ie manually annotated object masks).
  The bold and underlined numbers indicate the top two results, respectively.
  }
  \label{table:Coseg}
  \centering
  \vspace{4mm}
  \resizebox{\linewidth}{!} 
  {
  \begin{tabular}{l|c|cc|cc|cc|cc}
  \toprule
  \multirow{2}{*}{Method} & \multirow{2}{*}{Descriptor} & \multicolumn{2}{c|}{Airplane} & \multicolumn{2}{c|}{Car} & \multicolumn{2}{c|}{Horse} & \multicolumn{2}{c}{Avg.} \\
  & & $\cal{P}$ & $\cal{J}$ & $\cal{P}$ & $\cal{J}$ & $\cal{P}$ & $\cal{J}$ & $\cal{P}$ & $\cal{J}$ \\
  \midrule
  DOCS~\cite{li2018deep}$^*$ & VGG-16~\cite{VGG} & 0.946 & 0.64 & 0.940 & 0.83 & 0.914 & 0.65 & 0.933 & 0.70 \\
  \midrule
  Sun~\etal~\cite{Sun16} & HOG~\cite{HoG} & 0.886 & 0.36 & 0.870 & 0.73 & 0.876 & 0.55 & 0.877 & 0.55 \\
  Joulin~\etal~\cite{Joulin10} & SIFT~\cite{SIFT} & 0.493 & 0.15 & 0.587 & 0.37 & 0.638 & 0.30 & 0.572 & 0.27 \\
  Joulin~\etal~\cite{Joulin12} & SIFT~\cite{SIFT} & 0.475 & 0.12 & 0.592 & 0.35 & 0.642 & 0.30 & 0.570 & 0.24 \\
  Kim~\etal~\cite{kim2011distributed} & SIFT~\cite{SIFT} & 0.802 & 0.08 & 0.689 & 0.0004 & 0.751 & 0.06 & 0.754 & 0.05 \\
  Rubinstein~\etal~\cite{rubinstein} & SIFT~\cite{SIFT} & 0.880 & 0.56 & 0.854 & 0.64 & 0.828 & 0.52 & 0.827 & 0.43 \\
  Chen~\etal~\cite{XChen14} & HOG~\cite{HoG} & 0.902 & 0.40 & 0.876 & 0.65 & \underline{0.893} & 0.58 & 0.890 & 0.54 \\
  Quan~\etal~\cite{Quan16} & SIFT~\cite{SIFT} & 0.910 & 0.56 & 0.885 & 0.67 & \underline{0.893} & 0.58 & 0.896 & 0.60 \\
  Hati~\etal~\cite{Hati16} & SIFT~\cite{SIFT} & 0.777 & 0.33 & 0.621 & 0.43 & 0.738 & 0.20 & 0.712 & 0.32 \\
  Chang~\etal~\cite{Chang15} & SIFT~\cite{SIFT} & 0.726 & 0.27 & 0.759 & 0.36 & 0.797 & 0.36 & 0.761 & 0.33 \\
  MRW~\cite{Lee15} & SIFT~\cite{SIFT} & 0.528 & 0.36 & 0.647 & 0.42 & 0.701 & 0.39 & 0.625 & 0.39 \\
  Jerripothula~\etal~\cite{Jerripothula16} & SIFT~\cite{SIFT} & 0.905 & 0.61 & 0.880 & 0.71 & 0.883 & \underline{0.61} & 0.889 & 0.64 \\
  Jerripothula~\etal~\cite{Jerripothula17} & SIFT~\cite{SIFT} & 0.818 & 0.48 & 0.847 & 0.69 & 0.813 & 0.50 & 0.826 & 0.56 \\
  Hsu~\etal~\cite{hsu2018co} & VGG-16~\cite{VGG} & \underline{0.936} & \textbf{0.66} & \underline{0.914} & \underline{0.79} & 0.876 & 0.59 & \underline{0.909} & \underline{0.68} \\
  %
  %
  Ours & VGG-16~\cite{VGG} & 0.928 & \underline{0.65} & 0.912 & 0.78 & 0.865 & \underline{0.61} & 0.902 & \underline{0.68} \\
  Ours & ResNet-101~\cite{ResNet} & \textbf{0.941} & \underline{0.65} & \textbf{0.940} & \textbf{0.82} & \textbf{0.922} & \textbf{0.63} & \textbf{0.935} & \textbf{0.70} \\
  \bottomrule
  \end{tabular}
  }
\end{table}

\setlength{\elevenimg}{0.078\textwidth}

\setlength{\rowmargin}{0.5mm}

\begin{figure*}[t]
\begin{center}

\vspace{\rowmargin}
\mpage{\elevenimg}{\includegraphics[width=\linewidth]{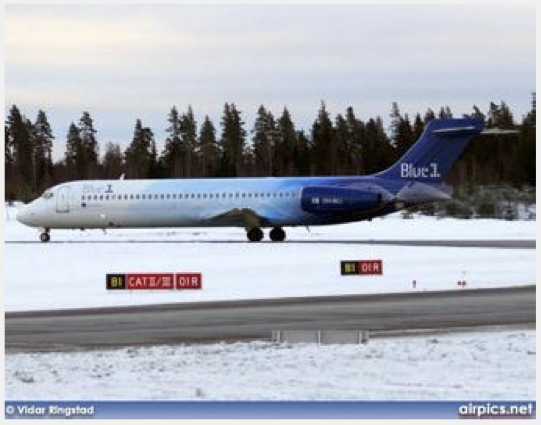}} \hfill
\mpage{\elevenimg}{\includegraphics[width=\linewidth]{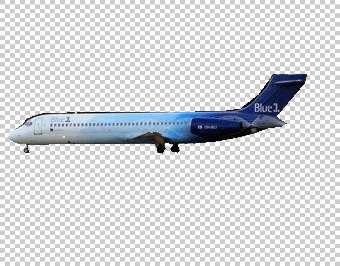}} \hfill
\mpage{\elevenimg}{\includegraphics[width=\linewidth]{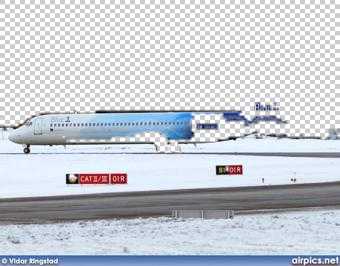}} \hfill
\mpage{\elevenimg}{\includegraphics[width=\linewidth]{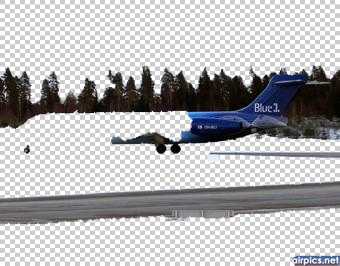}} \hfill
\mpage{\elevenimg}{\includegraphics[width=\linewidth]{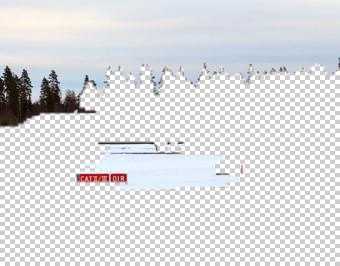}} \hfill
\mpage{\elevenimg}{\includegraphics[width=\linewidth]{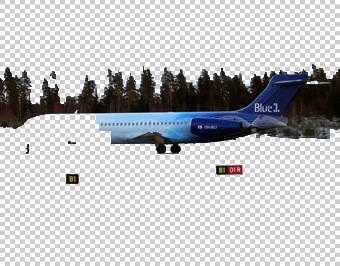}} \hfill
\mpage{\elevenimg}{\includegraphics[width=\linewidth]{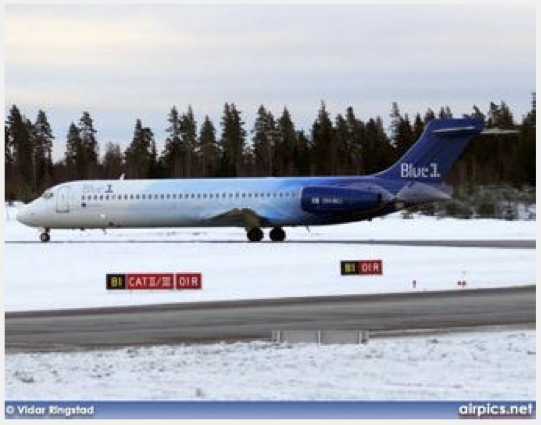}} \hfill
\mpage{\elevenimg}{\includegraphics[width=\linewidth]{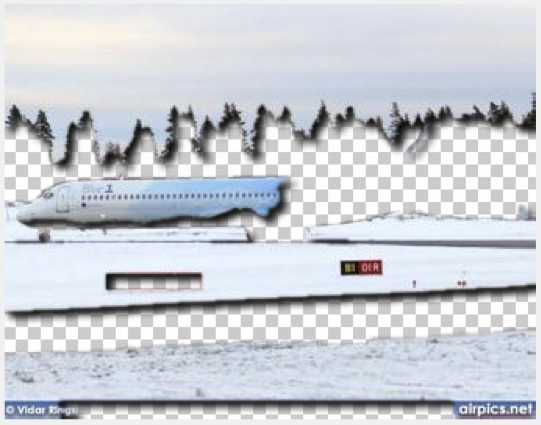}} \hfill
\mpage{\elevenimg}{\includegraphics[width=\linewidth]{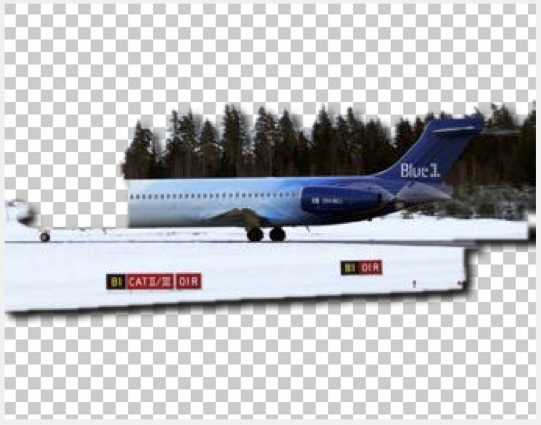}} \hfill
\mpage{\elevenimg}{\includegraphics[width=\linewidth]{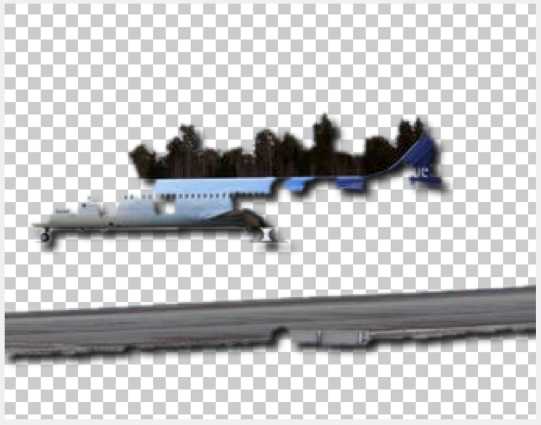}} \hfill
\mpage{\elevenimg}{\includegraphics[width=\linewidth]{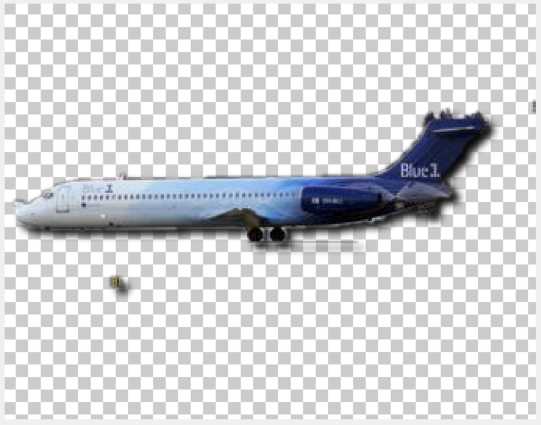}} \\

\vspace{\rowmargin}
\mpage{\elevenimg}{\includegraphics[width=\linewidth]{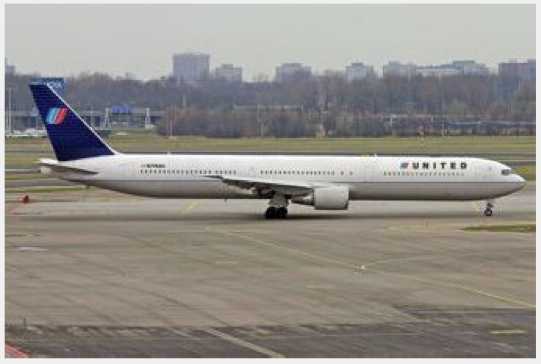}} \hfill
\mpage{\elevenimg}{\includegraphics[width=\linewidth]{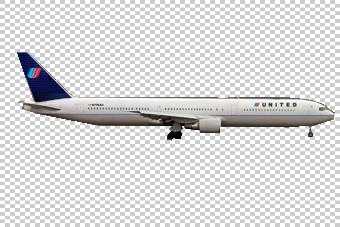}} \hfill
\mpage{\elevenimg}{\includegraphics[width=\linewidth]{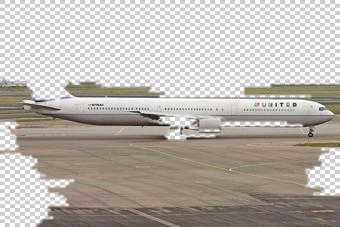}} \hfill
\mpage{\elevenimg}{\includegraphics[width=\linewidth]{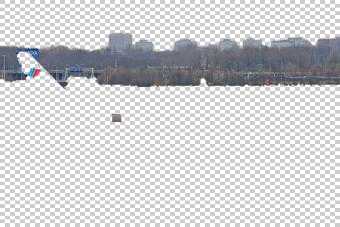}} \hfill
\mpage{\elevenimg}{\includegraphics[width=\linewidth]{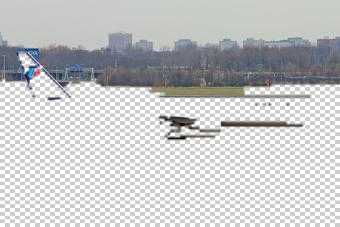}} \hfill
\mpage{\elevenimg}{\includegraphics[width=\linewidth]{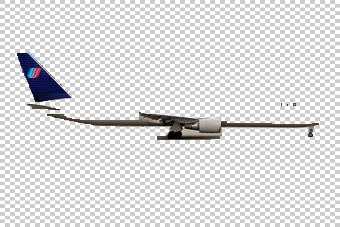}} \hfill
\mpage{\elevenimg}{\includegraphics[width=\linewidth]{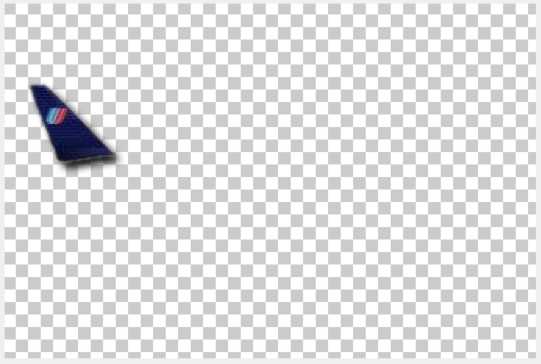}} \hfill
\mpage{\elevenimg}{\includegraphics[width=\linewidth]{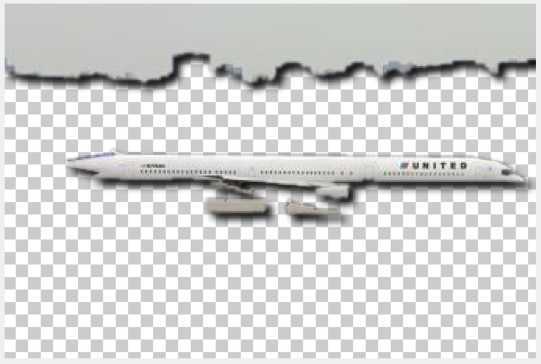}} \hfill
\mpage{\elevenimg}{\includegraphics[width=\linewidth]{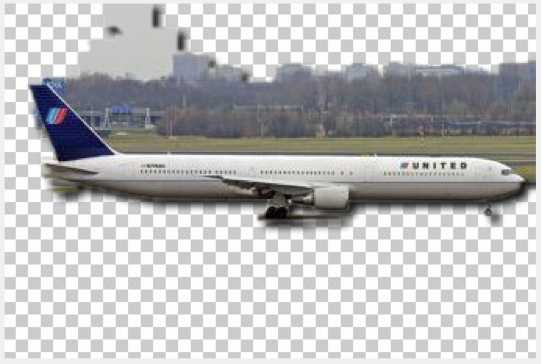}} \hfill
\mpage{\elevenimg}{\includegraphics[width=\linewidth]{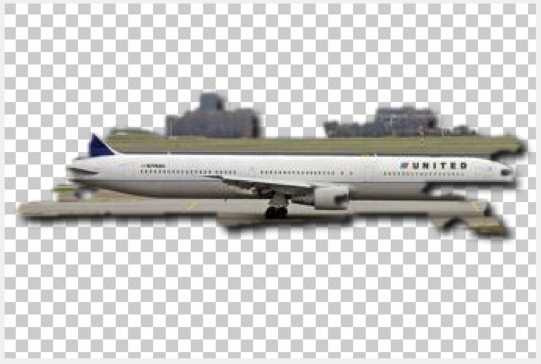}} \hfill
\mpage{\elevenimg}{\includegraphics[width=\linewidth]{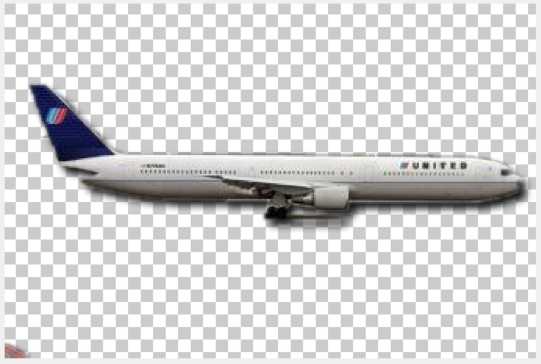}} \\

\vspace{\rowmargin}
\mpage{\elevenimg}{\includegraphics[width=\linewidth]{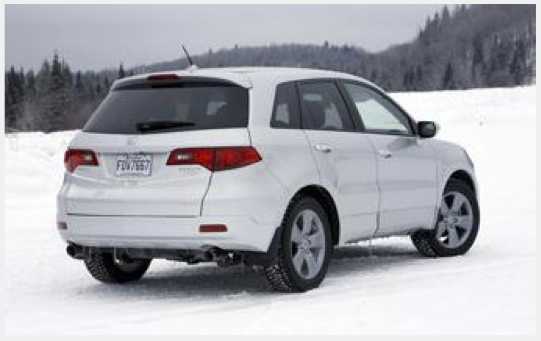}} \hfill
\mpage{\elevenimg}{\includegraphics[width=\linewidth]{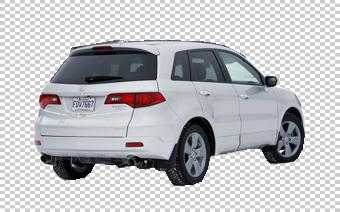}} \hfill
\mpage{\elevenimg}{\includegraphics[width=\linewidth]{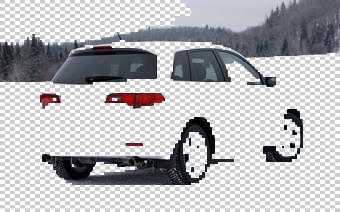}} \hfill
\mpage{\elevenimg}{\includegraphics[width=\linewidth]{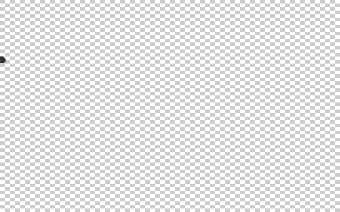}} \hfill
\mpage{\elevenimg}{\includegraphics[width=\linewidth]{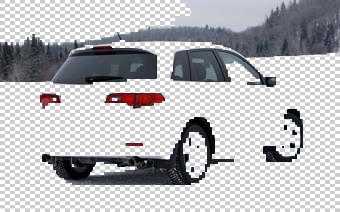}} \hfill
\mpage{\elevenimg}{\includegraphics[width=\linewidth]{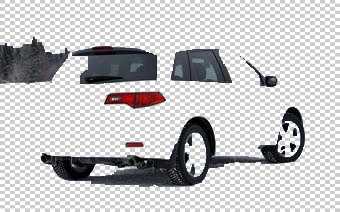}} \hfill
\mpage{\elevenimg}{\includegraphics[width=\linewidth]{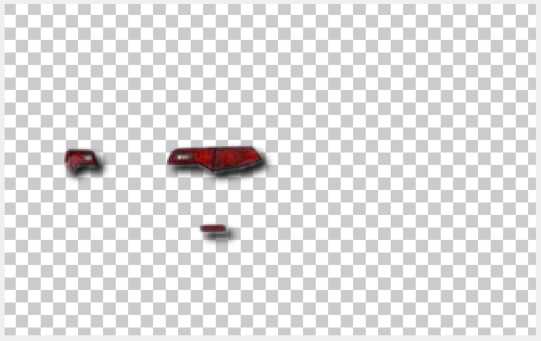}} \hfill
\mpage{\elevenimg}{\includegraphics[width=\linewidth]{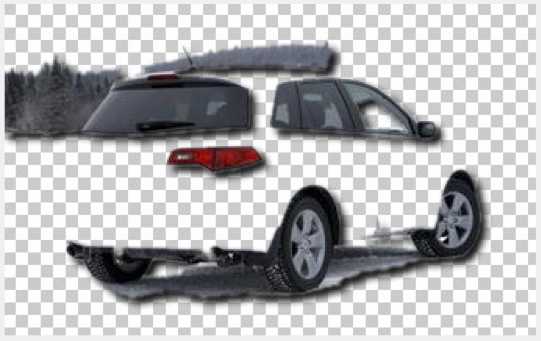}} \hfill
\mpage{\elevenimg}{\includegraphics[width=\linewidth]{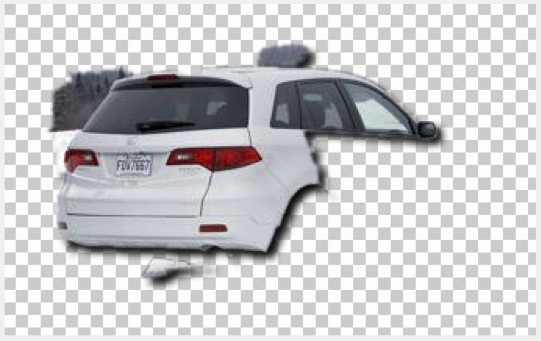}} \hfill
\mpage{\elevenimg}{\includegraphics[width=\linewidth]{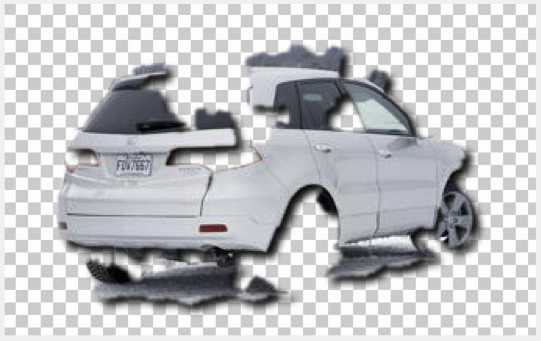}} \hfill
\mpage{\elevenimg}{\includegraphics[width=\linewidth]{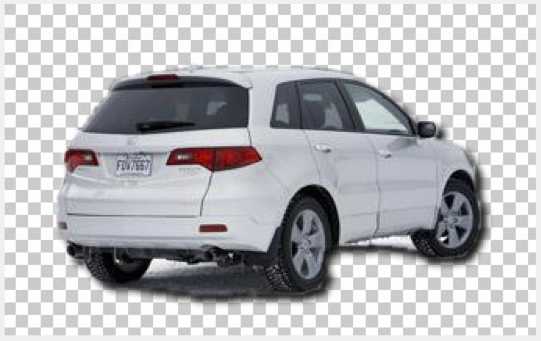}} \\

\vspace{\rowmargin}
\mpage{\elevenimg}{\includegraphics[width=\linewidth]{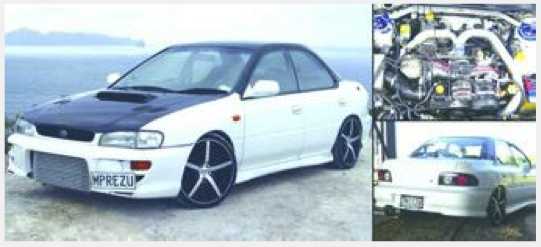}} \hfill
\mpage{\elevenimg}{\includegraphics[width=\linewidth]{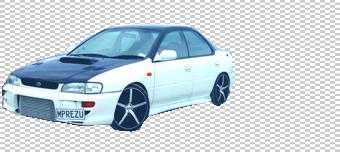}} \hfill
\mpage{\elevenimg}{\includegraphics[width=\linewidth]{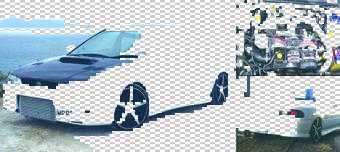}} \hfill
\mpage{\elevenimg}{\includegraphics[width=\linewidth]{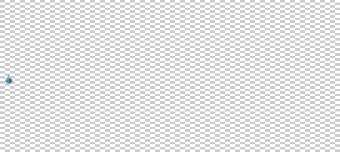}} \hfill
\mpage{\elevenimg}{\includegraphics[width=\linewidth]{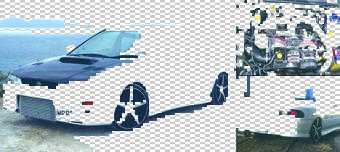}} \hfill
\mpage{\elevenimg}{\includegraphics[width=\linewidth]{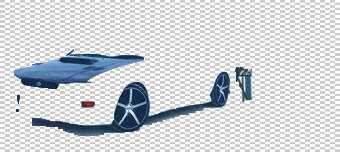}} \hfill
\mpage{\elevenimg}{\includegraphics[width=\linewidth]{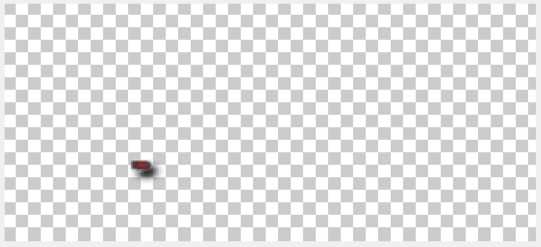}} \hfill
\mpage{\elevenimg}{\includegraphics[width=\linewidth]{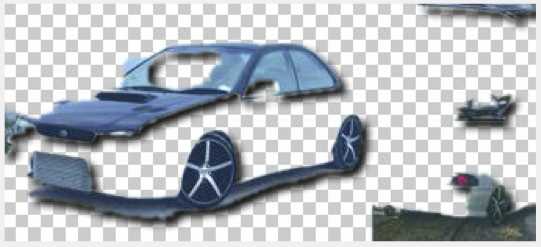}} \hfill
\mpage{\elevenimg}{\includegraphics[width=\linewidth]{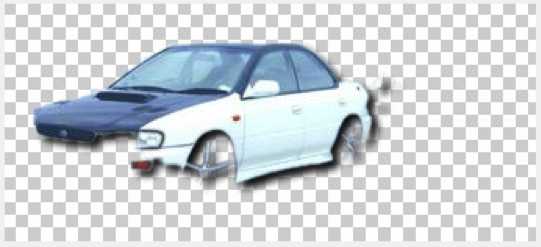}} \hfill
\mpage{\elevenimg}{\includegraphics[width=\linewidth]{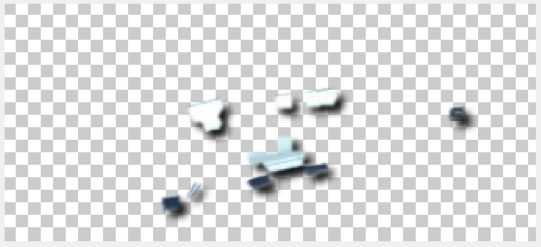}} \hfill
\mpage{\elevenimg}{\includegraphics[width=\linewidth]{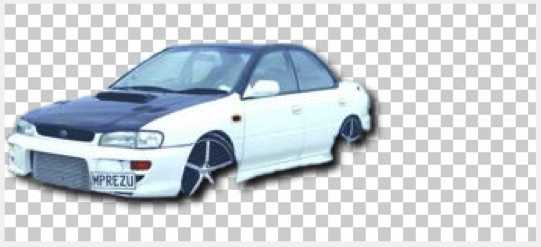}} \\

\vspace{\rowmargin}
\mpage{\elevenimg}{\includegraphics[width=\linewidth]{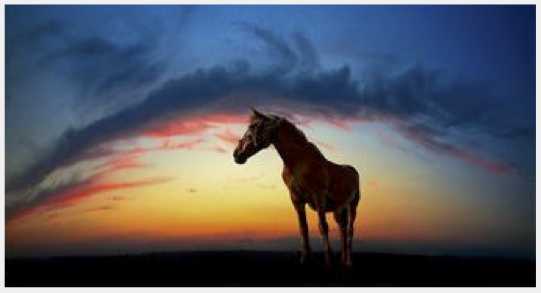}} \hfill
\mpage{\elevenimg}{\includegraphics[width=\linewidth]{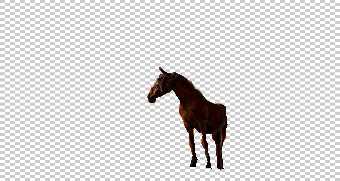}} \hfill
\mpage{\elevenimg}{\includegraphics[width=\linewidth]{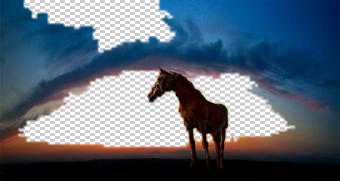}} \hfill
\mpage{\elevenimg}{\includegraphics[width=\linewidth]{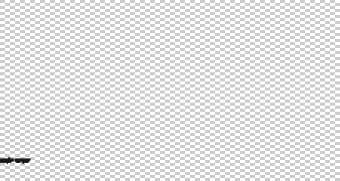}} \hfill
\mpage{\elevenimg}{\includegraphics[width=\linewidth]{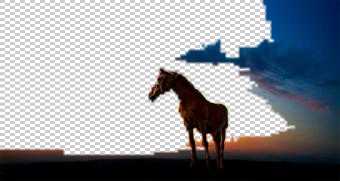}} \hfill
\mpage{\elevenimg}{\includegraphics[width=\linewidth]{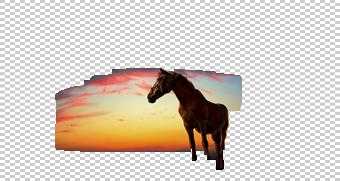}} \hfill
\mpage{\elevenimg}{\includegraphics[width=\linewidth]{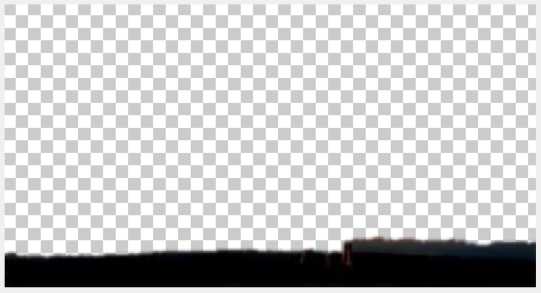}} \hfill
\mpage{\elevenimg}{\includegraphics[width=\linewidth]{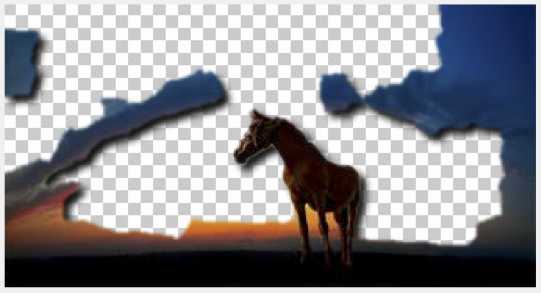}} \hfill
\mpage{\elevenimg}{\includegraphics[width=\linewidth]{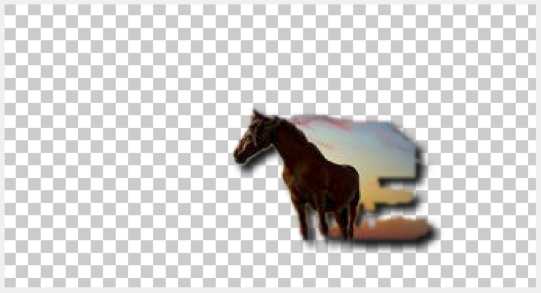}} \hfill
\mpage{\elevenimg}{\includegraphics[width=\linewidth]{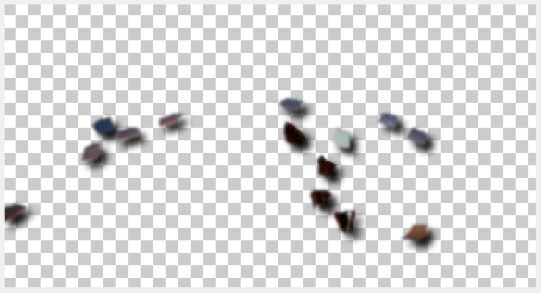}} \hfill
\mpage{\elevenimg}{\includegraphics[width=\linewidth]{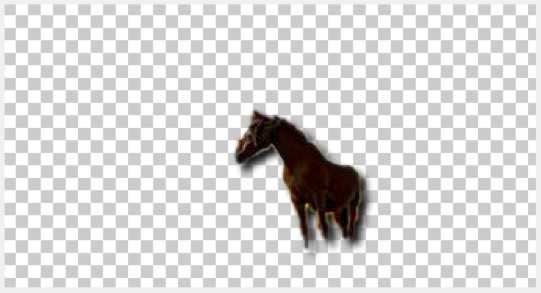}} \\

\vspace{\rowmargin}
\mpage{\elevenimg}{\includegraphics[width=\linewidth]{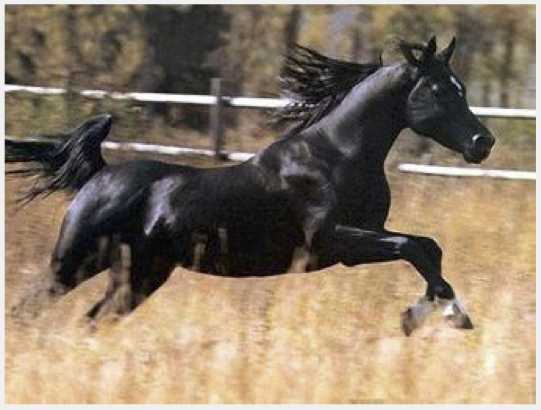}} \hfill
\mpage{\elevenimg}{\includegraphics[width=\linewidth]{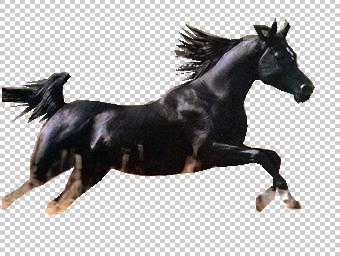}} \hfill
\mpage{\elevenimg}{\includegraphics[width=\linewidth]{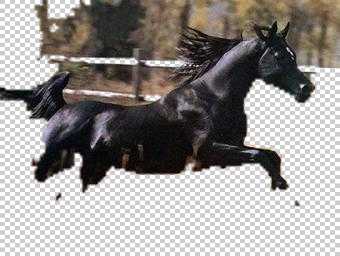}} \hfill
\mpage{\elevenimg}{\includegraphics[width=\linewidth]{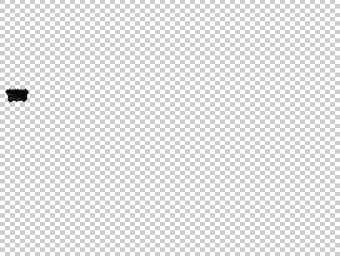}} \hfill
\mpage{\elevenimg}{\includegraphics[width=\linewidth]{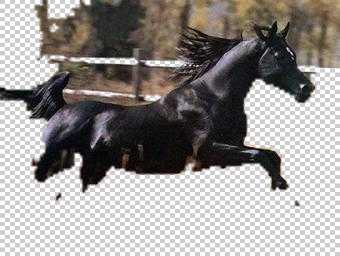}} \hfill
\mpage{\elevenimg}{\includegraphics[width=\linewidth]{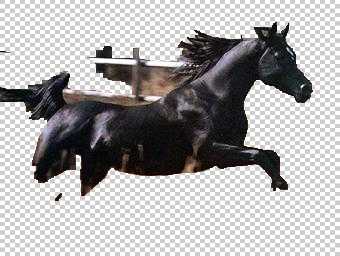}} \hfill
\mpage{\elevenimg}{\includegraphics[width=\linewidth]{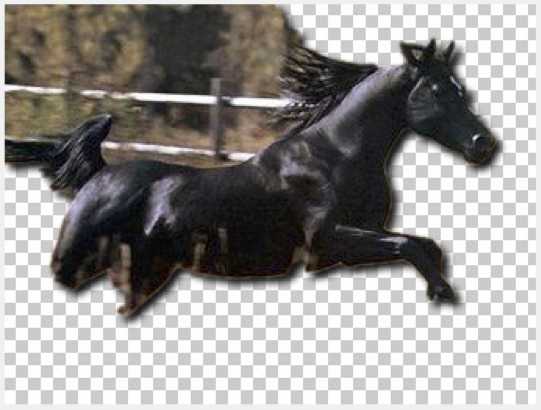}} \hfill
\mpage{\elevenimg}{\includegraphics[width=\linewidth]{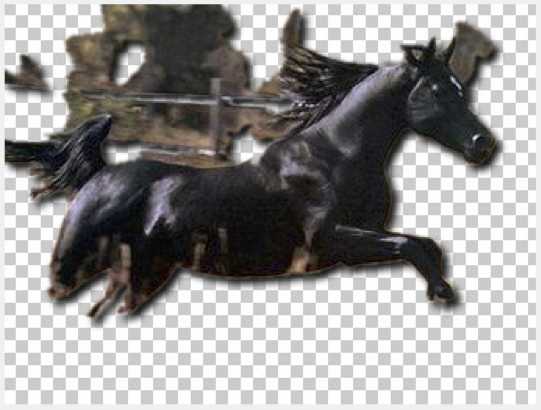}} \hfill
\mpage{\elevenimg}{\includegraphics[width=\linewidth]{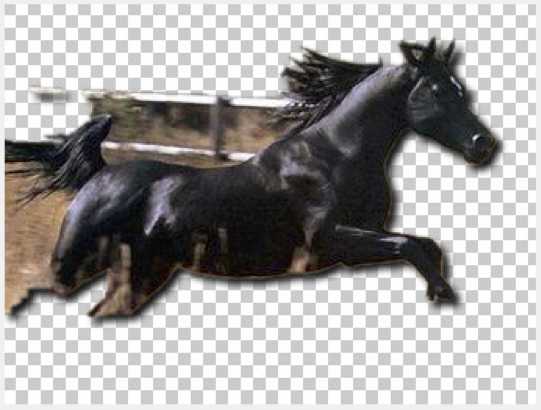}} \hfill
\mpage{\elevenimg}{\includegraphics[width=\linewidth]{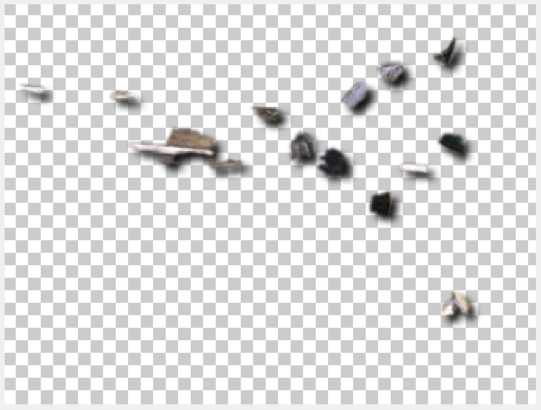}} \hfill
\mpage{\elevenimg}{\includegraphics[width=\linewidth]{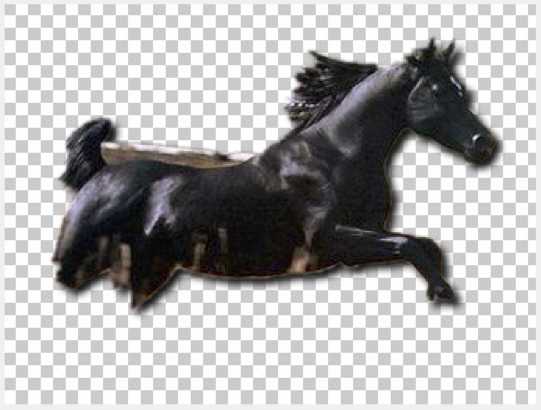}} \\

\mpage{\elevenimg}{\scriptsize Images} \hfill
\mpage{\elevenimg}{\scriptsize GT} \hfill
\mpage{\elevenimg}{\scriptsize \cite{Joulin10}} \hfill
\mpage{\elevenimg}{\scriptsize \cite{kim2011distributed}} \hfill
\mpage{\elevenimg}{\scriptsize \cite{Joulin12}} \hfill
\mpage{\elevenimg}{\scriptsize \cite{rubinstein}} \hfill
\mpage{\elevenimg}{\scriptsize \cite{Chang15}} \hfill
\mpage{\elevenimg}{\scriptsize \cite{Lee15}} \hfill
\mpage{\elevenimg}{\scriptsize \cite{Jerripothula17}} \hfill
\mpage{\elevenimg}{\scriptsize \cite{Hati16}} \hfill
\mpage{\elevenimg}{\scriptsize Ours} \\

\caption{\textbf{Qualitative results of object co-segmentation on the Internet~\cite{rubinstein} dataset.} Our method is capable of delineating accurate co-occurring object masks under large intra-class variations and background clutter.
}

\label{fig:internet-coseg}

\end{center}
\end{figure*}

\subsection{Object co-segmentation}
Table~\ref{table:Coseg} reports the quantitative results on the challenging Internet dataset~\cite{rubinstein}.
In this experiment, we set the hyper-parameters as follows: $\lambda_\mathrm{cycle}=5$, $\lambda_\mathrm{trans}=5$, $\lambda_\mathrm{contrast}=20$, and $\lambda_\mathrm{task}=10$.
%
Our results show that our method compares favorably against existing weakly-supervised methods and achieves competitive performance when compared with a strongly-supervised approach~\cite{li2018deep}.
%
%
The performance gain over the best competitor under the same experimental setting~\cite{hsu2018co} is $2.6\%$ in precision and $2\%$ in Jaccard index.
%
Our results demonstrate that our method is capable of adapting itself well to unseen object categories by training with weak image-level supervision provided by the dataset.
Figure~\ref{fig:internet-coseg} presents the visual comparisons of object co-segmentation with existing methods.
From the visual results, we observe that our method is more robust to intra-class appearance variations and viewpoint changes, and produces more accurate and consistent object co-segmentation results when compared with existing methods.

\setlength{\eightimg}{0.11\textwidth}

\setlength{\rowmargin}{0.5mm}

\begin{figure*}[t]
\begin{center}

\mpage{\eightimg}{\includegraphics[width=\linewidth]{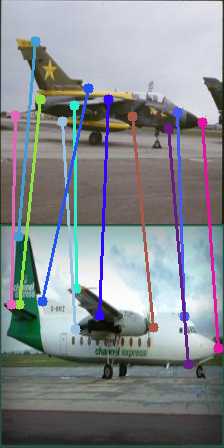}} 
\mpage{\eightimg}{\includegraphics[width=\linewidth]{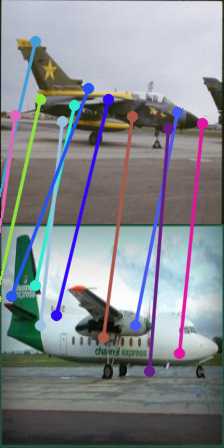}} 
\mpage{\eightimg}{\includegraphics[width=\linewidth]{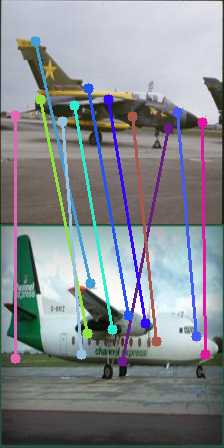}} 
\mpage{\eightimg}{\includegraphics[width=\linewidth]{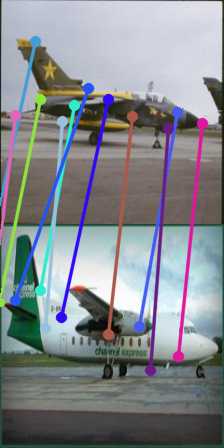}} \hfill
\mpage{\eightimg}{\includegraphics[width=\linewidth]{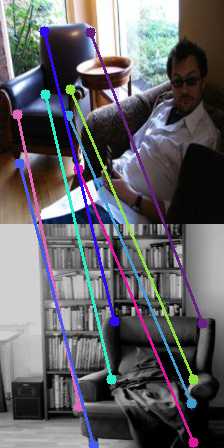}} 
\mpage{\eightimg}{\includegraphics[width=\linewidth]{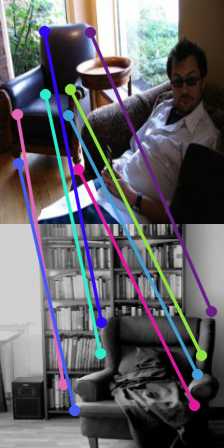}} 
\mpage{\eightimg}{\includegraphics[width=\linewidth]{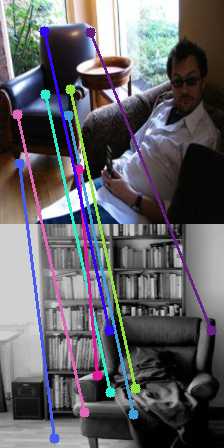}} 
\mpage{\eightimg}{\includegraphics[width=\linewidth]{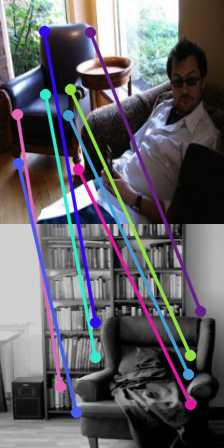}} \\

\vspace{\rowmargin}
\mpage{\eightimg}{\includegraphics[width=\linewidth]{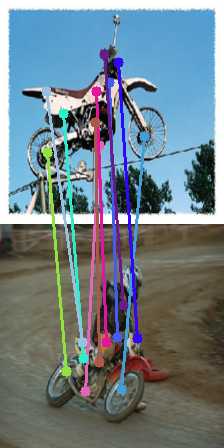}} 
\mpage{\eightimg}{\includegraphics[width=\linewidth]{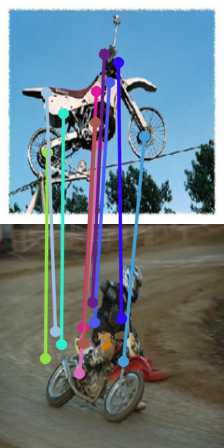}} 
\mpage{\eightimg}{\includegraphics[width=\linewidth]{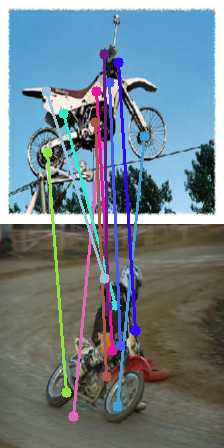}} 
\mpage{\eightimg}{\includegraphics[width=\linewidth]{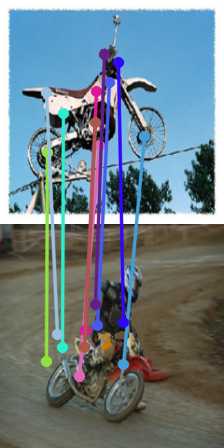}} \hfill
\mpage{\eightimg}{\includegraphics[width=\linewidth]{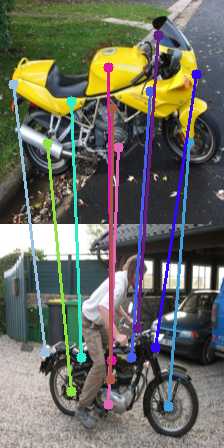}} 
\mpage{\eightimg}{\includegraphics[width=\linewidth]{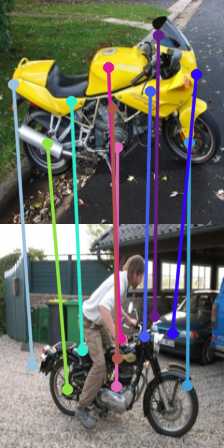}} 
\mpage{\eightimg}{\includegraphics[width=\linewidth]{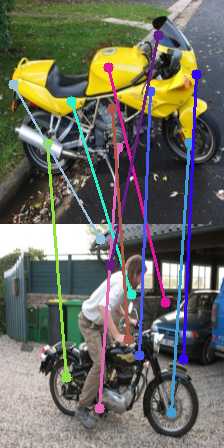}} 
\mpage{\eightimg}{\includegraphics[width=\linewidth]{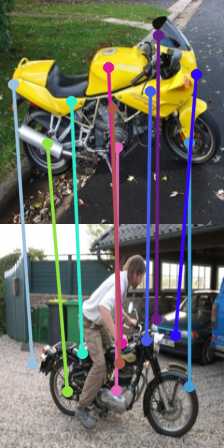}} \\

\mpage{\eightimg}{\scriptsize Ground truth} 
\mpage{\eightimg}{\scriptsize WeakMatchNet~\cite{WeakMatchNet}}
\mpage{\eightimg}{\scriptsize NC-Net\cite{ncnet}} 
\mpage{\eightimg}{\scriptsize Ours} \hfill
\mpage{\eightimg}{\scriptsize Ground truth} 
\mpage{\eightimg}{\scriptsize WeakMatchNet~\cite{WeakMatchNet}}
\mpage{\eightimg}{\scriptsize NC-Net\cite{ncnet}} 
\mpage{\eightimg}{\scriptsize Ours} \\

\captionof{figure}{\textbf{Qualitative results of semantic matching.} 
We present the qualitative comparisons of semantic matching with the state-of-the-art algorithms on the PF-PASCAL~\cite{ProposalFlow} (\emph{top} row) and PF-WILLOW~\cite{ProposalFlow} (\emph{bottom} row) datasets.}

\label{fig:pf-match}

\end{center}

\end{figure*}

\subsection{Semantic matching}
To evaluate the proposed method on semantic matching, we conduct experiments on the PF-PASCAL~\cite{ProposalFlow}, PF-WILLOW~\cite{ProposalFlow}, and SPair-$71$k~\cite{min2019spair} datasets.

\begin{table*}[t]
  \scriptsize
  \ra{1.2}
  \begin{center}
  \caption{
  \textbf{Experimental results of semantic matching on the PF-PASCAL dataset~\cite{ProposalFlow}.}
  The bold and underlined numbers indicate the top two results, respectively.}
  \label{table:Matching-PF-PASCAL}
  \resizebox{\linewidth}{!}
  {
  \begin{tabular}{l|c|cccccccccccccccccccc|c}
  \toprule
  Method & Descriptor & aero & bike & bird & boat & bottle & bus & car & cat & chair & cow & d.table & dog & horse & moto & person & plant & sheep & sofa & train & tv & mean\\
  \midrule
  Proposal Flow+LOM~\cite{ProposalFlow} & HOG~\cite{HoG} & 73.3 & 74.4 & 54.4 & 50.9 & 49.6 & 73.8 & 72.9 & 63.6 & 46.1 & 79.8 & 42.5 & 48.0 & 68.3 & 66.3 & 42.1 & 62.1 & 65.2 & 57.1 & 64.4 & 58.0 & 62.5\\
  UCN~\cite{UCN} & GoogLeNet~\cite{szegedy2015going} & 64.8 & 58.7 & 42.8 & 59.6 & 47.0 & 42.2 & 61.0 & 45.6 & 49.9 & 52.0 & 48.5 & 49.5 & 53.2 & 72.7 & 53.0 & 41.4 & \underline{83.3} & 49.0 & \underline{73.0} & 66.0 & 55.6\\
  \revised{A2Net~\cite{A2Net}} & ResNet-101~\cite{ResNet} & - & - & - & - & - & - & - & - & - & - & - & - & - & - & - & - & - & - & - & - & 59.0 \\
  \revised{GSF~\cite{novotny2018self}} & ResNet-50~\cite{ResNet} & - & - & - & - & - & - & - & - & - & - & - & - & - & - & - & - & - & - & - & - & 66.5 \\
  SCNet-A~\cite{SCNet} & VGG-16~\cite{VGG} & 67.6 & 72.9 & 69.3 & 59.7 & 74.5 & 72.7 & 73.2 & 59.5 & 51.4 & 78.2 & 39.4 & 50.1 & 67.0 & 62.1 & 69.3 & 68.5 & 78.2 & 63.3 & 57.7 & 59.8 & 66.3\\
  SCNet-AG~\cite{SCNet} & VGG-16~\cite{VGG} & 83.9 & 81.4 & 70.6 & 62.5 & 60.6 & 81.3 & 81.2 & 59.5 & 53.1 & 81.2 & \textbf{62.0} & 58.7 & 65.5 & 73.3 & 51.2 & 58.3 & 60.0 & \underline{69.3} & 61.5 & \underline{80.0} & 69.7\\
  SCNet-AG+~\cite{SCNet} & VGG-16~\cite{VGG} & 85.5 & 84.4 & 66.3 & 70.8 & 57.4 & 82.7 & 82.3 & 71.6 & \underline{54.3} & \textbf{95.8} & \underline{55.2} & 59.5 & \underline{68.6} & 75.0 & 56.3 & 60.4 & 60.0 & \textbf{73.7} & 66.5 & 76.7 & 72.2\\
  \revised{CNNGeo~\cite{CNNGeo}} & VGG-16~\cite{VGG} & 79.5 & 80.9 & 69.9 & 61.1 & 57.8 & 77.1 & 84.4 & 55.5 & 48.1 & 83.3 & 37.0 & 54.1 & 58.2 & 70.7 & 51.4 & 41.4 & 60.0 & 44.3 & 55.3 & 30.0 & 62.6\\
  \revised{CNNGeo~\cite{CNNGeo}} & ResNet-101~\cite{ResNet} & 83.0 & 82.2 & 81.1 & 50.0 & 57.8 & 79.9 & 92.8 & 77.5 & 44.7 & 85.4 & 28.1 & 69.8 & 65.4 & 77.1 & 64.0 & 65.2 & \textbf{100.0} & 50.8 & 44.3 & 54.4 & 69.5\\
  CNNGeo w/ Inlier~\cite{End-to-end} & ResNet-101~\cite{ResNet} & 84.7 & \underline{88.9} & 80.9 & 55.6 & 76.6 & 89.5 & \underline{93.9} & 79.6 & 52.0 & 85.4 & 28.1 & 71.8 & 67.0 & 75.1 & 66.3 & 70.5 & \textbf{100.0} & 62.1 & 62.3 & 61.1 & 74.8\\
  NC-Net~\cite{ncnet} & ResNet-101~\cite{ResNet} & \textbf{86.8} & 86.7 & \textbf{86.7} & 55.6 & \underline{82.8} & 88.6 & 93.8 & \textbf{87.1} & \underline{54.3} & 87.5 & 43.2 & \textbf{82.0} & 64.1 & \textbf{79.2} & \underline{71.1} & \underline{71.0} & 60.0 & 54.2 & \textbf{75.0} & \textbf{82.8} & \underline{78.9} \\
  WeakMatchNet~\cite{WeakMatchNet} & ResNet-101~\cite{ResNet} & \underline{85.6} & \textbf{89.6} & 82.1 & \textbf{83.3} & \textbf{85.9} & \underline{92.5} & \underline{93.9} & 80.2 & 52.2 & 85.4 & \underline{55.2} & 75.2 & 64.0 & \underline{77.9} & 67.2 & \textbf{73.8} & \textbf{100.0} & 65.3 & 69.3 & 61.1 & 78.0 \\
  %
  %
  Ours & ResNet-101~\cite{ResNet} & 83.4 & 87.4 & \underline{85.3} & \underline{72.2} & 76.6 & \textbf{94.6} & \textbf{94.7} & \underline{86.6} & \textbf{54.9} & \underline{89.6} & 52.6 & \underline{80.2} & \textbf{70.6} & \textbf{79.2} & \textbf{73.3} & 70.5 & \textbf{100.0} & 63.0 & 66.3 & 64.4 & \textbf{79.0} \\
  \bottomrule
  \end{tabular}
  }
  \end{center}
\end{table*}

{\flushleft {\bf Results on the PF-PASCAL dataset.}} 
Table~\ref{table:Matching-PF-PASCAL} shows the quantitative results of semantic matching on the PF-PASCAL~\cite{ProposalFlow} dataset.
In this experiment, we set the hyper-parameters as follows: $\lambda_\mathrm{cycle}=20$, $\lambda_\mathrm{trans}=10$, $\lambda_\mathrm{contrast}=2.5$, and $\lambda_\mathrm{task}=2.5$.
The proposed approach performs favorably against the state-of-the-art methods, achieving an overall PCK of $79.0\%$.
The advantage of integrating object co-segmentation over performing foreground detection on the feature maps can be assessed by comparing the proposed method with the WeakMatchNet~\cite{WeakMatchNet}.
The proposed method improves the performance by $1.0\%$ in terms of PCK evaluated at $\alpha=0.1$.
The top row of Figure~\ref{fig:pf-match} shows semantic matching results of the evaluated methods.
Estimating geometric transformations leads to more geometrically consistent matching results than approaches that establish correspondences without using any geometric transformation models (\eg NC-Net~\cite{ncnet}).
%
%

\begin{table}[t]
  \scriptsize
  \caption{
  \textbf{Experimental results of semantic matching on the PF-WILLOW dataset~\cite{ProposalFlow}.}
  The bold and underlined numbers indicate the top two results, respectively.}
  \label{table:Matching-PF-WILLOW}
  \centering
  \resizebox{\linewidth}{!} 
  {
  \begin{tabular}{l|c|c|c|c}
  \toprule
  Method & Descriptor & $\alpha$ = 0.05 & $\alpha$ = 0.1 & $\alpha$ = 0.15 \\
  \midrule
  SIFT Flow~\cite{SIFTFlow} & SIFT~\cite{SIFT} & 0.247 & 0.380 & 0.504 \\
  %
  %
  SIFT Flow~\cite{SIFTFlow} & VGG-16~\cite{VGG} & 0.324 & 0.456 & 0.555 \\
  \revised{CNNGeo~\cite{CNNGeo}} & ResNet-101~\cite{ResNet} & 0.448 & 0.777 & 0.899 \\
  CNNGeo w/ Inlier~\cite{End-to-end} & ResNet-101~\cite{ResNet} & 0.477 & 0.812 & 0.917 \\
  Proposal Flow + LOM~\cite{ProposalFlow} & HOG~\cite{HoG} & 0.284 & 0.568 & 0.682 \\
  UCN~\cite{UCN} & GoogLeNet~\cite{szegedy2015going} & 0.291 & 0.417 & 0.513 \\
  SCNet-A~\cite{SIFTFlow} & VGG-16~\cite{VGG} & 0.390 & 0.725 & 0.873 \\
  SCNet-AG~\cite{SIFTFlow} & VGG-16~\cite{VGG} & 0.394 & 0.721 & 0.871 \\
  SCNet-AG+~\cite{SIFTFlow} & VGG-16~\cite{VGG} & 0.386 & 0.704 & 0.853 \\
  \revised{A2Net~\cite{A2Net}} & ResNet-101~\cite{ResNet} & - & 0.680 & - \\
  WeakMatchNet~\cite{WeakMatchNet} & ResNet-101~\cite{ResNet} & 0.484 & 0.816 & 0.918 \\
  %
  %
  RTNs~\cite{RTN} & ResNet-101~\cite{ResNet} & 0.413 & 0.719 & 0.862 \\
  NC-Net~\cite{ncnet} & ResNet-101~\cite{ResNet} & \underline{0.514} & \underline{0.818} & \underline{0.927} \\
  %
  %
  Ours & ResNet-101~\cite{ResNet} & \textbf{0.538} & \textbf{0.854} & \textbf{0.939} \\
  \bottomrule
  \end{tabular}
  }
\end{table}

{\flushleft {\bf Results on the PF-WILLOW dataset.}} 
To evaluate the generalization capability of the proposed method, we evaluate the proposed model trained on the PF-PASCAL dataset~\cite{ProposalFlow} to the PF-WILLOW dataset~\cite{ProposalFlow} without fine-tuning.
Table~\ref{table:Matching-PF-WILLOW} shows the quantitative results of semantic matching on the PF-WILLOW~\cite{ProposalFlow} dataset.
Our method performs favorably against existing methods on all three evaluated PCK thresholds.
The performance gain over the second best method~\cite{ncnet} is $2.0\%$ at $\alpha = 0.05$ or $3.6\%$ at $\alpha = 0.1$.
The results suggest that sufficient generalization ability in establishing dense correspondences can be exhibited by our model.
The bottom row of Figure~\ref{fig:pf-match} shows two examples of visual results of the evaluated methods. 
The matching results by our method are more accurate and geometrically consistent.

\begin{table}[!ht]
  \caption{
  \textbf{Experimental results of semantic matching on the SPair-71k dataset~\cite{min2019spair}.} 
  Marker $^*$ denotes that the method requires keypoint supervision.
  The column Fine-tune denotes that method is fine-tuned on the SPair-71k training set~\cite{min2019spair}.
  The bold and underlined numbers indicate the top two results, respectively.
  }
  \scriptsize
  \label{table:spair}
  \begin{center}
    \begin{tabular}{l|c|c}
    \toprule
    Method & Fine-tune & Avg. \\
    \midrule
    HPF~\cite{min2019hyperpixel}$^*$ & $\checkmark$ & 28.2 \\
    \midrule
    CNNGeo~\cite{CNNGeo} & & 18.1 \\
    A2Net~\cite{A2Net} & & 20.1 \\
    CNNGeo w/ Inlier~\cite{End-to-end} & & 21.1 \\
    NC-Net~\cite{ncnet} & & \underline{26.4} \\
    Ours & & 25.8 \\
    \midrule
    CNNGeo~\cite{CNNGeo} & $\checkmark$ & 20.6 \\
    A2Net~\cite{A2Net} & $\checkmark$ & 22.3 \\
    CNNGeo w/ Inlier~\cite{End-to-end} & $\checkmark$ & 20.9 \\
    NC-Net~\cite{ncnet} & $\checkmark$ & 20.1 \\
    Ours & $\checkmark$ & \textbf{26.6} \\
    \bottomrule
    \end{tabular}
  \end{center}
\end{table}

\revised{
{\flushleft {\bf Results on the SPair-$71$k dataset.}} 
Table~\ref{table:spair} presents the quantitative results on the challenging SPair-$71$k dataset~\cite{min2019spair}.
Following \cite{min2019hyperpixel}, we report two different sets of results: 
(1) Training on the PF-PASCAL dataset~\cite{ProposalFlow} and 
(2) Training on the the SPair-71k training set~\cite{min2019spair}.
In the second experiment, we set the hyper-parameters as follows: $\lambda_\mathrm{cycle}=20$, $\lambda_\mathrm{trans}=20$, $\lambda_\mathrm{contrast}=1$, and $\lambda_\mathrm{task}=1$.
Our results show that our method compares favorably against existing weakly-supervised methods and achieves competitive performance when compared with an approach that requires keypoint annotations as supervision~\cite{min2019hyperpixel}.
}

\begin{table}[t]
    \scriptsize
    \caption{
    \textbf{Ablation studies of object co-segmentation on the Internet dataset~\cite{rubinstein}. }
    %
    %
    The bold and underlined numbers indicate the top two results.
    }
    \label{exp:abla-coseg-internet}
    \centering
    \resizebox{\linewidth}{!} 
    {
    \begin{tabular}{l|cc|cc|cc|cc}
    \toprule
    \multirow{2}{*}{Method} & \multicolumn{2}{c|}{Airplane} & \multicolumn{2}{c|}{Car} & \multicolumn{2}{c|}{Horse} & \multicolumn{2}{c}{Avg.} \\
    & $\mathcal{P}$ & $\mathcal{J}$ & $\mathcal{P}$ & $\mathcal{J}$ & $\mathcal{P}$ & $\mathcal{J}$ & $\mathcal{P}$ & $\mathcal{J}$ \\
    \midrule
    Ours (full model) & \textbf{0.941} & \textbf{0.65} & \textbf{0.940} & \textbf{0.82} & \textbf{0.922} & \textbf{0.63} & \textbf{0.935} & \textbf{0.70} \\
    %
    %
    Ours w/o $\mathcal{L}_\mathrm{matching}$ & \underline{0.940} & \underline{0.64} & 0.927 & 0.78 & 0.887 & 0.61 & 0.918 & 0.68 \\
    Ours w/o $\mathcal{L}_\mathrm{cycle-consis}$ & 0.936 & \underline{0.64} & \underline{0.929} & \underline{0.81} & \underline{0.920} & \underline{0.62} & 0.928 & \underline{0.69} \\
    Ours w/o $\mathcal{L}_\mathrm{trans-consis}$ & 0.939 & \textbf{0.65} & 0.928 & \underline{0.81} & \textbf{0.922} & \underline{0.62} & \underline{0.930} & \underline{0.69} \\
    Ours w/o $\mathcal{L}_\mathrm{task-consis}$ & 0.938 & \textbf{0.65} & 0.915 & 0.78 & 0.883 & 0.60 & 0.912 & 0.68 \\
    Ours w/o $\mathcal{L}_\mathrm{contrast}$ & 0.444 & 0.39 & 0.302 & 0.26 & 0.475 & 0.43 & 0.425 & 0.38 \\
    \bottomrule
    \end{tabular}
    }
  \end{table}

\subsection{Ablation study}

{\flushleft {\bf Removing one loss at a time.}} 
To analyze the importance of each adopted loss function, we conduct an ablation study by turning off one of the loss terms at a time.
For object co-segmentation, we carry out experiments on the Internet dataset~\cite{rubinstein}.
Table~\ref{exp:abla-coseg-internet} shows the experimental results.
For semantic matching, we train our model on the PF-PASCAL dataset~\cite{ProposalFlow} and turn off a loss function at a time.
We report the performance of semantic matching on the PF-WILLOW dataset~\cite{ProposalFlow} recorded at three different PCK thresholds, \ie $\alpha \in \{0.05, 0.1, 0.15\}$.
This ablation study shows the effect of each loss function with respect to the generalization capability.
Table~\ref{exp:abla-match-pf-willow} reports the experimental results. 

For semantic matching, without the foreground-guided matching loss $\mathcal{L}_\mathrm{matching}$, there is no explicit supervision to maximize the similarity between the corresponding features of an image pair.
Thus, our model suffers from a significant performance drop by $4.4\%$ at $\alpha = 0.05$. 
For object co-segmentation, while the performance drops are moderate when the foreground-guided matching loss $\mathcal{L}_\mathrm{matching}$ is turned off, the results indicate that having better ability in predicting geometric transformations can further improves object co-segmentation.

Without the forward-backward consistency loss $\mathcal{L}_\mathrm{cycle-consis}$, our model only enforces the consistency across multiple images through the transitivity consistency loss $\mathcal{L}_\mathrm{trans-consis}$.
Experimental results show that the performance drops by $0.9\%$ at $\alpha = 0.05$ for semantic matching, and $0.7\%$ in precision and $1\%$ in Jaccard index for object co-segmentation.

Without the transitivity consistency loss $\mathcal{L}_\mathrm{trans-consis}$, our model only enforces the consistency on the estimated geometric transformations between an image pair (\ie the forward-backward consistency loss $\mathcal{L}_\mathrm{cycle-consis}$ is still in effect).
Experimental results show that the performance drops by $0.6\%$ at $\alpha = 0.05$ for semantic matching, and $0.5\%$ in precision and $1\%$ in Jaccard index for object co-segmentation.

Without the perceptual contrastive loss $\mathcal{L}_\mathrm{contrast}$, there is no other loss to explicitly guide the object co-segmentation network stream to predict object masks.
For object co-segmentation, significant performance drops of $51\%$ in precision and $32\%$ in Jaccard index occur since our model no longer segments the co-occurrent objects in an image collection even though the cross-network consistency loss $\mathcal{L}_\mathrm{task-consis}$ facilitates supervision (\ie dense correspondence fields) for the output of the decoder $\mathcal{D}$.
For semantic matching, since our model does not learn to perform object co-segmentation, our model suffers from the negative impact caused by background clutters, resulting in a $3.6\%$ performance drop at $\alpha = 0.05$.

When the cross-network consistency loss $\mathcal{L}_\mathrm{task-consis}$ is turned off, there is no explicit supervision to enforce the predicted object masks to be geometrically consistent across images.
For object co-segmentation, the model thus tends to segment only the discriminative parts of the objects as reflected in Figure~\ref{Fig:TSS-Coseg}, resulting in performance drops of $2.3\%$ in precision and $2\%$ in Jaccard index.
For semantic matching, since the predicted object masks may not precisely highlight the entire objects, our model may not effectively suppress the impact caused by background clutters when incorporating such object masks.
A $2.4\%$ performance drop by our method occurs when $\alpha$ is set to $0.05$.

\begin{table}[t]
    \scriptsize
    \caption{
    \textbf{Ablation study of semantic matching on the PF-WILLOW dataset~\cite{ProposalFlow} under three different PCK thresholds $\alpha$.} 
    %
    %
    The bold and underlined numbers indicate the top two results.
    }
    \label{exp:abla-match-pf-willow}
    \centering
    {
    \begin{tabular}{l|c|c|c}
    \toprule
    Method & $\alpha = 0.05$ & $\alpha = 0.10$ & $\alpha = 0.15$ \\
    \midrule
    %
    %
    %
    Ours (full model)  & \textbf{0.538} & \textbf{0.854} & \textbf{0.939} \\
    Ours w/o $\mathcal{L}_\mathrm{matching}$ & 0.494 & 0.822 & 0.927 \\
    Ours w/o $\mathcal{L}_\mathrm{cycle-consis}$ & 0.529 & 0.847 & \underline{0.938} \\
    Ours w/o $\mathcal{L}_\mathrm{trans-consis}$ & \underline{0.532} & \underline{0.851} & 0.930 \\
    Ours w/o $\mathcal{L}_\mathrm{task-consis}$ & 0.514 & 0.842 & 0.928 \\
    Ours w/o $\mathcal{L}_\mathrm{contrast}$ & 0.502 & 0.823 & 0.922 \\
    %
    %
     WeakMatchNet~\cite{WeakMatchNet} & 0.491 & 0.819 & 0.922 \\
    \bottomrule
    \end{tabular}
    }
  \end{table}

\begin{table}[!t]
  \caption{
  \textbf{Ablation study of object co-segmentation on the TSS dataset~\cite{Taniai} using different post-processing methods.}
  The bold and underlined numbers indicate the top two results, respectively.
  }
  \scriptsize
  \label{table:coseg-post-process}
  \begin{center}
    \begin{tabular}{l|cc}
    \toprule
    Method & $\mathcal{P}$ & $\mathcal{J}$ \\
    \midrule
    DenseCRF~\cite{krahenbuhl2011efficient} & \underline{0.933} & \underline{0.68} \\
    Otsu's~\cite{otsu1979threshold} & 0.927 & 0.65 \\
    GrabCut~\cite{rother2004grabcut} & \textbf{0.935} & \textbf{0.70} \\
    \bottomrule
    \end{tabular}
  \end{center}
\end{table}

\begin{figure}[!ht]
  \centering
  \includegraphics[width=\linewidth]{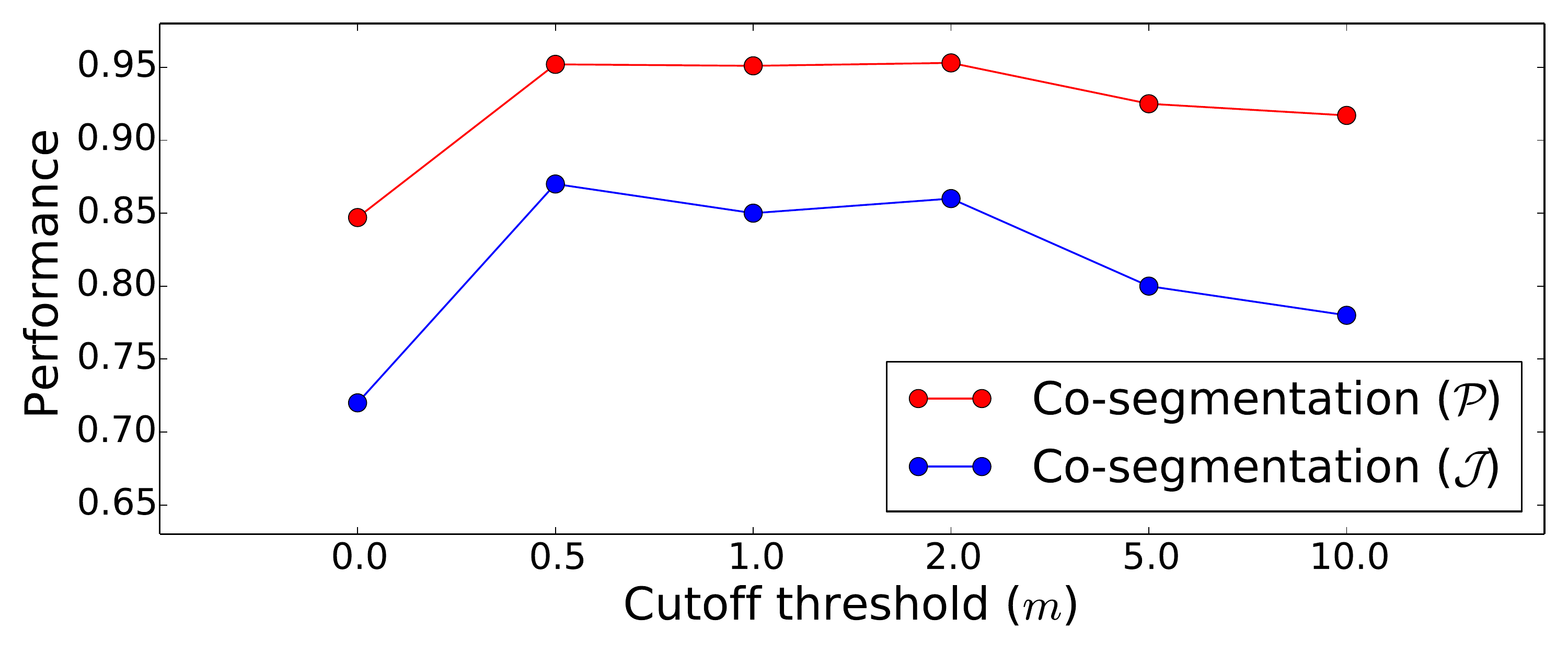}
  \caption{
  \textbf{Sensitivity analysis of the cutoff threshold $m$ on object co-segmentation on the TSS dataset~\cite{Taniai}.}
  The performance of our approach remains stable when the cutoff threshold lies within a reasonable range.
  }
  \label{exp:sensitivity-margin}
\end{figure}


\setlength{\threeimg}{0.33\textwidth}
\begin{figure*}[t!]
  \centering
  \begin{subfigure}[!t]{\threeimg}
    \includegraphics[width=1.0\linewidth]{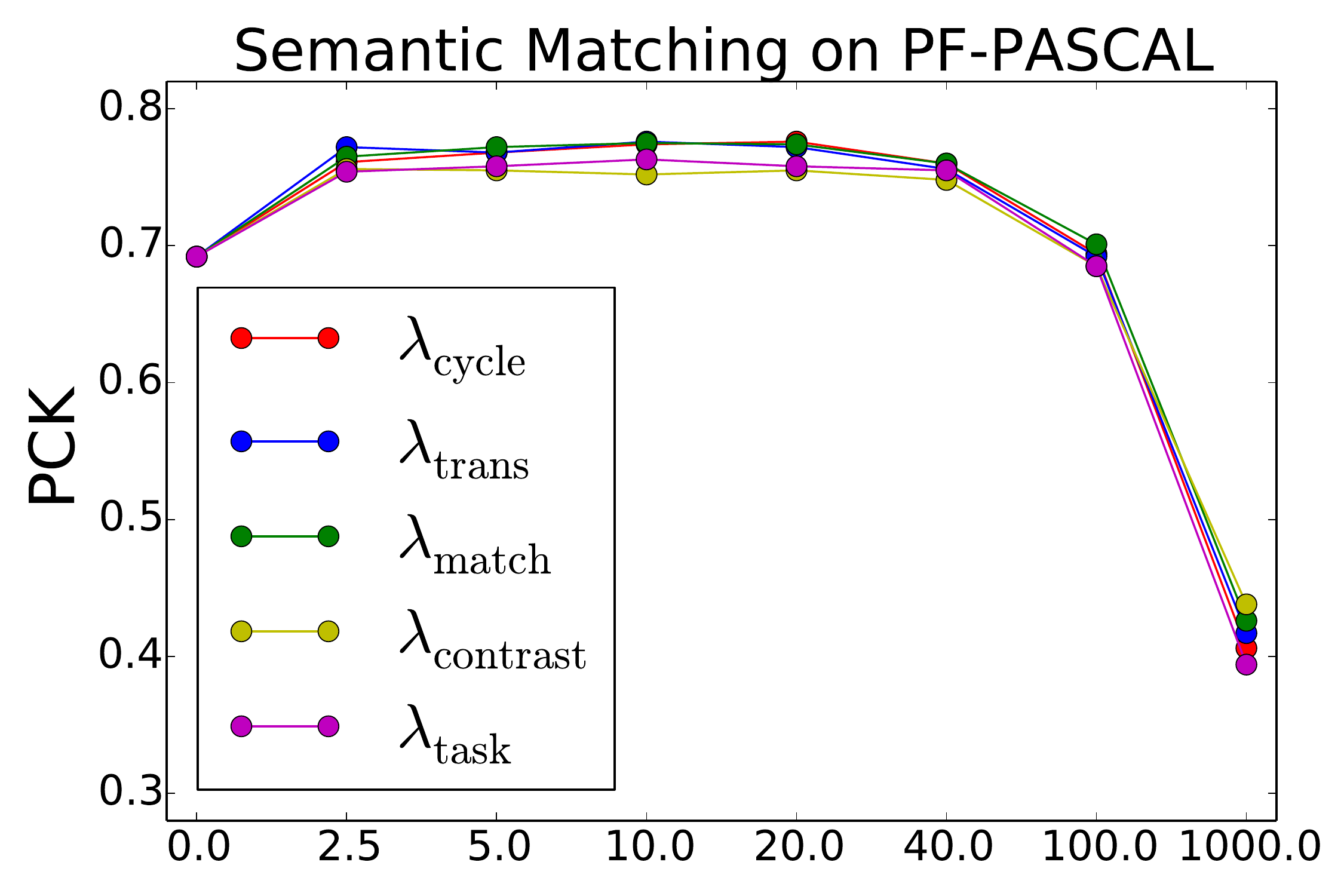}
    \vspace{-5mm}
    \caption{Semantic matching (PCK)}
  \end{subfigure}
  \hfill
  \begin{subfigure}[!t]{\threeimg}
    \includegraphics[width=1.0\linewidth]{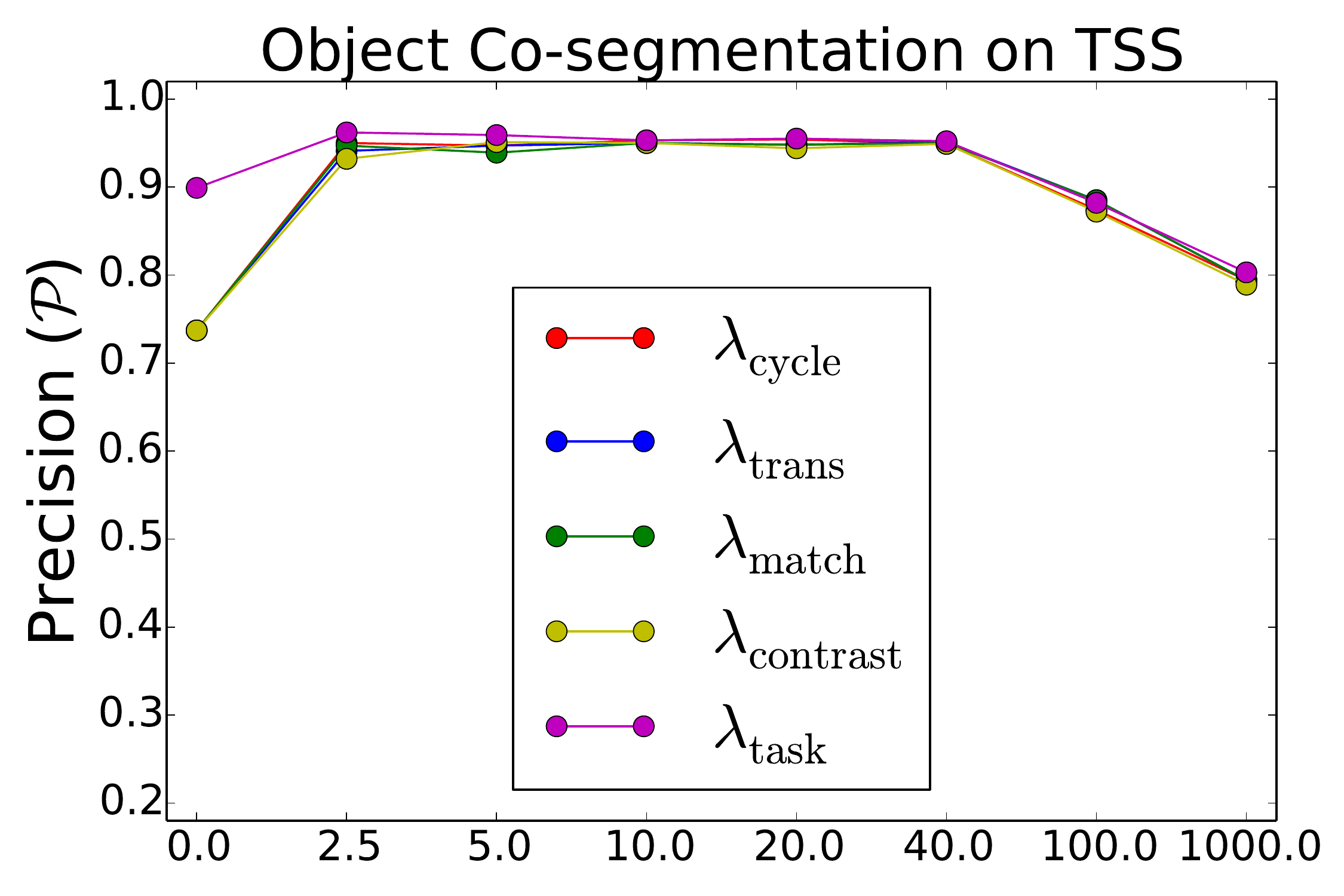}
    \vspace{-6mm}
    \caption{Object co-segmentation ($\mathcal{P}$)}
  \end{subfigure}
  \hfill
  \begin{subfigure}[!t]{\threeimg}
    \includegraphics[width=1.0\linewidth]{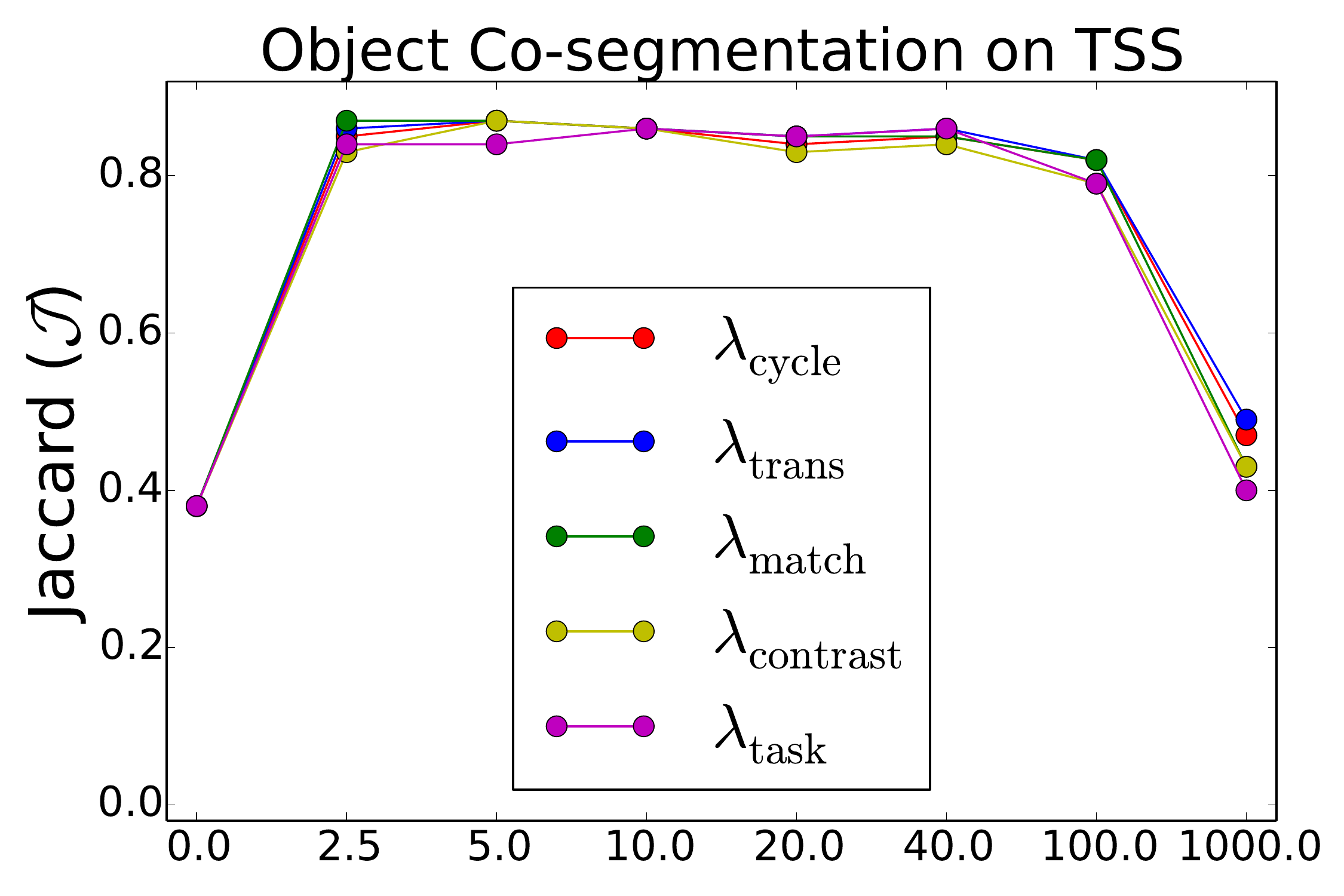}
    \vspace{-6mm}
    \caption{Object co-segmentation ($\mathcal{J}$)}
  \end{subfigure}
  
  \vspace{-2mm}
  \caption{\textbf{Sensitivity analysis of hyper-parameters.}
  The performance of our approach remains stable when the weights for the five loss terms are within a reasonable range.
  %
  }
  \label{fig:sensitivity-hyper-param}
\end{figure*}

The ablation study for object co-segmentation demonstrates that the proposed cross-network consistency loss $\mathcal{L}_\mathrm{task-consis}$ and the perceptual contrastive loss $\mathcal{L}_\mathrm{contrast}$ are crucial to achieving high performance.
On the other hand, the foreground-guided matching loss $\mathcal{L}_\mathrm{matching}$, forward-backward consistency loss $\mathcal{L}_\mathrm{cycle-consis}$, and transitivity consistency loss $\mathcal{L}_\mathrm{trans-consis}$ facilitate object co-segmentation.
For semantic matching, the foreground-guided matching loss $\mathcal{L}_\mathrm{matching}$ and  perceptual contrastive loss $\mathcal{L}_\mathrm{contrast}$ are important to our proposed method.
On the other hand, the cross-network consistency loss $\mathcal{L}_\mathrm{task-consis}$, forward-backward consistency loss $\mathcal{L}_\mathrm{cycle-consis}$, and transitivity consistency loss $\mathcal{L}_\mathrm{trans-consis}$ are helpful for enhancing the generalization ability in semantic matching.


\revised{
{\flushleft {\bf Effect of using different post-processing methods.}} 
To analyze the effect of different post-processing methods for object co-segmentation, we evaluate the performance of our approach using the DenseCRF~\cite{krahenbuhl2011efficient}, Otsu's~\cite{otsu1979threshold}, and GrabCut~\cite{rother2004grabcut} methods on the TSS dataset~\cite{Taniai}.
Table~\ref{table:coseg-post-process} reports the experimental results.
Our results show that using these three post-processing methods achieve similar performance.
}

{\flushleft {\bf Effect of cutoff threshold $m$.}} 
To analyze the sensitivity of our model against the cutoff threshold $m$ in \revised{Eq.}~(\ref{eq:contrast}), we conduct sensitivity analysis on the TSS dataset~\cite{Taniai} by varying the value of the cutoff threshold $m$.
Figure~\ref{exp:sensitivity-margin} shows the experimental results.
When the cutoff threshold $m$ is set to $0$, \ie $d_{AB}^- = 0$ in \revised{Eq.}~(\ref{eq:neg-d-AB}), the model enforces only the inter-image foreground similarity in \revised{Eq.}~(\ref{eq:pos-d-AB}).
Without enforcing intra-image figure-ground dissimilarity in \revised{Eq.}~(\ref{eq:neg-d-AB}), the model may not produce clean foreground-background separation, \revised{suffering from severe performance drops}.
%
%
When increasing the cutoff threshold $m$ to $2$, the results in both precision $\mathcal{P}$ and Jaccard index $\mathcal{J}$ are significantly improved.
When further increasing the cutoff threshold $m$ from $2$ to $5$ or $10$, minimizing the perceptual contrastive loss $\mathcal{L}_\mathrm{contrast}$ is dominated by maximizing the foreground-background distinctness.
The performance of our model drops instead.
Introducing the cutoff threshold $m$ can considerably enhance the perceptual contrastive loss $\mathcal{L}_\mathrm{contrast}$ by setting the cutoff threshold $m$ to an appropriate value.

{\flushleft {\bf Sensitivity analysis.}} 
We analyze the performance of the proposed model by varying the value of each hyper-parameter on the PF-PASCAL~\cite{ProposalFlow} \emph{validation} set for semantic matching, and on the TSS \emph{validation} set for object co-segmentation.
Figure~\ref{fig:sensitivity-hyper-param} presents the experimental results of sensitivity analysis.
For semantic matching, we report the results at PCK threshold $\alpha = 0.1$.
When each of the hyper-parameter is set to $0$ (\ie the corresponding loss function is turned off), our model suffers from performance drops.
When the individual hyper-parameters are set within a reasonable range, the performance is improved significantly.
These results show that each loss function contributes to our method.
However, when the hyper-parameter is set to a large value, \eg $1,000$, the corresponding loss term dominates the full training objective in \revised{Eq.}~(\ref{eq:FullObj}), leading to significant performance drop.
Similar conclusions can be drawn on the object co-segmentation task.

\subsection{Run-time analysis}
Given $800$ images of the TSS dataset~\cite{Taniai} for joint semantic matching and object co-segmentation,
it takes $280$ minutes on a machine with an Intel $i7$ $3.4$ GHz processor and a single NVIDIA GeForce GTX $1080$ graphics card with $11$GB memory. 
The average run-time for processing each image in the set is $21$ seconds.

\revised{
\subsection{Limitations and future work}
Our method may not work for images that contain multiple object instances.
For semantic matching, our method predicts only \emph{one} transformation matrix for a pair of images.
When multiple object instances are present in an image (\eg those in the iCoseg dataset~\cite{batra2011interactively}), our method may not work well since multiple geometric transformations are required.
For object co-segmentation, our method may fail if there exist background patches that are visually similar to the foreground objects.
We note that joint semantic matching and object co-segmentation from images containing multiple object instances can be addressed by instance co-segmentation methods~\cite{hsu2019deepco3} and instance-level semantic matching approaches~\cite{ncnet}.
We leave this problem as future work.
}

\section{Conclusions}

We propose a \emph{weakly-supervised} and \emph{end-to-end trainable} network for joint semantic matching and object co-segmentation.
The core technical novelty lies in the coupled training of both tasks.
We introduce a cross-network consistency loss to encourage the two-stream network to produce a consistent explanation of the given image pair.
The network training requires only weak image-level supervision, making the proposed method scalable to real-world applications.
Through joint optimization, semantic matching is improved owing to the object masks revealed by object co-segmentation, while object co-segmentation is enhanced by referring to cross-image geometric transformations estimated during semantic matching.
Experimental results demonstrate that our approach achieves the state-of-the-art performance for semantic matching and object co-segmentation.



\ifCLASSOPTIONcaptionsoff
  \newpage
\fi



\section*{Acknowledgments}
This work was supported in part by the Ministry of Science and Technology (MOST) under grants MOST 107-2628-E-009-007-MY3 and MOST 109-2634-F-007-013.
M.-H. Yang is supported in part by NSF Career Grant (1149783) and gifts from Google.
J.-B. Huang is supported in part by NSF CRII (1755785).
\bibliographystyle{IEEEtran}
\bibliography{IEEEabrv,reference.bib}

%

\begin{IEEEbiography}[{\includegraphics[width=1in,height=1.25in,clip,keepaspectratio]{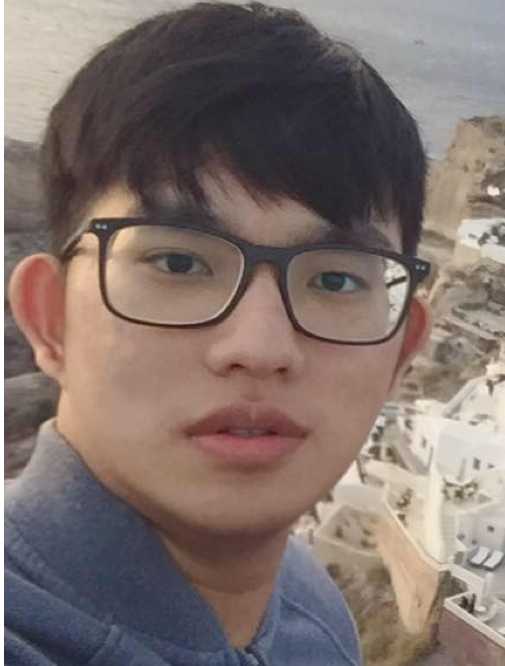}}]{Yun-Chun Chen} received the B.S. degree in electrical engineering from National Taiwan University, Taipei, Taiwan in 2018.
His current research interests include computer vision, deep learning, and machine learning.
\end{IEEEbiography}
\vspace{-8mm}
\begin{IEEEbiography}[{\includegraphics[width=1in,height=1.25in,clip,keepaspectratio]{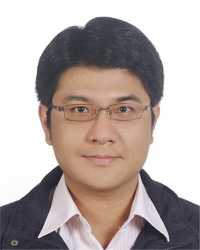}}]{Yen-Yu Lin} received the B.B.A. degree in information management, and the M.S. and Ph.D. degrees in computer science and information engineering from National Taiwan University, in 2001, 2003, and 2010, respectively. 
He is a Professor with the Department of Computer Science, National Chiao Tung University. 
Prior to that, he worked for the Research Center for Information Technology Innovation, Academia Sinica from January 2011 to July 2019.
His current research interests include computer vision, machine learning, and artificial intelligence. 
\end{IEEEbiography}
\vspace{-8mm}
\begin{IEEEbiography}[{\includegraphics[width=1in,height=1.25in,clip,keepaspectratio]{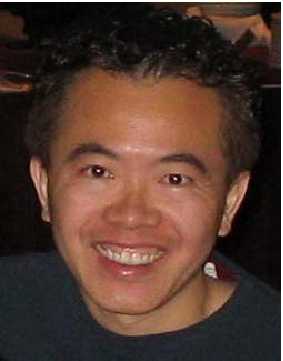}}]{Ming-Hsuan Yang}
is a Professor in Electrical
Engineering and Computer Science at University
of California at Merced and an adjunct professor at Yonsei University. 
%
Yang served as a program co-chair for the 2019 IEEE International Conference on Computer Vision.
He served as an associate editor of the IEEE Transactions on Pattern Analysis and Machine Intelligence from 2007 to 2011, and is an associate editor of the International Journal of Computer Vision, Image and Vision Computing and Journal of Artificial Intelligence Research. 
Yang received the NSF CAREER award and Google faculty award. 
He is a Fellow of the IEEE.
\end{IEEEbiography}
\vspace{-8mm}
\begin{IEEEbiography}[{\includegraphics[width=1in,height=1.25in,clip,keepaspectratio]{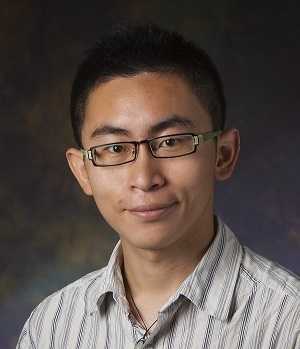}}]{Jia-Bin Huang} is an assistant professor in the Bradley Department of Electrical and Computer Engineering at Virginia Tech.
He received the B.S. degree in Electronics Engineering from National Chiao-Tung University, Hsinchu, Taiwan and his Ph.D. degree in the Department of Electrical and Computer Engineering at University of Illinois, Urbana-Champaign in 2016. 
\end{IEEEbiography}

%






\end{document}